\documentclass[10pt,twocolumn,letterpaper]{article}

\usepackage{wacv}
\usepackage{times}
\usepackage{epsfig}
\usepackage{graphicx}
\usepackage{amsmath}
\usepackage{amssymb}
\usepackage{xspace}
\usepackage{adjustbox}
\usepackage{booktabs}
\usepackage{amsmath}
\usepackage{relsize}
\usepackage[accsupp]{axessibility}

\usepackage[dvipsnames]{xcolor}
\usepackage{mathrsfs}
\usepackage{amsfonts}
\usepackage{pifont}

\usepackage{tabularx, makecell}
\usepackage{xcolor,colortbl}
\newcommand{\imgWidth}{0.2\textwidth}

\usepackage{subcaption}
\usepackage{cuted}

\makeatletter
\long\def\@makecaption#1#2{
   \setbox\@tempboxa\hbox{\small \noindent #1.~#2}
   \setlength{\@ctmp}{\hsize}
   \addtolength{\@ctmp}{-\@figindent}\addtolength{\@ctmp}{-\@figindent}
   \ifdim \wd\@tempboxa >\@ctmp
      {\small #1.~#2\par}
   \else
      \hbox to\hsize{\hfil\box\@tempboxa\hfil}
  \fi}
\makeatother

\usepackage{arydshln}

\usepackage{xspace}
\usepackage{adjustbox}
\usepackage{booktabs}
\usepackage{amsmath}
\usepackage{relsize}
\usepackage{mathrsfs}
\usepackage[ruled, vlined, noend]{algorithm2e}
\usepackage{nicefrac}
\usepackage{enumitem}

\usepackage{amsfonts}
\usepackage{pifont}
\newcommand{\cmark}{\ding{51}}%

\newcommand{\N}{\mathbb{N}}
\newcommand{\R}{\mathbb{R}}

\newcommand{\loss}{\mathcal{L}}
\newcommand{\clientset}{\mathcal{K}}
\newcommand{\clusterset}{\mathcal{C}}

\newcommand{\fl}{\textsc{FL}\xspace}

\newcommand{\semseg}{\textsc{SS}\xspace}
\newcommand{\sfda}{\textsc{SFDA}\xspace}

\newcommand{\setting}{\textsc{FFreeDA}\xspace}
\newcommand{\method}{\textsc{LADD}\xspace}

\DeclareMathOperator*{\argmax}{arg\,max}
\DeclareMathOperator*{\argmin}{arg\,min}

\definecolor{ForestGreen}{RGB}{34,139,34}
\definecolor{pinkcarino}{RGB}{255,0,255} 
\definecolor{lightgray}{RGB}{240,240,240}

\newcommand{\myparagraph}[1]{\vspace{4pt}\noindent\textbf{#1}}

\newcommand\blfootnote[1]{%
  \begingroup
  \renewcommand\thefootnote{}\footnote{#1}%
  \addtocounter{footnote}{-1}%
  \endgroup
}

\def\wacvPaperID{909} %

\wacvalgorithmstrack   %

\wacvfinalcopy %

\ifwacvfinal
\usepackage[breaklinks=true,bookmarks=false,hyperfootnotes=false]{hyperref}
\else
\usepackage[pagebackref=true,breaklinks=true,colorlinks,bookmarks=false]{hyperref}
\fi

\begin{document}

\title{Learning Across Domains and Devices: \\Style-Driven Source-Free Domain Adaptation in Clustered Federated Learning}

\author{Donald Shenaj$^{*,1}$, Eros Fan\`i$^{*,2}$, Marco Toldo$^1$, Debora Caldarola$^2$, Antonio Tavera$^2$, \\ 
Umberto Michieli$^{\dag,1}$, Marco Ciccone$^{\dag,2}$, Pietro Zanuttigh$^{\dag,1}$, and Barbara Caputo$^{\dag,2}$ \and
$^1$University of Padova, Italy \\  
\and
$^2$Politecnico di Torino, Italy 
}

\maketitle
\thispagestyle{empty}

\begin{abstract}
Federated Learning (\fl) has recently emerged as a possible way to tackle the domain shift in real-world Semantic Segmentation (\semseg) without compromising the private nature of the collected data.
However, most of the existing works on \fl unrealistically assume labeled data in the remote clients. %
Here we propose a novel task (\setting) in which the clients' data is unlabeled %
and the server accesses a source labeled dataset for pre-training only. To solve \setting, we propose \method, which leverages the knowledge of the pre-trained model by employing self-supervision with ad-hoc regularization techniques for local training and introducing a novel federated clustered aggregation scheme based on the clients' style. Our experiments show that our algorithm is able to efficiently tackle the new task outperforming existing approaches.
The code is available at \url{https://github.com/Erosinho13/LADD}.

\blfootnote{*: Equal contribution. \dag: Equal supervision.}
\end{abstract}

\vspace{-1cm}
\section{Introduction}

Federated Learning (\fl) \cite{fedavg, fedprox, feddyn, scaffold, fedproto, gboard, fl_intro, leaf} is a relatively new field of research that is attracting increasing interest. In \fl, a learning task is solved through a collaboration among several edge devices, \ie, clients, coordinated by a central server \cite{fedavg}. This learning paradigm is useful when data cannot be freely shared due to regulations, laws, and ethical principles: \fl allows training a global model without leaking the users' data, preserving their privacy.

\begin{figure}
    \centering
    \includegraphics[width=\columnwidth]{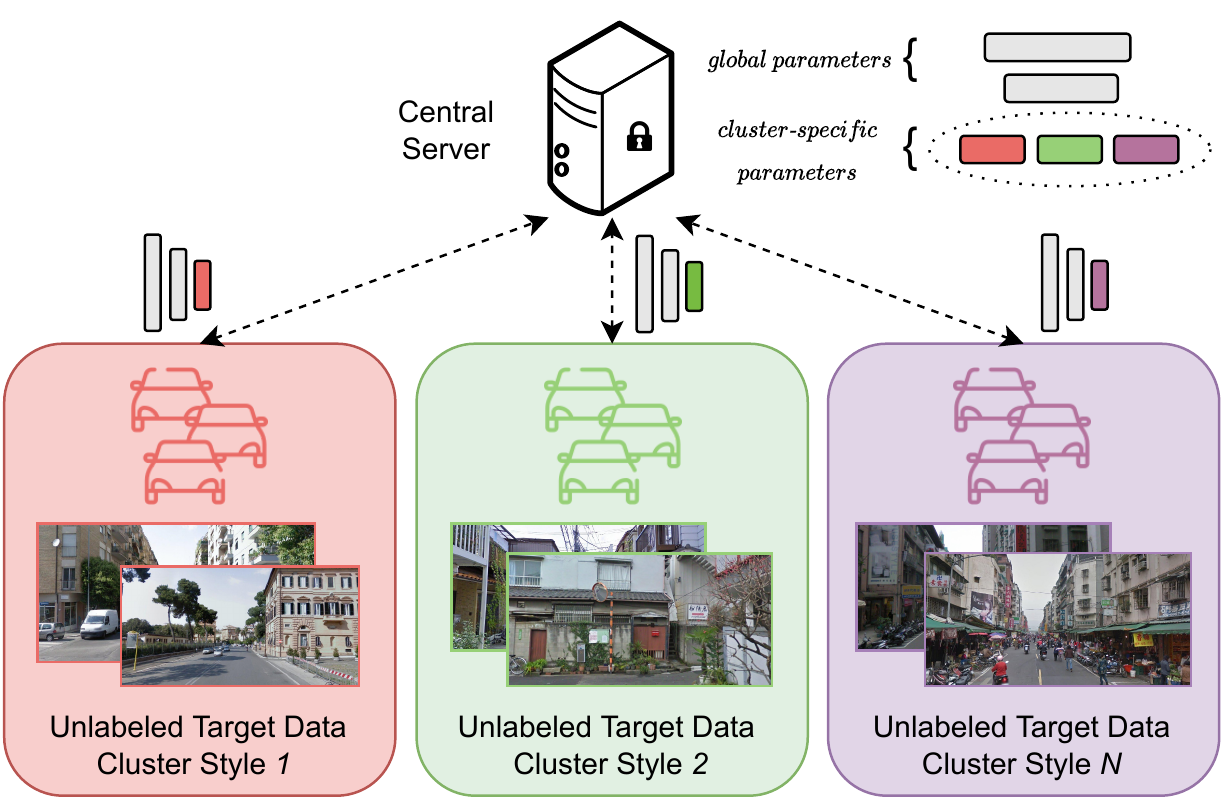}
    \caption{\setting overview: clients having similar appearance are clustered together, while local learning is carried out exploiting both global and cluster-specific parameters. The clients' data is unlabeled and the source labeled dataset is kept on the server.}
    \label{fig:ga}
\end{figure}

As an example, \fl also constitutes a practical solution to tackle real-world vision tasks with data collected from multiple users in different scenarios. For instance, in the case of Semantic Segmentation (\semseg), it can be employed by self-driving cars for obstacle detection, and avoidance \cite{FedDrive}. Most existing \fl works assume the availability of labeled data on the client side. This assumption is clearly unrealistic due to the high cost and amount of manual work needed for dense pixel-level annotations \cite{main_ref}. %

In this work, we focus on autonomous driving applications introducing a novel, more realistic setting for \semseg: \textbf{F}ederated source-\textbf{Free} \textbf{D}omain \textbf{A}daptation (\setting). In \setting, the server can pre-train the model on labeled source data. However, further accessing the source data is forbidden as in the Source-Free Domain Adaptation (\sfda) setting~\cite{sfda_original}. %
Clients access only their \textit{unlabeled} target dataset, which they cannot share with other clients or the server. In particular, we consider real-world scenarios with several clients, each with a limited amount of images. 

After the pre-training phase, the training setting is fully unsupervised. However, the objective of \setting is not only to solve a multi-target domain adaptation problem in \semseg, rather to tackle specific issues arising in \fl, such as \textit{statistical} and \textit{system heterogeneity} \cite{fedprox,fairavg}, \textit{communication bottleneck} \cite{hamer2020fedboost}, and clients' \textit{privacy preservation} \cite{li2020federated,bonawitz2019towards}. 
To the best of our knowledge, no previous works addressed this problem and their related issues at the same time.

To address the \setting problem, we propose \method, a novel federated algorithm that assumes the presence of multiple distributions hidden among the clients. To exemplify, it is reasonable to assume that self-driving cars within the same city collect similar images. Indeed, the geographical proximity of two self-driving cars and different weather conditions could make the local datasets more or less similar. Therefore, \method partitions the clients into clusters based on the styles of the images belonging to each client, trying to match them with their actual latent distribution. 
To minimize parameters duplication and improve communication efficiency, \method splits the model's parameters in \textit{shared}, globally aggregated across all clients, and \textit{cluster-specific} which are aggregated only across clients within the same cluster, as visible in Fig. \ref{fig:ga}.
Moreover, \method takes full advantage of the source dataset during the pre-training stage with style transfer data augmentation \cite{FDA}, randomly loading the target styles in the source images to mimic the target distributions. Finally, \method also leverages self-training through an ad-hoc pseudo-labeling strategy and stabilizes  training with regularization techniques. %

As \setting is a novel setting, we adapted several baselines from other settings. \method outperforms all baselines, showing the importance of designing specific algorithms for the proposed setting. To summarize: %
\setlist{nolistsep}
\begin{itemize}[noitemsep]
    \item We introduce \setting, a novel \semseg task for \fl where we dropped the unrealistic assumption of dense labeled data at client side. 
    \item We propose two realistic benchmarks for it, based on the Mapillary Vistas \cite{mapillary} and CrossCity \cite{crosscity} datasets. 
    \item We propose \method, a new federated algorithm tackling \setting based on style-transfer and clustering. 
    \item \method shows excellent performance on all benchmarks with a  source dataset (GTA5 \cite{gta5}) and three different targets (Cityscapes \cite{cityscapes}, CrossCity, Mapillary), with diversified splits of the data across the clients.
\end{itemize}
\section{Related Work}

\textbf{Semantic Segmentation (SS)}, \ie, classifying each pixel of an image with the corresponding semantic class, is an important challenge in many use cases such as self-driving cars \cite{feng2020deep}. 
State-of-the-art SS models rely on an encoder-decoder architectures, based on CNNs \cite{ref_fcn, ref_deeplab,deeplabv3,ref_pspnet, mobilenetv2,hassani2021escaping} or transformers~\cite{dosovitskiy2020vit,liu2021swin,chu2021twins,xie2021segformer} to generate dense predictions.
These approaches typically assume a simplified, centralized setting in which the whole training dataset is available on a central server. %
However, this is not always possible due to privacy and efficiency constraints, and distributed training solutions must be considered.

\textbf{Domain Adaptation (DA).}
Being a complex structured prediction task, SS generally requires expensive dense annotations. Recently, an increasing number of methods \cite{toldo20review, csurka2021unsupervised} tackle this by training on synthetic data generated in virtual environments~\cite{gta5, synthia, idda, selma}. Nonetheless, models trained on these data fail to generalize to the real world because of the inherent domain shift between the simulated and real distributions. DA %
aims at reducing the performance gap between a \textit{source domain} on which a model has been trained and a \textit{target} one. When the target data is unlabeled, this is called Unsupervised DA (UDA).
Initially, DA methods attempted to close the gap by measuring domain divergence \cite{pmlr-v37-long15, tzeng_mcd, MCD}. Another popular direction is adversarial training \cite{adaptsegnet, clan,michieli20tiv}, which includes the segmentation network and a domain discriminator competing in a \textit{minimax} game. Other applications attempt to reduce domain shift by employing image-to-image translation algorithms to generate images modified with the style of the other domain \cite{hoffman18cycada, pizzati2020domain,toldo20cycle}. Since this is a time-consuming technique, some non-trainable style translation algorithms, such as FDA \cite{FDA}, have been introduced. 
Modern approaches \cite{pycda, cbst, barbato2021latent, daformer} use self-learning techniques to create pseudo-labels from the target data, allowing the model to be fine-tuned even in a federated scenario in which each client observes its unlabeled domain.

\textbf{Main Challenges in \fl.} %
Clients in \fl  %
have different hardware  capabilities (\textit{system heterogeneity}) %
and their data may belong to different distributions (\textit{statistical heterogeneity}). 
Additionally, clients-server communication should be
efficient \cite{hamer2020fedboost} and privacy must be preserved
by preventing the server to access clients' local data \cite{li2020federated,bonawitz2019towards}.

\textbf{Vision Tasks in \fl.} Thanks to its many applications in the real world and its potential in managing sensitive data, \fl \cite{fedavg} has recently captured the interest of the research community \cite{fl_intro,kairouz2021advances,zhang2021survey}. However, most research papers focus on the theoretical aspects of \fl \cite{fedprox,scaffold,feddyn,FedUL}, 
neglecting its application to more complex vision tasks, \eg, SS, and realistic scenarios, \eg, heterogeneous domain distribution and unlabeled data observed at clients. A few exceptions are \cite{fedproto, FedDrive, caldarola2022improving}, which study \fl SS and \fl in the context of autonomous driving, and \cite{sheller2018multi,LiWenqi2019PFBT,BerceaCosminI2021FDFL,FedCross,xu2022closing} that leverage medical images. Their main limitation is the costly assumption of having labeled data available.

In \cite{FedUL}, the authors deal with the novel unsupervised \fl setting from strong theoretical assumptions whilst only focusing on classification tasks for simple datasets such as MNIST \cite{mnist}, and CIFAR10 \cite{cifar10}.
Focusing on SS, and proposing a more realistic approach, \cite{main_ref} introduces FMTDA (Federated Multi-target Domain Adaptation) to handle a few clients with unlabeled target local datasets belonging to different distributions while maintaining an open-access labeled source dataset on the server-side. 
Inspired by this work, we investigate the more complex setup of SFDA~\cite{liu2021source}, in which the source dataset is only visible on the server during the pre-training phase and is not available to the clients. Moreover, we study a more realistic scenario where many more clients collaborate in the training but access much less data. As in FMTDA, we assume that clients' data may differ in terms of visual domains, \eg, the scenes collected by the autonomous vehicles in different geographical locations may have different weather or light conditions or may not show some semantic classes. 

The study of DA in \fl (both UDA and SFDA) is still in its early stages: \cite{zhuang2022federated} leverages UDA techniques for face recognition, \cite{Peng2020Federated} tackles domain shift via adversarial approaches, while \cite{main_ref} sees each client as a distinct target domain. To the best of our knowledge, this is the first work adapting SFDA to \fl.
Additional insights on vision tasks in \fl other than SS and DA are reported in \cite{aledhari2020federated}.

\textbf{Clustered \fl (CFL).} In a real-world context, subsets of users typically share some common characteristics: for example, users in nearby geographic locations experience cities with similar architecture or weather conditions. Therefore, clients can be partitioned into clusters, each representing a specific set of conditions that we match to a corresponding \textit{style} \cite{genadapt}. %
This approach falls under the literature of CFL \cite{sattler2020clustered}, in which clustering is usually exploited for building personalized models that work well in a specific subdomain of interest \cite{fallah2020personalized,ghosh_clustering2020,caldarola2021cluster}. Differently from these methods, we cluster clients based on the styles extracted from the unlabeled samples seen by each client.

\section{Problem Setting}

\begin{figure*}
    \centering
    \includegraphics[width=2.1\columnwidth]{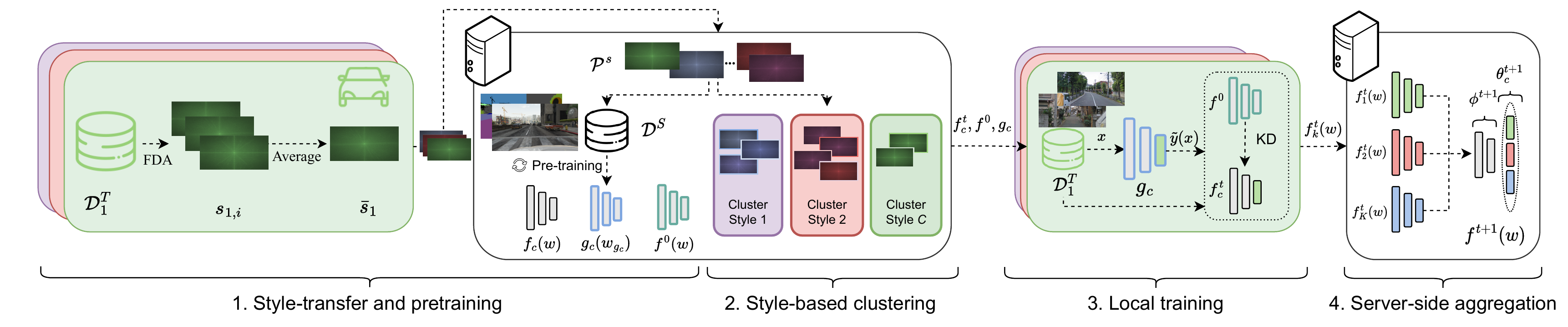}
    \caption{\method overview (best seen in colors). 1) Each client $k$ extracts the average style $\bar{s}_k$ of its local data $\mathcal{D}_k^T$ using FDA. At server-side, the collected styles $\mathcal{P}^s$ are applied to the source dataset $\mathcal{D}_s$ during the supervised pre-training. 2) Clients are clustered according to their style. 3) At client-side, the cluster-specific teacher $g_c$ outputs the pseudo-labels, used for training $f_c^t$, leveraging KD from the pre-trained model. 4) At the server-side aggregation, we distinguish between global ($\phi^{t+1}$) and cluster-specific parameters ($\theta_c^{t+1}$).}
    \label{fig:arch}
\end{figure*}

In this section, we formalize the proposed Federated Source Free Domain Adaptation (\setting) setting. %

Given a central server and the set of all clients $\clientset$ with $\left | \clientset \right | = K$, the input space $\mathcal{X}$, the output space $\mathcal{Y}$ and $N_p$ pixels in each image, the datasets are distinguished as follows: the source dataset $\mathcal{D}^S$ is kept on the server-side and is 
made of pairs of image and segmentation label %
$(x^S, y^S) \sim \mathscr{P}^S(\boldsymbol{x}, \boldsymbol{y})$, where $\boldsymbol{x}$ and $\boldsymbol{y}$ are the random variables following the distribution $\mathscr{P}^S$, associated with $x^S\in\mathcal{X} \text{ and } y^S\in\mathcal{Y}^{N_p}$ %
respectively; the $K$ target training datasets $\mathcal{D}^T_k = \{x_{k,i}^T\in \mathcal{X} \: \forall i \in |\mathcal{D}^T_k|\}$ are local to each client $k\in[K]:=\{0,1,...,K-1\}$   %
and 
$x_{k,i}^T \sim \mathscr{P}^T_k(\boldsymbol{x})$. %

By definition of the SFDA scenario, the source and test datasets share the same set of categories $\mathcal{Q}=\mathcal{Q}^S = \mathcal{Q}^T$. As for the federated setting, $K$ is reasonably large and the local datasets differ in terms of both size and distributions but have typically a much smaller size than the source dataset. Since users may share some common characteristics, it may happen that $\mathscr{P}^T_k(\boldsymbol{x}) = \mathscr{P}^T_h(\boldsymbol{x})$ for some $k, h\in[K]$. We assume the local datasets to be drawn from the same meta-distribution, which contains $G$ latent visual domains (\eg, different cities), and each $\mathcal{D}_k^T$ contain only images from one of the $G$ latent domains.  %
The test dataset $\mathcal{D}_{test}^{T}$ follows the target distribution $\mathscr{P}^T$ and is used to evaluate the final model learned across domains and devices.

Given the model $f(w): \mathcal{X} \rightarrow \R^{N_p\times|\mathcal{Y}|}$ parametrized by $w$, the global objective is to obtain optimal segmentation performance on the target data distribution $\mathscr{P}^T(\boldsymbol{x})$, and it can be achieved by minimizing a suitable loss function, \ie:
\begin{equation}
    w^* = \argmin_w \sum_{k\in[K]} \frac{|\mathcal{D}_k^T|}{|\mathcal{D}^T|}\loss_k(w)
\end{equation}
where $\loss_k$ is the local loss function and $\mathcal{D}^T=\bigcup_{k \in \clientset} \mathcal{D}^T_k$. %

\section{Method}
In this section, we describe in detail our FL algorithm by detailing the pre-training strategy (Sec.~\ref{sec:pretraining}), the aggregation (Sec.~\ref{sec:style_aggr}) and the adaptation techniques (Sec.~\ref{sec:adaptation}).
The procedure is summarized in Fig.~\ref{fig:arch} and Algorithm~\ref{alg:ladd}.

\begin{algorithm}[t]
\scriptsize
\SetAlgoLined
\SetKwBlock{Clustering}{Clustering of the clients $\clientset$ and Pre-Training of $f$ on $\mathcal{D}^S$}{}
\SetKwBlock{Adaptation}{Adaptation of $f$ on $\mathcal{D}^T$}{}
\textbf{Require:}\\
    Source (labeled) dataset $\mathcal{D}^S$, clients $k\in\clientset$ with target (unlabeled) datasets $\mathcal{D}^T_k$, global model $f(w)=f(\{\theta,\phi\})$\\%, $T$ rounds\\
\Clustering{
    Extract the styles $\mathcal{P}^s_k$ for each $k\in\mathcal{K}$\\
    Define the style-based clusters $\clusterset$ (refer to Algorithm \ref{alg:clustering})\\
    Train $f(w)$ on $\mathcal{D}^S$ with style-transfer from $\mathcal{P}^s = \bigcup_{k \in \clientset} \mathcal{P}_k^s$
}
\Adaptation{
    \textbf{Initialize:} \\
    Cluster models $f_c(w_c) = f(w)$ and teachers $g_c(w_{g_c}) = f(w)$ \\
    \For{\textbf{each} round $t \in [T]$}{
        Randomly extract $\clientset^t \subset \clientset$. Let $c := {\Gamma_\clusterset(k)}$.\\
        \For{$k \in \clientset^t$ in parallel}{
            Set $f_k(w_k) = f_c(w_c)$ \\
            $\phi_k^t, \theta_k^t \leftarrow$ \textsc{ClientUpdate}($f_k$, $g_c$, $f$, $\mathcal{D}^T_k$) (Sec. \ref{sec:adaptation})
        }
        $\phi^{t+1} \leftarrow$ Aggregate $\phi_k^t$ globally \\
        $\theta^{t+1}_c \leftarrow$ Aggregate $\theta_k^t$ within the cluster $c$ \\
        \If{$t \mod \omega \equiv 0$}{
            \If{$t \geq t_\textsc{start}$}{
                $g_c^{t+1}(w_{g_c}) = \textsc{SWAtUpdate}\big(g_c^t\big) \: \forall c$ (Sec. \ref{sec:adaptation})
            }
            \Else{
                $g_c(w_{g_c}) = f_c^t(w_c^t) \: \forall c\in \mathcal{C}$
            }
        }
    }
}
\caption{\small{\method} \scriptsize{(Learning Across Domains and Devices)}} 
\label{alg:ladd}
\end{algorithm}

\subsection{Server Pre-training} \label{sec:pretraining}
The first step of \method is a pre-training stage on the labeled source dataset $\mathcal{D}^S$. Before training, to bring the styles of source and target images closer together and improve the generalization of the pre-trained model, we apply the FDA \cite{FDA} style transfer technique. First, the clients' style is transferred to the server and then applied to $\mathcal{D}^S$, on which the model is trained. Specifically, the $k$-th client extracts the style $s_k$ from each of its images, given by a window of width $l_s$ located at the center of the amplitude spectrum of that image \cite{FDA}, \ie representing the amplitude of the lowest spatial frequency coefficients. Critically, these coefficients do not contain relevant information on the scene content, thus not breaking the user's privacy. The pool of styles $\mathcal{P}_k^s$ extracted from client $k$ is populated by sending the 
average of its extracted styles, \ie $\mathcal{P}_k^s = \{\bar{s}_k\}$. %
On the server-side, the randomly initialized model $f(w)$ is trained on the source dataset $\mathcal{D}^S$, augmenting the source images with random styles extracted from the set $\mathcal{P}^s = \bigcup_{k \in \clientset} \mathcal{P}_k^s$. 
It is worth noting that these styles are never shared among the clients, and even a few images are sufficient to compute $\bar{s}_k$ on each client. Once the pre-training stage is completed, the source dataset is no longer used.

\subsection{Style-based Aggregation}
\label{sec:style_aggr}

In a realistic \fl setting, different clients may observe similar samples, \eg self-driving cars in the same region are likely to collect similar images and are not subject to statistical heterogeneity \cite{fedprox} during the server aggregation. On the other side, this premise does not apply to self-driving cars scattered throughout various distant places, which may learn conflicting information, thus affecting performances if naively aggregated.
In addition, users may have access to a limited number of images, hindering clients from generalizing only from local optimization~\cite{caldarola2022improving}.

Taking these factors into account, we propose to explicitly cluster the clients to find their $G$ latent visual domains.
To this end, we partition $\clientset$ in a set of non-empty clusters $\clusterset$ with $\left | \clusterset \right | = C$ and $\sum_{c \in \clusterset} \left | c \right | \! = \! K$ basing on the clients' transferred styles. The centroid $\mu_c$ of each cluster is computed using $\bar{s}_k, \! \forall  k\in \mathcal{C}$. We summarize our approach in Algorithm~\ref{alg:clustering}. 
We refer to $\textsc{h}$-Means instead of K-Means for symbolic convenience. First, we compute $\textsc{h}$-Means $N$ times $\forall \textsc{h} \in [n]_m$, with $N$, $n$, $m$ positive integers. Then, for each value of $\textsc{h}$, we select the partition $\clusterset_\textsc{h}$ with the smallest intra-cluster distance, and compute its Silhouette Score. Finally, we select the clustering $\clusterset$ with the highest Silhouette Score.

\begin{algorithm}
\scriptsize
\SetAlgoLined
\SetKwBlock{intracl}{$\textsc{IntraClusterDist}(\clusterset, k)$}{}
\SetKwBlock{intercl}{$\textsc{InterClusterDist}(\clusterset, k)$}{}
\SetKwBlock{silhouette}{$\textsc{SilhouetteScore}(\clusterset, a, b)$}{}
Let $d(\cdot, \cdot)$ be the \textit{L2-norm} operator. \\
\vspace{0.7\baselineskip}
\textbf{Require}: \\
Clients $k\in\clientset$, target datasets $\mathcal{D}^T_k \hspace{.5em} \forall k \in [K]$, function $\Gamma$ assigning each client to one of the $C$ clusters. Hyper-params $n, m, N \in \N_0$, $m < n$\\
\For{$\textsc{h} \in [n]_m := \{m, m+1, ..., n-1\}$}{
    \For{$n \in [N]$}{
        Change random seed $rs$\\
        $\clusterset_\textsc{h}^{rs} = \textsc{h-means}$\\
        Compute $a_k(\clusterset_\textsc{h}^{rs}) = \textsc{IntraClusterDist}(\clusterset_\textsc{h}^{rs}, k) \hspace{.5em} \forall k \in \clientset$
    }
    $\clusterset_\textsc{h} = \argmin_{\clusterset^{rs}_{\textsc{h}}} \sum_{k \in \clientset}a_k(\clusterset^{rs}_{\textsc{h}})$ \\
    Define $a_k^\textsc{h} := a_k(\clusterset_\textsc{h}) \hspace{.5em} \forall k \in \clientset$ \\
    Compute $b_k^\textsc{h} := b_k(\clusterset_\textsc{h}) = \textsc{InterClusterDist}(\clusterset_\textsc{h}, k) \hspace{.5em} \forall k \in \clientset$ \\
    Compute $\bar{\sigma}(\clusterset_\textsc{h}) = \textsc{SilhouetteScore}(a_k^\textsc{h}, b_k^\textsc{h} \hspace{.5em} \forall k \in \clientset)$
}
\textbf{return} ${\clusterset = \argmax_\textsc{h} \bar{\sigma}(\clusterset_\textsc{h})}$ \\
\vspace{0.7\baselineskip}
\intracl{
    \textbf{return} $\frac{1}{\left | \Gamma_\clusterset(k) \right | - 1} \sum_{h \in \Gamma_\clusterset(k), h \neq k} d(k, h)$ 
}
\intercl{
    \textbf{return} $\min_{c \in \clusterset, c \neq \Gamma_\clusterset(k)} \frac{1}{\left | c \right |} \sum_{h \in c} d(k, h)$
}
\silhouette{
    $\sigma_k = \frac{b_k - a_k}{\max \left (a_k, b_k \right )}$ if $\left | \Gamma_\clusterset(k) \right | > 1$, 0 otherwise, $\forall k \in \clientset$ \\
    \textbf{return} $\frac{1}{K}\sum_{k \in \clientset} \sigma_k$
}
\caption{Clustering Selection algorithm.}
\label{alg:clustering}
\end{algorithm}

As for the server-side aggregation, %
instead of averaging the updates of the selected clients at each round as done by the standard FedAvg \cite{fedavg} algorithm, \method introduces a 
clustered and layer-aware aggregation policy. We define as: (i) $w_k^t$  the weights of the model of client $k$ after $E$ local epochs of training at round $t$; (ii) $\theta_k^t$ and $\phi_k^t$ the group of \textit{cluster-specific} and \textit{global} parameters of the local model, such that $w_k^t = \theta_k^t \cup \phi_k^t$ and $\theta_k^t \cap \phi_k^t = \emptyset$;
(iii) $\clientset^t \subset \clientset$ the subset of clients selected at round $t$. 
We globally aggregate the global parameters $\phi_k^t$ over all the selected clients in $\clientset^t$ to obtain the new parameter set $\phi^{t+1}$. On the other side the cluster-specific parameters $\theta_k^t$ are averaged within the clusters, resulting in $C$ specific parameter sets $\theta_c^{t+1}$ and 
$C$ models $f_c^{t+1}(\boldsymbol{x}; w_c^{t+1}), \forall\: c \in \clusterset$ where $w_c^{t+1} = \phi^{t+1} \cup \theta_c^{t+1}$. Note that the server is not required to store independent models for each cluster, it is sufficient to save only the cluster-specific parameters $\theta_c^{t+1} \hspace{.5em} \forall c \in \clusterset$ and the global parameters $\phi^{t+1}$, loading them when needed. 
At test time, %
given the $i$-th target test image, we (i) extract the style $s_{\textsc{test}, i}$, (ii) compute the \textit{L2-norm} between $s_{\textsc{test}, i}$ and all the cluster centroids $\mu_c \hspace{.5em} \forall c \in \clusterset$%
, (iii) select the cluster $c$ with the smallest \textit{L2-norm}, and (iv) use the model $f_c^{t+1}(\boldsymbol{x}; w_c^{t+1})$ to evaluate the model on the $i$-th test image. 
    
\subsection{Client Adaptation}
\label{sec:adaptation}
    
    Here we introduce the main components of the local loss $\loss_k(w)$ and the employed regularization techniques. %
    
    \myparagraph{Self-Training.} At round $t$, given an image $x$ on the $k$-th client, we train the local model $f^t_k(\boldsymbol{x}; w_k^t)$ by employing hard one-hot pseudo-labels $\tilde{y}(x) \in \R^{Q\times N_p}$ with the same thresholding mechanism as proposed in \cite{FDA}, where ${Q = |\mathcal{Q}|}$ is the number of classes. %
    To reduce the computation burden on the client, we avoid having client-specific teacher networks. On the contrary,
    the pseudo-labels are computed using a cluster-specific teacher network $g_c^t(\boldsymbol{x}; w_{g_c}^t)$, which outputs the predictions ${\hat{y}(x) := g_c^t(x; w_{g_c})}$. %
    The teacher parameters $w_{g_c}^t$ are first initialized as $w_{g_c}^0 = w$ and then updated every $\omega$ rounds as $w_{g_c}^t = w_c^t$.

    \myparagraph{Regularization.} Pseudo-labels allow the clients to mimic the presence of the labels. However, after a few training iterations, the learning curve starts dropping \cite{urma}. %
    At first, self-training allows to reduce the gap between the knowledge extracted by $\mathcal{D}^S$ and the one needed to perform well on the target datasets $\mathcal{D}^T_k$. Later on though the network starts being too confident on its predictions, reducing its effectiveness and making more miss-classifications. It therefore becomes of the utmost importance to try to reverse this trend. To this end, we exploit a \textbf{Knowledge Distillation} (KD) loss $\loss_\textsc{KD}$ 
    \cite{kdloss} based on the soft predictions given by the pre-trained model to prevent $f^t(w)$ from forgetting the knowledge acquired during the pre-training phase. However, our experiments showed KD on its own was not enough for avoiding overfitting, since the learning curve starts slightly dropping again during the last rounds of adaptation (see Suppl.\ Mat.\ for more details). Inspired by the recent success of Stochastic Weight Averaging (SWA) \cite{swa} in FL \cite{caldarola2022improving}, we apply a moving average to the clients' teachers $g_c$ after a starting round $t_\textsc{start}$ as
        $w_{g_c}^{t+\omega} = \nicefrac{(w_{g_c}^{t} n_{g_c}^t + w_c^{t+\omega})}{(n_{g_c}^t + 1)}$,
    where $n_{g_c}^t = \nicefrac{(t - t_\textsc{start})}{\omega}$. We name this technique \textbf{SWA teacher} (SWAt). SWAt allows noise reduction, further stabilizes the learning curve and enables the model to better converge to the local minimum of the total loss $\mathcal{L} = \mathcal{L}_\textsc{pseudo} + \lambda_\textsc{kd}\mathcal{L}_\textsc{kd}$, where $\lambda_\textsc{kd}$ is an hyper-parameter to control the KD.

\section{Experiments}

\subsection{Experimental Setup}
\label{sec:exp_setup}

\begin{table}[t]
    \caption{Federated SS splits employed in our work. %
    }
    \setlength{\tabcolsep}{2.5pt}
    \label{tab:setup}
    \centering
    \footnotesize
    \begin{tabular}{lcccccc}
    \toprule
        \textbf{Split} & $Q$ & \textbf{$|\mathcal{D}^T|$} & \textbf{$|\mathcal{D}_{test}^{T}|$} & $|\mathcal{K}|$ &  {\textbf{\# Img/Client} (range)} \\
        \midrule
        Cityscapes & 19 & 2975 & 500 & 144 & $[10, 45]$ \\ 
        CrossCity &  13 & 12800 & 400 & 476 &  $[17, 37]$\\
        Mapillary & 19 & 17969 & 2000 & 357 & $[16, 100]$\\
        \hdashline
        CrossCity (split of \cite{main_ref}) & 13 & 12800 & 400 & 4 & 3200\\
         
    \bottomrule
    \end{tabular}
    \vspace{-5pt}
\end{table}

\noindent
\textbf{Datasets.}
We evaluate the proposed framework in  synthetic-to-real experimental setups for autonomous driving applications, which are commonly used as benchmark for domain adaptation methods.
For the source domain, we opt for the synthetic GTA5 dataset \cite{gta5}.
It comprises 24966 highly realistic road scenes of typical US-like urban and suburban environments.
As for the real (\ie, target) domain, we experiment with three different datasets: Cityscapes \cite{cityscapes}, CrossCity \cite{crosscity} and Mapillary Vistas \cite{mapillary}. 
We use unlabeled training samples from all the datasets. Results are reported on the original validation split for Cityscapes and Mapillary, while on the test split for CrossCity.

Cityscapes provides street-view images from 50 cities in Central Europe.
CrossCity includes more diverse locations and appearances, collecting driving scenes from multiple cities around the world (\ie,  Rome, Rio, Tokyo, and Taipei). 
Finally, the Mapillary Vistas dataset collects geo-localized street-view images from all around the world.
We consider the largest number of overlapping classes among GTA5 and the real datasets (\ie, 19 for Cityscapes and Mapillary, and 13 for CrossCity).
\\
We propose a federated partitioning of the target datasets among the clients for each of the target dataset (\ie, a \textit{split}), as summarized in Table~\ref{tab:setup} and detailed in Suppl.\ Mat. For Cityscapes, we use the \textit{heterogeneous} split from \cite{FedDrive}, where 144 clients observe
images taken only from one city. We emulated the same kind of split also for the CrossCity dataset. %
Finally, we used the GPS information of the Mapillary dataset to discover clients with spatially near images.

\noindent
\textbf{Baselines and Competitors. }
To support the efficacy of the proposed approach in the unexplored \setting setup, we compare with multiple methods on both centralized and federated setups.
As the lower bound, we consider the na\"ive \textit{source only} approach, which entails the sole use of source labeled data during training.
At the other end, as the upper bound we propose the FTDA and Oracle comparisons. Both the methods assume the availability of supervised target data, either on the server side (\ie, centralized framework) or on the client side (\ie, federated framework). However, the FTDA method implies that target data is used to fine-tune the model after a source-only pre-training, while the Oracle simply consists in a supervised FedAvg training on the labeled version of the target dataset.
Moreover, we re-implemented the Maximum Classifier Discrepancy (MCD) \cite{MCD} method, and we adapted it to the federated setting as the authors of \cite{main_ref} did. Furthermore, we compare our method with  DAFormer \cite{daformer}, that we regard as the current state-of-the-art UDA approach. %
We remark that both  baselines are evaluated in the UDA setting, \ie, a simpler scenario where source and target datasets are jointly available.
Concerning the \fl aggregation, \cite{qinbin, caldarola2022improving} show that algorithms like FedProx \cite{fedprox} and SCAFFOLD \cite{scaffold} typically do not provide an improvement for vision tasks, therefore we focused on other aspects of the training, like domain adaptation, clustering and style transfer techniques.

\noindent
\textbf{Server Pre-train.} We pre-trained the model on GTA5 using a power-law decreasing learning rate $\eta$, starting from ${\eta = 5.0 \cdot 10^{-3}}$ with power $0.9$ and using SGD optimizer with momentum equal to $0.9$ and no weight decay. We pre-train the model for 15k steps.
Each client computes the style on all its images 
using a window of size $3\times3$ and sends the mean style to the server, before the pre-training starts.

\newpage
\noindent
\textbf{Federated Adaption.} In CrossCity, we run the experiments with fixed $\eta = 1.0 \cdot 10^{-2}$, training on 4 clients per round for a number of rounds $T=1000$, with $\lambda_{\textsc{KD}}=20$; we update the pseudo-label teacher model every round ($\omega=1$) and set $t_\textsc{start}=400$ for SWAt. In Cityscapes, we trained on 5 client per round, with $T=300$, fixed $\eta=5e-5$, $\lambda_{KD}=10$, $\omega = 5$ and $t_\textsc{start}=200$. For Mapillary we used $\eta = 1.0 \cdot 10^{-2}$, $\lambda_{KD}=10$, 6 clients per round, with $T=100$, same pseudo-label policy of Cityscapes and $t_\textsc{start} = 50$.
In both settings, for all datasets, the batch size was 16. We performed data augmentation as follows: random scaling ($0.7, 2$), random crop of $1024\times512$, color jitter with brightness, contrast and saturation equal to 0.5, and image normalization. For Mapillary instead of random scaling we forced a fixed rescaling with width equal to 1024.

\subsection{Experimental Results}

\noindent
\textbf{GTA5$\rightarrow$Cityscapes}
Our first setup is the GTA5 $\rightarrow$ Cityscapes adaptation.
Experimental results are reported in Table \ref{tab:city_split_by_city}.
Even though providing high quality realistic images, the GTA5 dataset still suffers a domain gap compared to real-world images, as those included in Cityscapes.
We notice that simply training over supervised source data (\ie, \textit{source only}) leads to a significant performance discrepancy compared to the full target supervision.
Even applying the state-of-the-art DAFormer method \cite{daformer} in an UDA setting (\ie, assuming joint availability of source supervised and aggregated target unsupervised data, which violates the assumptions of our setup), there is a noticeable performance drop of around $25\%$ of mIoU from the supervised oracle.

In our setup, we assume a federated learning framework with private target data distributed among multiple clients and a large-scale source dataset only available in a central server for the pre-training stage only. 
This introduces additional challenges not present in standard centralized domain adaptation settings. 
In particular, we assume that source and target data are not accessible on the same device, and that target data is available in small batches scattered among devices and not in a single place.
Furthermore, target data is heterogeneously distributed among clients.
The increase in the task complexity is noticeable from the performance drop of the supervised target oracle and FTDA methods (which still assume target supervision).
This is also true for the MCD \cite{MCD} UDA approach, which loses almost $10\%$ of mIoU when tested in a federated setting.

The proposed method is able to obtain robust results in this challenging setting achieving a mIoU of around $36.5 \%$.
In particular, the efficient pre-training based on domain stylization, along with the self-training optimization scheme, allows to tackle the lack of source data at the client side. 
We additionally improve training stability, which is hindered by the small amount of target data available within single clients, with KD and SWAt, as shown by the very small standard deviation of the results in Table \ref{tab:city_split_by_city}.
Finally, we provide an enhanced aggregation mechanism, which indirectly shares task information in a effective manner among clients sharing similar input statistics (\ie, with smaller domain gap), according to our style-based client clustering.
By observing results in Table~\ref{tab:city_split_by_city}, we notice that LADD keeps a similar performance gap w.r.t.\ the target oracle compared to what DAFormer achieves in a centralized UDA setting. 
Competitive results are provided by different variations of the proposed style-based clustering, \ie, by keeping only the decoder (\ie, \textit{LADD (cls)}) or the whole network (\ie, \textit{LADD (all)}) as cluster-specific during the aggregation. Additional analyses are provided in Sec.\ \ref{sec:ablation}.

\begin{table}[t]
    \caption{Results on the heterogeneous split of Cityscapes. %
    }
    \setlength{\tabcolsep}{3.5pt}
    \label{tab:city_split_by_city}
    \centering
    \footnotesize
    \begin{tabular}{llcc}
    \toprule
        \textbf{Setting} & \textbf{Method} & \textbf{mIoU (\%)} \\
        \midrule
          centralized & Oracle & $66.64 \pm 0.33$ \\
          centralized & Source Only & $24.05 \pm 1.14$ \\ %
          centralized & FTDA & $65.74 \pm 0.48$ \\ %
         \hdashline centralized & MCD \cite{MCD} & $20.55 \pm 2.66$ \\ %
          centralized & DAFormer \cite{daformer} & $42.31 \pm 0.20$ \\
         \hdashline federated & Oracle & $58.16 \pm 1.02$ \\
          federated & FTDA &  $59.35 \pm 0.61$ \\
             \hdashline
          FL-UDA & MCD \cite{MCD} &  $10.86 \pm 0.67$ \\ %
          \textbf{\setting} & FedAvg$^\dag$  \cite{fedavg}  + Self-Tr. & $35.10 \pm 0.73$ \\
          \textbf{\setting} & \textbf{\method (cls)} & $\textbf{36.49} \pm \textbf{0.13}$\\
          \textbf{\setting} & \textbf{\method (all)} & $\textbf{36.49} \pm \textbf{0.14}$\\
    \bottomrule
    \end{tabular}
    \vspace{-5pt}
\end{table}

\noindent
\textbf{GTA5$\rightarrow$CrossCity}
We further investigate the performance of the proposed approach in the GTA5$\rightarrow$CrossCity scenario.
Quantitative results are reported in Table \ref{tab:crosscity_split_27_10}. 
We compare with the na\"ive source only baseline, as well as with MCD \cite{MCD}.
Due to the lack of target supervision on training images, the upper bound of the target oracle cannot be provided, nor the result of FTDA.

The diverse content and appearance of CrossCity's road scenes, due to variable geographic origin of its samples, provide a heterogeneous target distribution.
The enhanced heterogeneity w.r.t.\ the more uniform Cityscapes dataset in turn leads to a tougher challenge for federated training.
For instance, the MCD method when extended from a centralized to a federated learning framework suffers from a substantial performance reduction.
Instead, \method provides a much higher accuracy in a federated setting, with more than $17\%$ gain over federated MCD, while also not requiring reuse of source data after the initial pre-training. 
This is indicative of the robustness of our method w.r.t.\ the statistical diversity of client target data.

Finally, we note that, by allowing only a minimal amount of network parameters to be cluster-dependent, we achieve a final accuracy very close to our best result, obtained without any parameter sharing across clusters of clients. 
This result shows that \method demands limited communication overhead w.r.t.\  standard FedAvg.

\begin{table}[t]
    \caption{Results on the proposed CrossCity split. 
    }
    \label{tab:crosscity_split_27_10}
    \setlength{\tabcolsep}{4pt}
    \centering
    \centering
    \footnotesize
    \begin{tabular}{llcc}
    \toprule
        \textbf{Setting} & \textbf{Method} & \textbf{mIoU (\%)} \\ %
        \midrule
           centralized & Source Only &  $26.49 \pm 1.46$ \\  %
           centralized & MCD \cite{MCD} &  $27.15 \pm 0.87$ \\ %
             \hdashline
            FL-UDA & MCD \cite{MCD} &   $24.80 \pm 1.56$ \\ %
            \textbf{\setting}  & FedAvg$^\dag$  \cite{fedavg} + Self-Tr. & $33.59 \pm 1.25$ \\
         \textbf{\setting} & \textbf{\method (cls)} & $39.87 \pm 0.14$ \\
          \textbf{\setting} & \textbf{\method (all)} & $\textbf{40.09} \pm \textbf{0.19}$ \\
    \bottomrule
    \end{tabular}
    \vspace{-5pt}
\end{table}

\blfootnote{\dag: Same pretrain as LADD.}

\vspace{-10pt}
\noindent
\textbf{GTA5$\rightarrow$Mapillary}
Finally, we provide an experimental analysis with target data from Mapillary Vistas, while the GTA5 still serves as source dataset. 
Table \ref{tab:mapillary} contains numerical results of the evaluation.
The diverse assortment of target data collected around the world, and dispersed among clients according to geographic location (see Sec.\ \ref{sec:exp_setup}), makes client data distribution even more heterogeneous than the previous setups. 
We notice that with target supervision (\ie, oracle method), there exists a significant performance drop of around $11.5\%$ of mIoU from centralized to federated settings. This is even more noticeable with the MCD \cite{MCD} UDA approach, which struggles when tested under the considered federated learning setup, suffering from a similar mIoU decrease of $12\%$. 
The proposed \method framework instead provides considerable performance, with results surpassing the source only optimization by a large margin (more than $8\%$ of mIoU) in its best version with only classifier weights kept cluster-specific (\textit{LADD (cls)}), and approaching the federated oracle result. 
We further observe that LADD outperforms (in its best configuration) the simpler framework based on FedAvg and self-training. 
This supports the effectiveness of the proposed additional modules (concerning local training regularization and federated aggregation, see Sec.\ \ref{sec:style_aggr} and \ref{sec:adaptation}) in tackling the domain adaptation problem in a distributed learning setting.
Finally, we remark that the 
lightweight
aggregation scheme with a cluster-specific classifier achieves the best results.

\begin{table}[t]
    \caption{Results on the proposed Mapillary split.
    }
    \label{tab:mapillary}
    \centering
    \footnotesize
    \begin{tabular}{llcc}
    \toprule
        \textbf{Setting} & \textbf{Method} & \textbf{mIoU\%}\\% & \textbf{std} \\
        \midrule
          centralized & Oracle & $61.46 \pm 0.21$ \\
          centralized & Source Only & $32.40 \pm 0.71$ \\
          centralized & MCD & $31.93 \pm 1.89$  \\
            \hdashline 
          federated & Oracle & $49.91 \pm 0.49$ \\
          FL-UDA & MCD &  $19.15 \pm 0.75$\\
             \textbf{\setting} &  FedAvg$^\dag$  \cite{fedavg}+ Self-Tr. & $38.97 \pm 0.21$ \\
          \textbf{\setting} & \textbf{\method (cls)} & $\textbf{40.16} \pm  \textbf{1.02}$ \\
          \textbf{\setting} & \textbf{\method (all)} & $38.78 \pm 1.82$ \\
    \bottomrule
    \end{tabular}
     \vspace{-5pt}
\end{table}

\subsection{Ablation Studies} 
\label{sec:ablation}
\begin{table}[t]
    \caption{Ablation of our optimization framework, performed on CrossCity dataset for the presented split.}
    \label{tab:ablation_crosscity_split_27_10_num}
    \centering
    \begin{adjustbox}{width=\columnwidth}
    \centering
    \footnotesize
    \begin{tabular}{ccccccc}
    \toprule
         \textbf{FDA}  & \textbf{ST}  & \textbf{KD}  &  \textbf{SWAt} & \textbf{Cluster Aggr}  & \textbf{mIoU (\%)}\\% & \textbf{std}\\
        \midrule
           & & & &  & $26.49 \pm  1.46$ \\
            & \textbf{\cmark} &  & & & $30.58 \pm 0.59$ \\
             \cmark &  & & &  & $32.43 \pm 0.61$ \\
               & \textbf{\cmark} & \textbf{\cmark} & \textbf{\cmark} & \textbf{\cmark} & $ 32.78 \pm 0.09 $ \\
             \cmark & \cmark & & &  & $33.59 \pm 1.25$\\
             \cmark & \cmark & \cmark & & & $37.49 \pm 0.14$\\
             \cmark & \cmark & \cmark & \cmark & & $38.83 \pm 0.12$ \\
             \cmark & \cmark & \cmark  & & \cmark  & $39.18 \pm 0.24$ \\
             \textbf{\cmark} & \textbf{\cmark} & \textbf{\cmark} & \textbf{\cmark} & \textbf{\cmark} & $\textbf{40.09} \pm  \textbf{0.19}$ \\
    \bottomrule
    \end{tabular}
    \end{adjustbox}
    \vspace{-5pt}
\end{table}
\noindent
\textbf{Impact of Optimization Modules.} 
We now study the contribution of each component of our method.
In Table ~\ref{tab:ablation_crosscity_split_27_10_num}, we report the target mIoU computed with modules incrementally activated, showing the gain brought by each of them (GTA5$\rightarrow$CrossCity setup). 
We notice that the introduction of the style transfer technique during server pre-training generates an improvement of almost $6\%$ of mIoU, where the lower bound (source only pre-training) coincides with the centralized source only experiment. The na\"ive federated adaptation using the proposed self-training routine (and FedAvg aggregation) allows to gain an initial improvement of $2\%$ of mIoU, but leads to unstable training curves (see Suppl.\ Mat.). 
Adding the KD module we further boost the performance (by almost $3\%$ mIoU) 
and prevent clients' optimization from undertaking unsteady behaviors, with the initial stable configuration as anchor point. 
By activating SWAt we get an extra boost in terms of final mIoU and training stability, and training convergence is achieved much more consistently. 
We remark how the std drops when enabling KD and SWAt.
When further introducing cluster aggregation, but leaving SWAt disabled, we get a small mIoU increase, at the price of higher instability. 
Finally, by adding the cluster aggregation along with SWAt we get our complete method, for a final score of $40.09\%$ of mIoU, achieved with stable training. 

\noindent
\textbf{Style-based Pre-training.}
We analyze the impact of the stylization mechanism on the pre-training.
We test different style extraction schemes, by varying the size of the Fourier amplitude window.
Table \ref{tab:ablation_crosscity_split_27_10_window_pretrain} reports quantitative results of the ablation study (GTA5$\rightarrow$CrossCity setup).
It is possible to observe that a window of 1x1 is sufficient to capture and transfer useful domain-dependent information across domains.
However, by increasing the dimension of the style window to $3\times3$ and $5\times5$ pixels we get an improvement of $1\%$ mIoU.
In addition, even though providing similar results to the $5 \times 5$ size in terms of final mIoU, the $3\times3$ window leads to a more stable pre-training (testified by lower std), due to less artifacts being introduced in the stylized images. 

We remark that the style data occupies a very limited amount of memory (in the order of a few bytes), and thus requires little communication overhead to be transmitted from client to server before federated rounds start.
The style window corresponds to a small portion of the Fourier Transform and, prior to its transmission, is averaged over all data samples within each client.
Recall also that  shape information is mapped to the phase data, while we transmit only amplitude information. Therefore, we argue that it encloses a negligible fraction of the overall information of the local image data, and thus does not violate data privacy limitations.

\begin{figure}
    \centering
    \includegraphics[width=0.85\columnwidth,trim=0 0.2cm 0 1.5cm,clip]{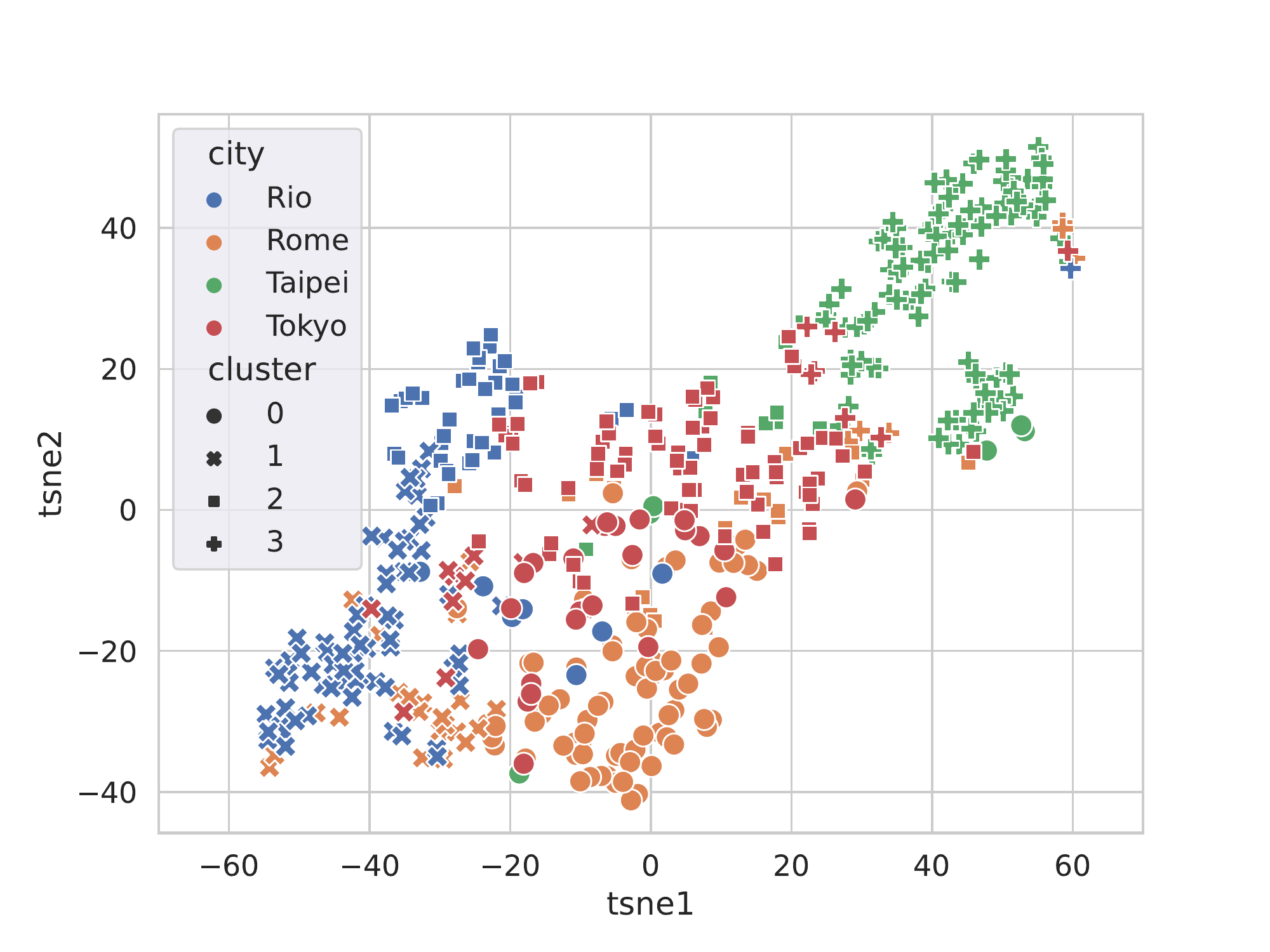}
    \caption{t-SNE of the styles. The colors represent the city ground-truth, while the symbols represent the inferred clusters. The clustering accuracy, considering each cluster a city, is equal to 0.68.}
    \label{fig:tsne_crosscity_27_10}
    \vspace{-5pt}
\end{figure}

\noindent
\textbf{Cluster-level Aggregation.}
We study how different cluster-based aggregation schemes affect the overall adaptation performance.
In particular, we select different groups of model parameters for cluster-specific and global aggregations. 
Results obtained in the GTA5$\rightarrow$CrossCity setup are collected in Table~\ref{tab:ablation_crosscity_split_27_10}.
We notice that keeping only batch-norm parameters cluster-dependent (second row) provides similar results as the standard FedAvg (first row), where all model parameters are global.
When holding per-cluster backbone and classifier blocks individually, we get improved  performance.
The accuracy is further slightly boosted when the entire  model is kept cluster-specific, showing that backbone and classifier both enclose cluster-dependent information.

Finally, we observe that solely treating the lightweight classifier block as cluster-specific gives comparable performance to the best full-model intra-cluster aggregation.
Therefore, we found a less computational and memory demanding version of the proposed \method approach, still providing robust performance.

\noindent
\textbf{Style-based Client Clustering.}
We analyze the distribution of clients across the style-based clusters identified by our approach in an unsupervised fashion.
The study is conducted when the CrossCity target dataset is employed, so that we can compare style-based clusters with those determined by the city of origin.
In Fig.~\ref{fig:tsne_crosscity_27_10} we associate each cluster with a point in a 2D space according to its style tensor.
In particular, each style tensor is flattened resulting into a 27-D vector, and the t-SNE \cite{van2008visualizing} dimensional reduction method is used to project it into a 2D space. We can observe that clients from different cities (each of which identified by a different color) tend to be clustered together. 
At the same time, we notice that clients with similar styles (\ie, are associated to the same style-based cluster by our approach, see Sec.~\ref{sec:style_aggr}) are projected in adjacent regions. %
Furthermore, both city- and style-based partitions appear to be highly overlapping, signifying that style-based clustering is effectively able to capture domain-dependent information, which is highly correlated to the geographical location.

\noindent
\textbf{Comparison in a simpler CrossCity split.}
We compare with the FMTDA method \cite{main_ref} in the GTA5$\rightarrow$CrossCity adaptation setup they propose, with target data distributed over 4 total clients, each containing images of one of the 4 CrossCity's cities. For a fair comparison, we use the same segmentation network adopted in \cite{main_ref}. %
In this simpler federated setting, a restricted version of \method  without the cluster-level aggregation is still effective. In Table \ref{tab:ablation_crosscity_split_4}, we report the mIoU computed on each city, along with the average value, for different approaches directly taken from \cite{main_ref}.
We observe a consistent improvement, which in the average target mIoU reaches almost $5\%$, demonstrating the superiority of the proposed \method approach.

\begin{table}[t]
\begin{minipage}[t]{3.9cm}
    \caption{Ablation on window size used for FDA pre-training.}
    \label{tab:ablation_crosscity_split_27_10_window_pretrain}
    \centering
    \footnotesize
    \begin{tabular}{lcc}
    \toprule
        \textbf{Size}  &  \textbf{mIoU (\%)} \\%& \textbf{std}\\
        \midrule
            None & $26.49 \pm 1.46$ \\
            $1\times1$ & $31.59 \pm 0.68$ \\
            $3\times3$ & $32.43 \pm 0.61$ \\
            $5\times5$ & $32.51 \pm 0.75$ \\
    \bottomrule
    \end{tabular}
    \vspace{-5pt}
\end{minipage}
\begin{minipage}[t]{3.9cm}
    \caption{Ablation on cluster-specific layers (CrossCity).}
    \label{tab:ablation_crosscity_split_27_10}
    \centering
    \footnotesize
    \begin{tabular}{lcc}
    \toprule
        \textbf{ Layers}  &  \textbf{mIoU (\%)} \\%& \textbf{std} \\
        \midrule
            None & $38.83 \pm 0.12$\\
            BN & $38.72 \pm 0.20$ \\
            Backbone & $39.31 \pm 0.13$  \\
            Classifier & $39.87 \pm 0.14$ \\
            \textbf{All} & $\textbf{40.09} \pm \textbf{0.19}$ \\
    \bottomrule
    \end{tabular}
    \vspace{-5pt}
\end{minipage}
\end{table}

\begin{table}[t]
    \caption{Results on the CrossCity split proposed in \cite{main_ref}.}
    \label{tab:ablation_crosscity_split_4}
    \centering
    \setlength{\tabcolsep}{3pt}
    \footnotesize
    \begin{tabular}{lcccccc}
    \toprule
        \textbf{Method}  &  Rio & Rome & Taipei & Tokyo & \textbf{Avg}\\
        \midrule
            Source Only & 27.9 & 27.6 & 26.0 & 28.2 & 27.4  \\
            Cent-MCD \cite{MCD} & 31.3 & 30.6 & 28.8 & 31.6 & 30.5  \\
        \midrule
            Fed-DAN \cite{long2017deep} & 27.3 & 26.4 & 26.0 & 28.5 & 27.1\\
            Fed-DANN \cite{ganin2015unsupervised} & 28.6 & 26.0 & 26.6 & 28.6 & 27.5\\
            Fed-MCD \cite{MCD} &  27.7 &  27.3 &  26.5 &  29.0 &  27.6 \\
            DualAdapt \cite{main_ref} &  29.2 &  28.0  & 27.6  & 30.7  & 28.9 \\
            \method (ours) & \textbf{35.4} & \textbf{34.0} & \textbf{31.5} & \textbf{32.4} & \textbf{33.3}  \\
    \bottomrule
    \end{tabular}
    \vspace{-5pt}
\end{table}

\section{Conclusion}

In this work we introduced \setting, a new challenging and realistic setting for Source-Free Domain Adaptation in Federated Learning for Semantic Segmentation. In \setting, a server-side labeled dataset is used for pre-training the model, while local training uses only the unlabeled clients' data.
We introduced \method, an innovative algorithm to solve \setting, employing (i) style-transfer, knowledge distillation and SWA teacher on the pseudo-labels for regularizing learning, and (ii) style-driven clustering for learning both global and personalized parameters.
\method has no direct competitors due to the novel setup \setting, but is still able to achieve competing results compared to %
state-of-the-art algorithms. %
We also provided two new splits adapting the CrossCity and Mapillary Vistas datasets to the federated scenario as a reference for future research in \fl SS.

\newpage

{\small

\bibliographystyle{ieee_fullname}
\bibliography{strings, main}
}

\clearpage

\renewcommand{\thefigure}{S\arabic{figure}}
\renewcommand{\theHfigure}{S\arabic{figure}}

\renewcommand{\thesection}{S\arabic{section}}
\renewcommand{\theHsection}{S\arabic{section}}

\renewcommand{\theequation}{S\arabic{equation}}
\renewcommand{\theHequation}{S\arabic{equation}}

\renewcommand{\thetable}{S\arabic{table}}
\renewcommand{\theHtable}{S\arabic{table}}

\setcounter{equation}{0}
\setcounter{figure}{0}
\setcounter{table}{0}
\setcounter{section}{0}
\setcounter{page}{1}

\clearpage

\begin{strip}
{
\null
\vskip .375in
\begin{center}
    {\Large \bf 
    \textit{Supplementary Material}\\ 
    Learning Across Domains and Devices: \\Style-Driven Source-Free Domain Adaptation in Clustered Federated Learning
    \par}
      \vspace*{12pt}
  \end{center}
  }
\end{strip}

This document contains supporting material for the paper \textit{Learning Across Domains and Devices: Style-driven Source-Free Domain Adaptation in Clustered Federated Learning}. 
Here, we include additional details on the federated splits employed in the paper along with analyses of the convergence stability of our approach when compared to competing strategies adapted to our federated setup.
Finally, we show some qualitative segmentation maps.

\section{Additional Details on Splits}

In this section, we complete the description of how the federated splits used in our experiments are generated.

\begin{figure}[t]
    \newcolumntype{Y}{>{\centering\arraybackslash}X}
    \centering
    \includegraphics[width=\linewidth]{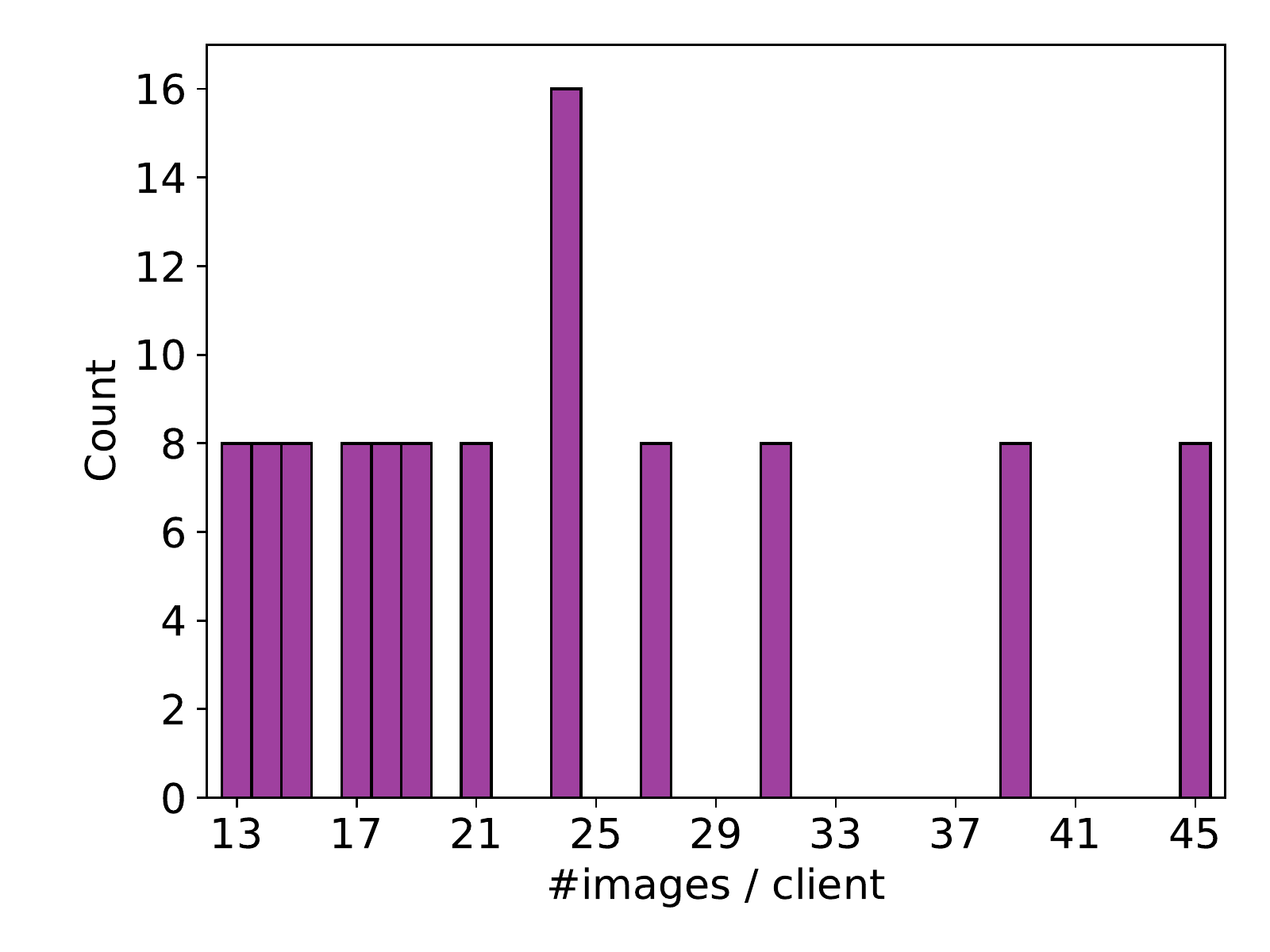}
    \caption{Histogram of images per client in the federated Cityscapes split.}
    \label{fig:clients_cityscapes}
\end{figure}

\textbf{Cityscapes.} We used the \textit{heterogeneous} federated split of Cityscapes \cite{cityscapes} proposed in \cite{FedDrive}. The split comprises $144$ clients, where each client has between $10$ and $45$ samples belonging to a single city from the dataset. Further details on the distribution of the number of images per client are shown in Figure~\ref{fig:clients_cityscapes}. 

\begin{figure*}
    \newcolumntype{Y}{>{\centering\arraybackslash}X}
    \centering
    \begin{subfigure}{\imgWidth}
        \caption{\scriptsize{Rio}}
        \includegraphics[width=\textwidth]{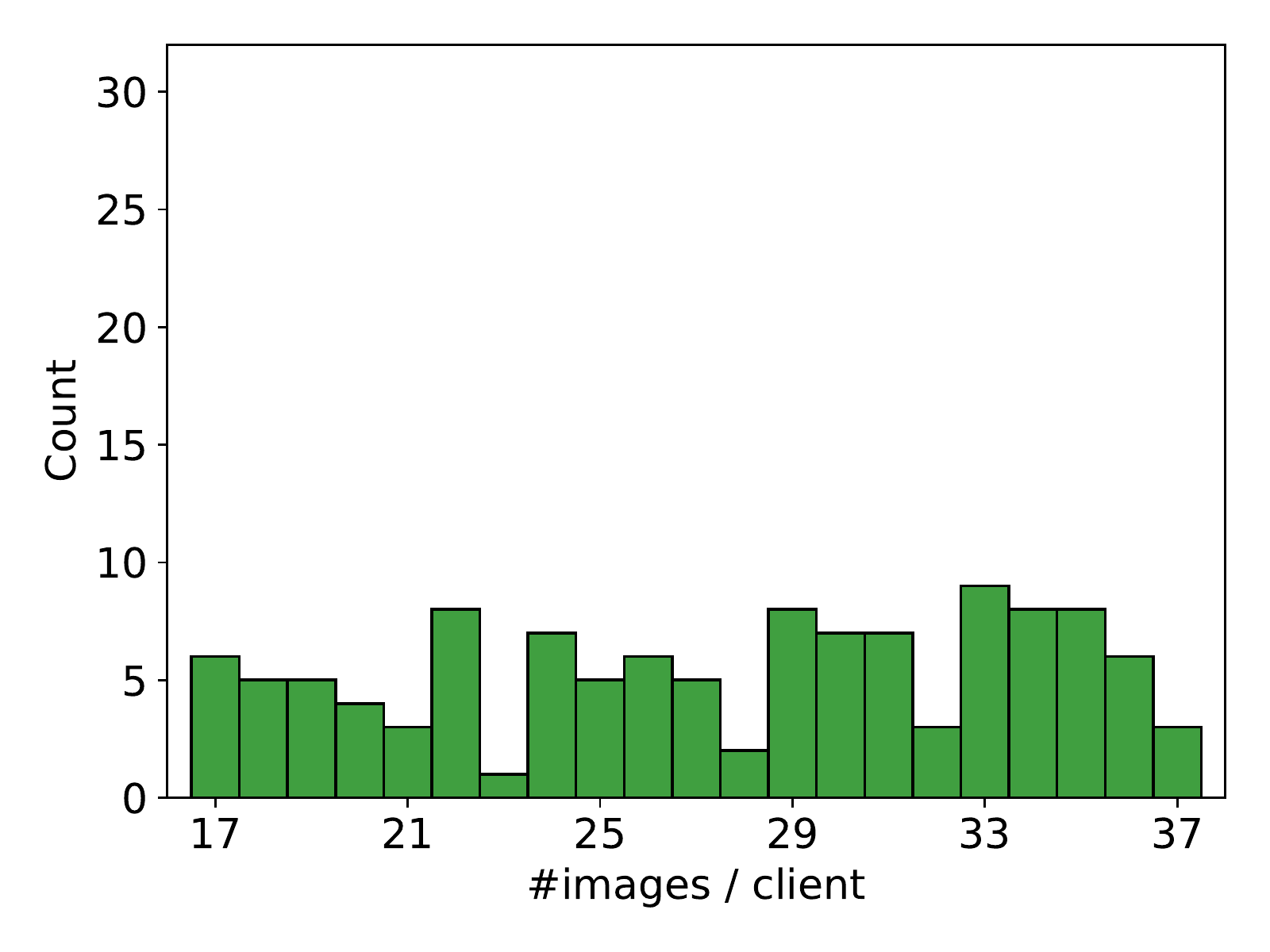}
    \end{subfigure}%
    \begin{subfigure}{\imgWidth}
        \caption{\scriptsize{Rome}}
        \includegraphics[width=\textwidth]{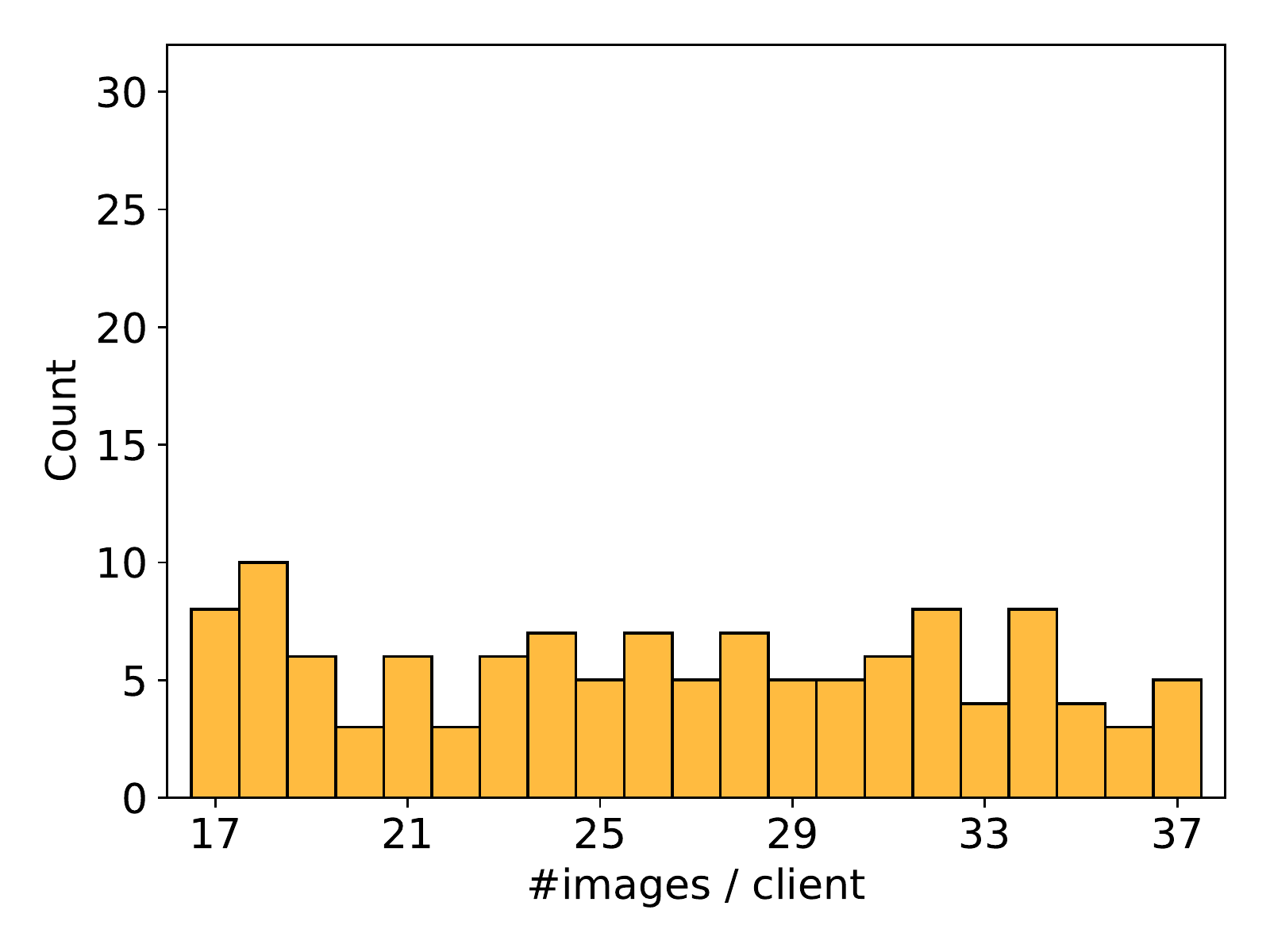}
    \end{subfigure}%
    \begin{subfigure}{\imgWidth}
        \caption{\scriptsize{Taipei}}
        \includegraphics[width=\textwidth]{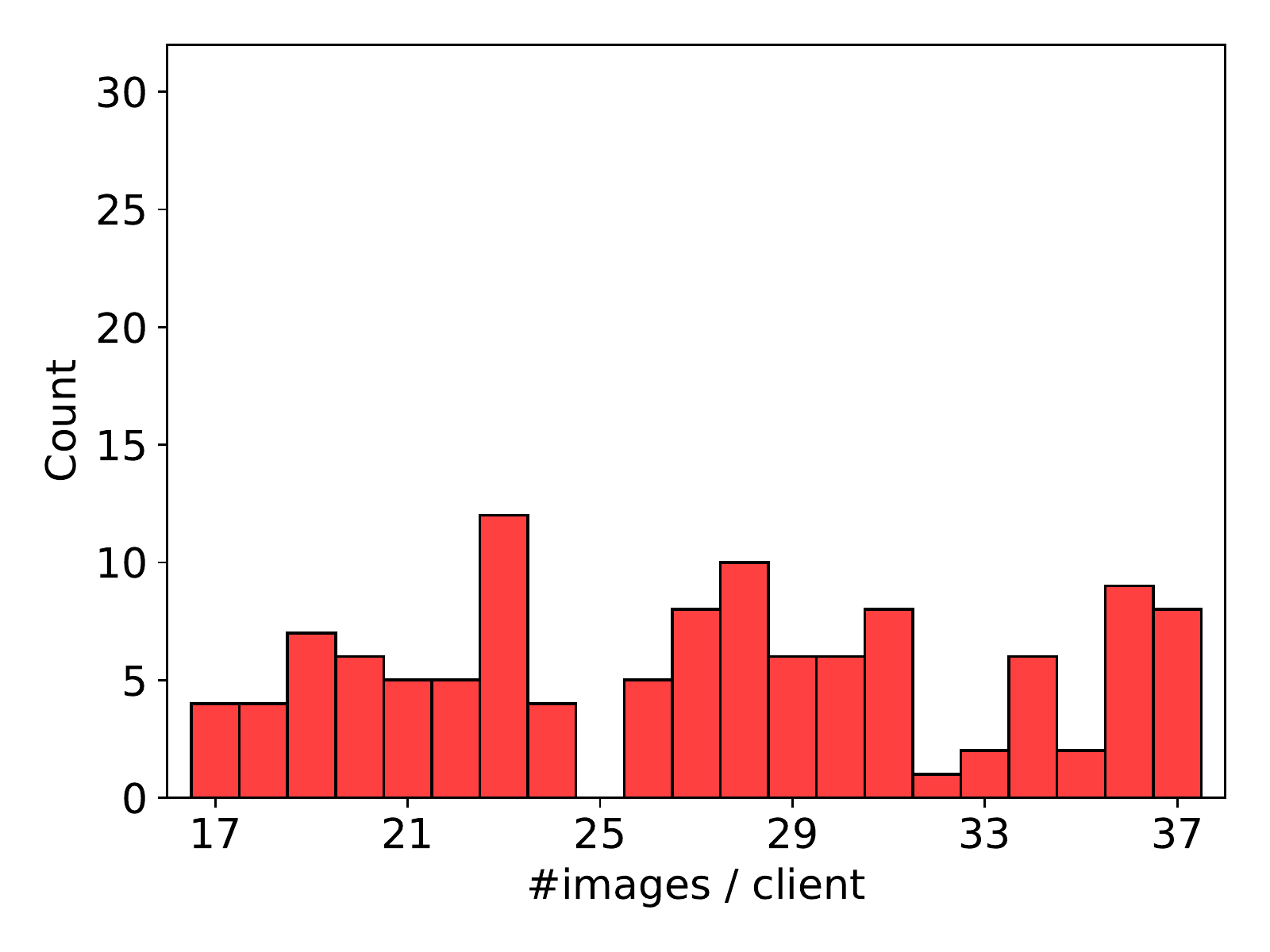}
    \end{subfigure}%
    \begin{subfigure}{\imgWidth}
        \caption{\scriptsize{Tokyo}}
        \includegraphics[width=\textwidth]{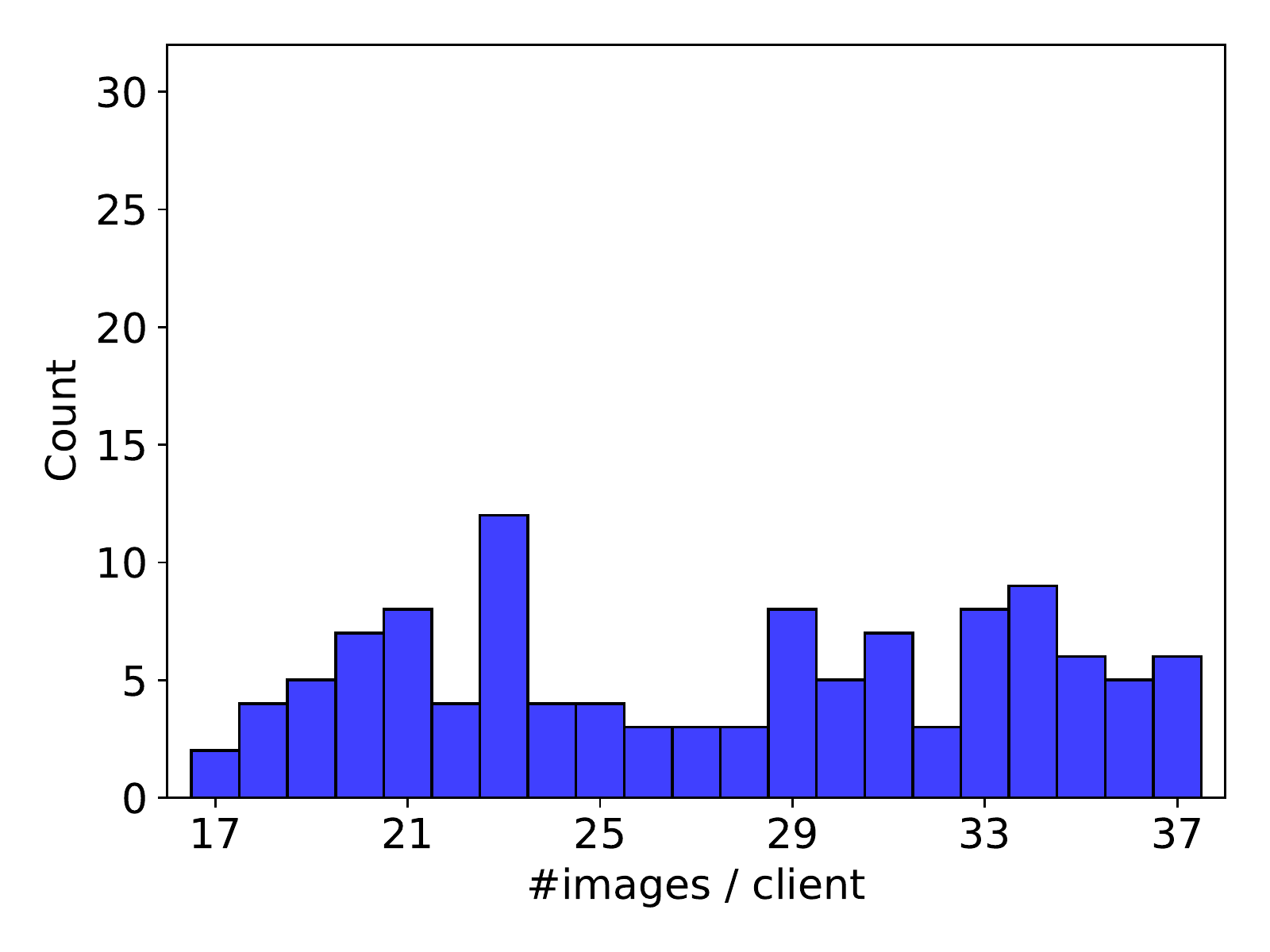}
    \end{subfigure}%
    \begin{subfigure}{\imgWidth}
        \caption{\scriptsize{Cumulative}}
        \includegraphics[width=\textwidth]{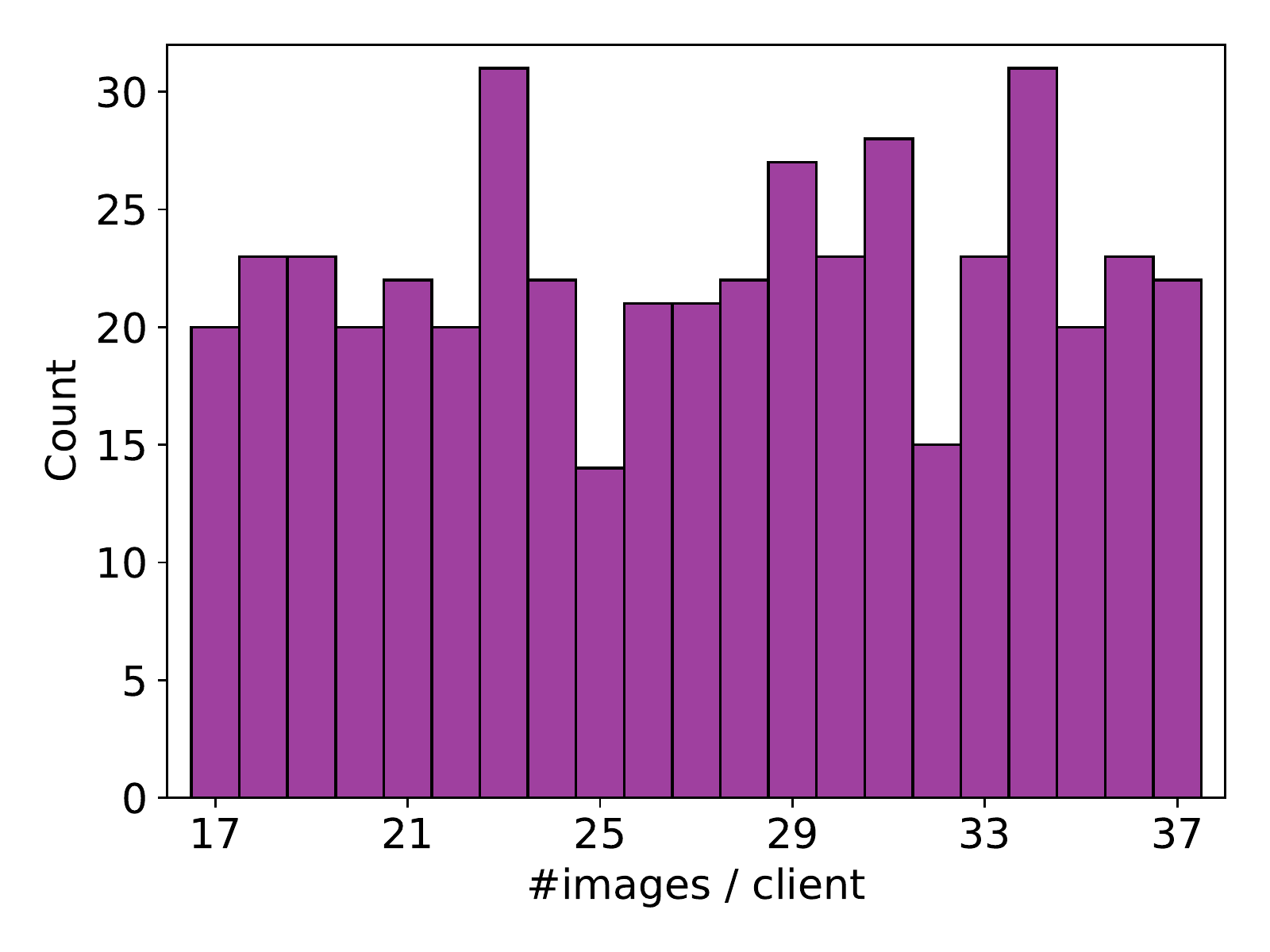}
    \end{subfigure}%
    \caption{Histogram of images per clients in the proposed federated CrossCity split.}
    \label{fig:clients_crosscity}
\end{figure*}

\textbf{CrossCity.} We generated the federated split of the CrossCity \cite{crosscity} dataset by assigning $27\pm 10$ images taken from the same city to each client, where the number of samples per client is uniformly sampled. The final distributions of the number of images per client are shown in Figure~\ref{fig:clients_crosscity} both per city and overall. We observe how the distributions are balanced across the four cities.

\begin{figure*}[ht]
    \newcolumntype{Y}{>{\centering\arraybackslash}X}
    \centering
    \begin{minipage}{.69\linewidth}
        \begin{minipage}{.32\linewidth}
        \begin{subfigure}{\linewidth}
        \caption{\scriptsize{Africa}}
        \includegraphics[width=\textwidth]{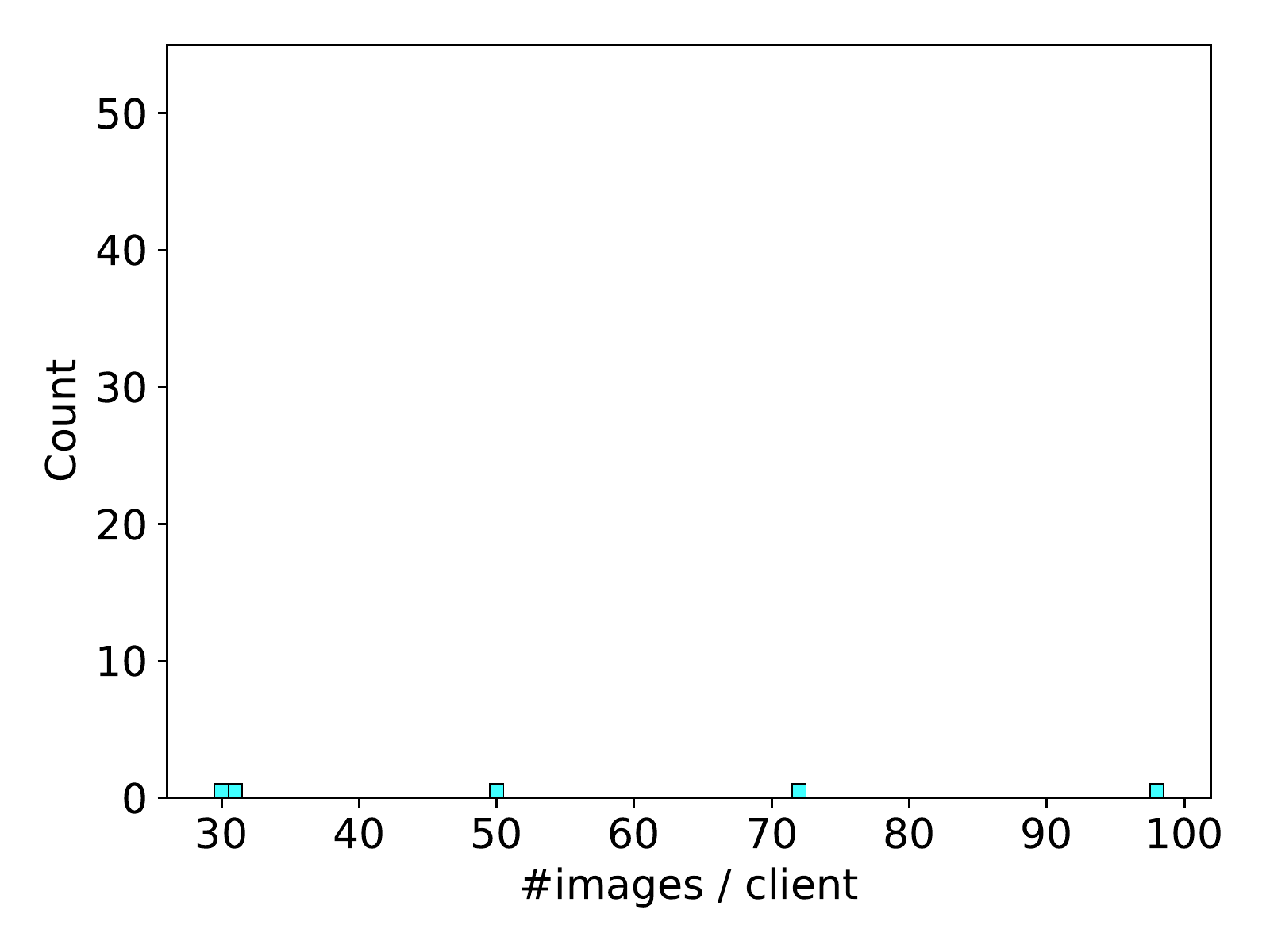}
        \end{subfigure}
        \end{minipage}
        \begin{minipage}{.32\linewidth}
        \begin{subfigure}{\linewidth}
        \caption{\scriptsize{Asia}}
        \includegraphics[width=\textwidth]{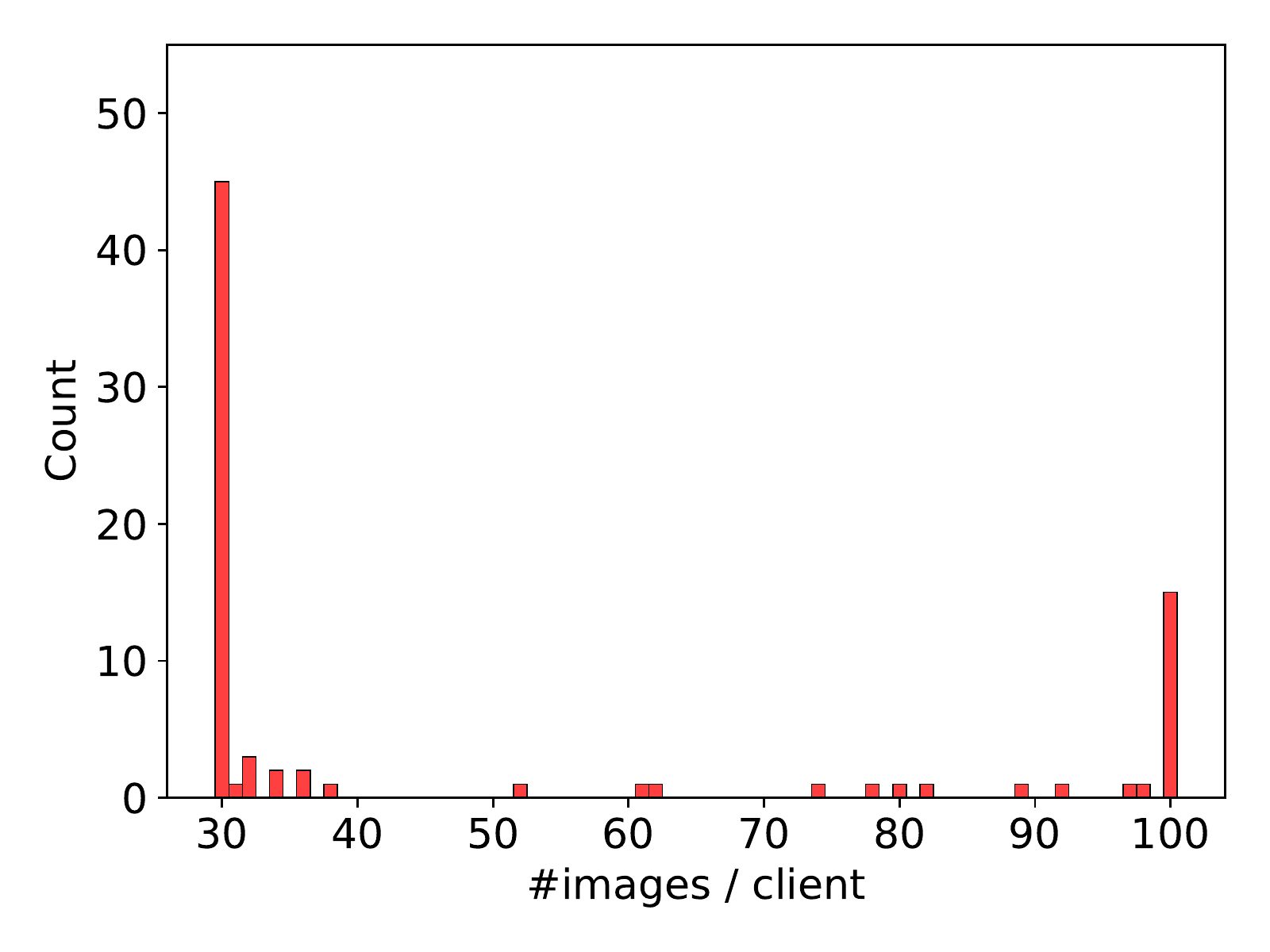}
        \end{subfigure}
        \end{minipage}
        \begin{minipage}{.32\linewidth}
        \begin{subfigure}{\linewidth}
        \caption{\scriptsize{Europe}}
        \includegraphics[width=\textwidth]{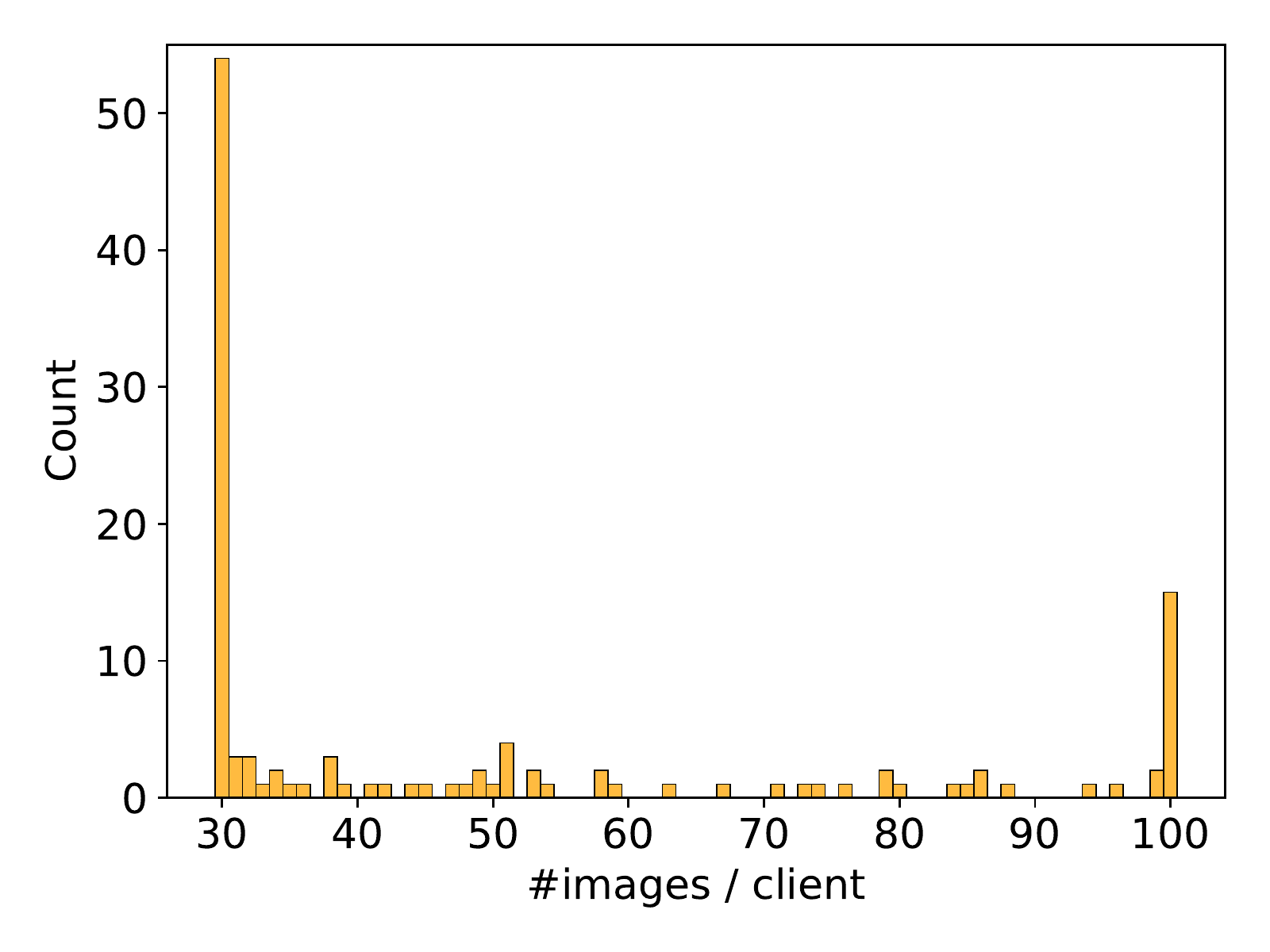}
        \end{subfigure}
        \end{minipage}
        \begin{minipage}{.33\linewidth}
        \begin{subfigure}{\linewidth}
        \caption{\scriptsize{North America}}
        \includegraphics[width=\textwidth]{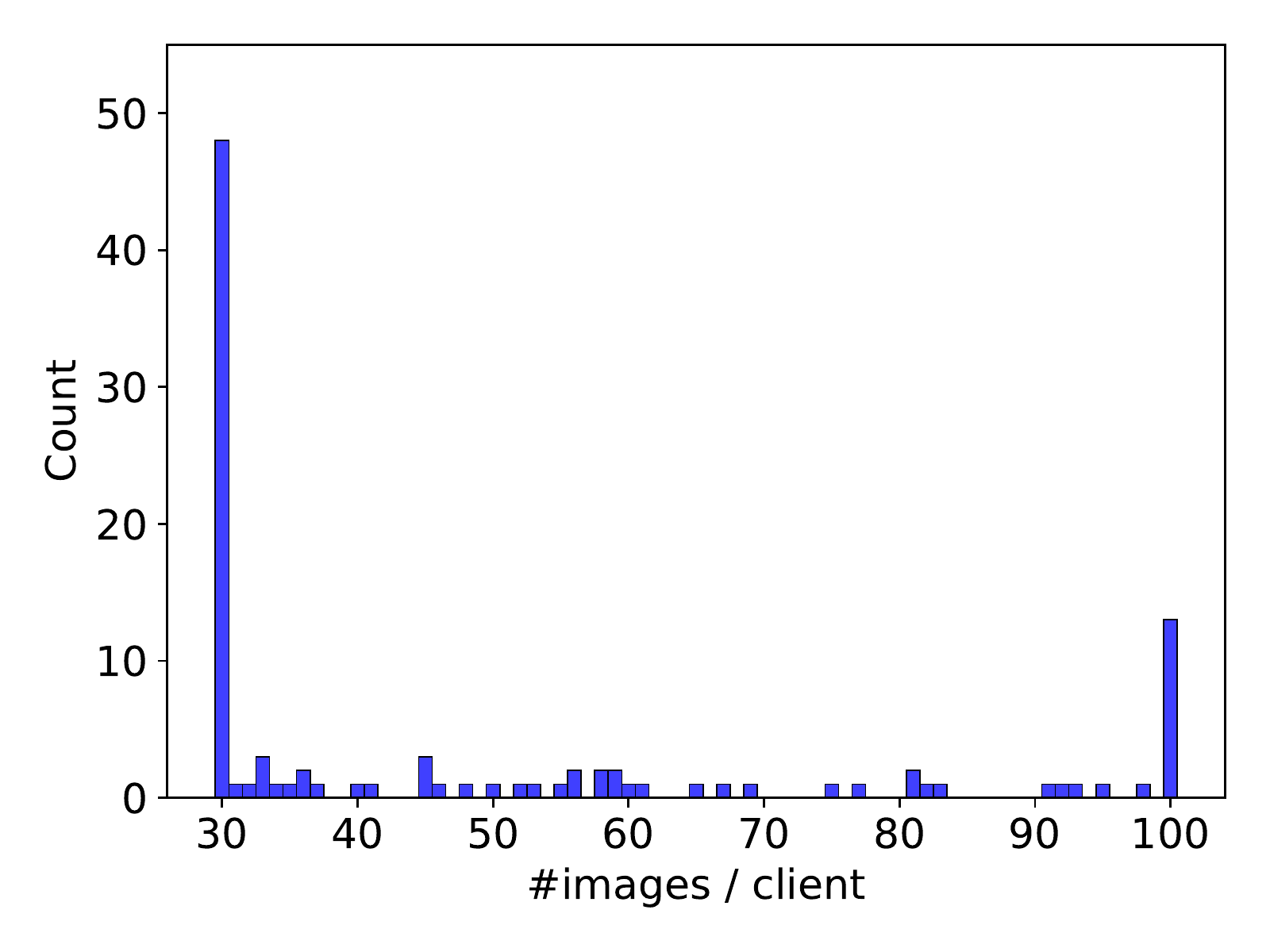}
        \end{subfigure}
        \end{minipage}
        \begin{minipage}{.33\linewidth}
        \begin{subfigure}{\linewidth}
        \caption{\scriptsize{Oceania}}
        \includegraphics[width=\textwidth]{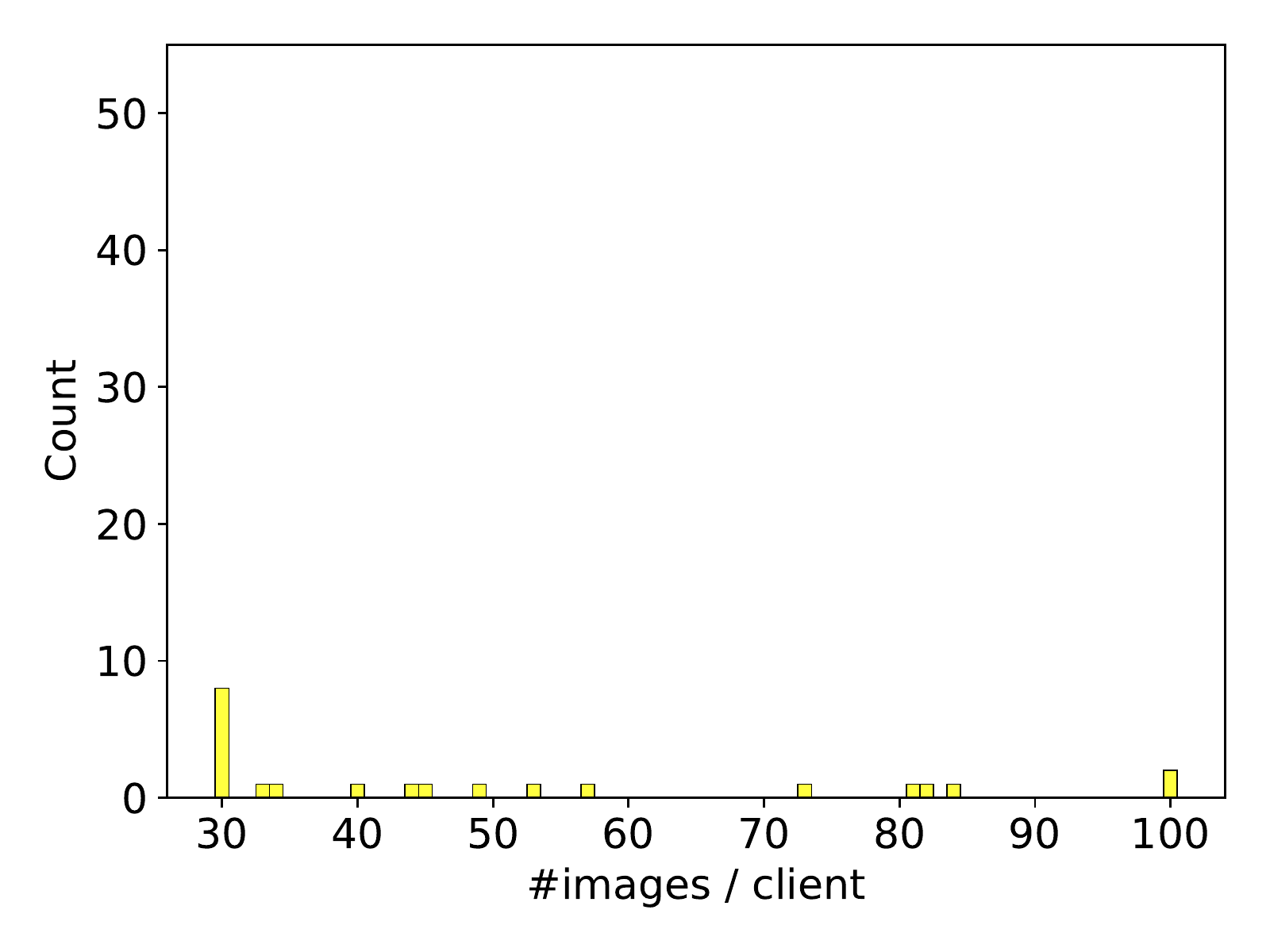}
        \end{subfigure}
        \end{minipage}
        \begin{minipage}{.32\linewidth}
        \begin{subfigure}{\linewidth}
        \caption{\scriptsize{South America}}
        \includegraphics[width=\textwidth]{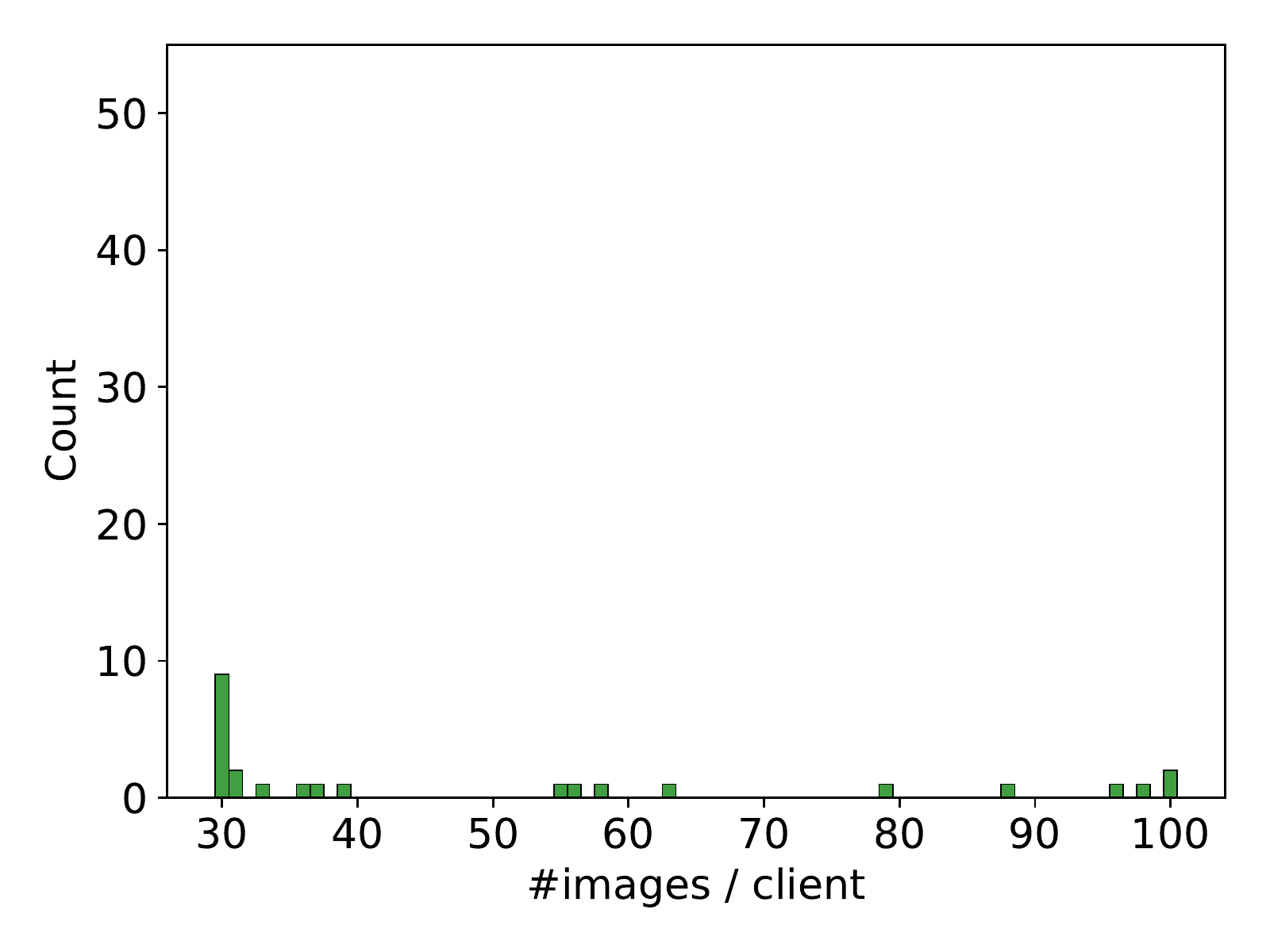}
        \end{subfigure}
        \end{minipage}
    \end{minipage}
    \begin{minipage}{.3\linewidth}
        \begin{subfigure}{\linewidth}
            \caption{\scriptsize{Cumulative}}
            \includegraphics[width=\textwidth]{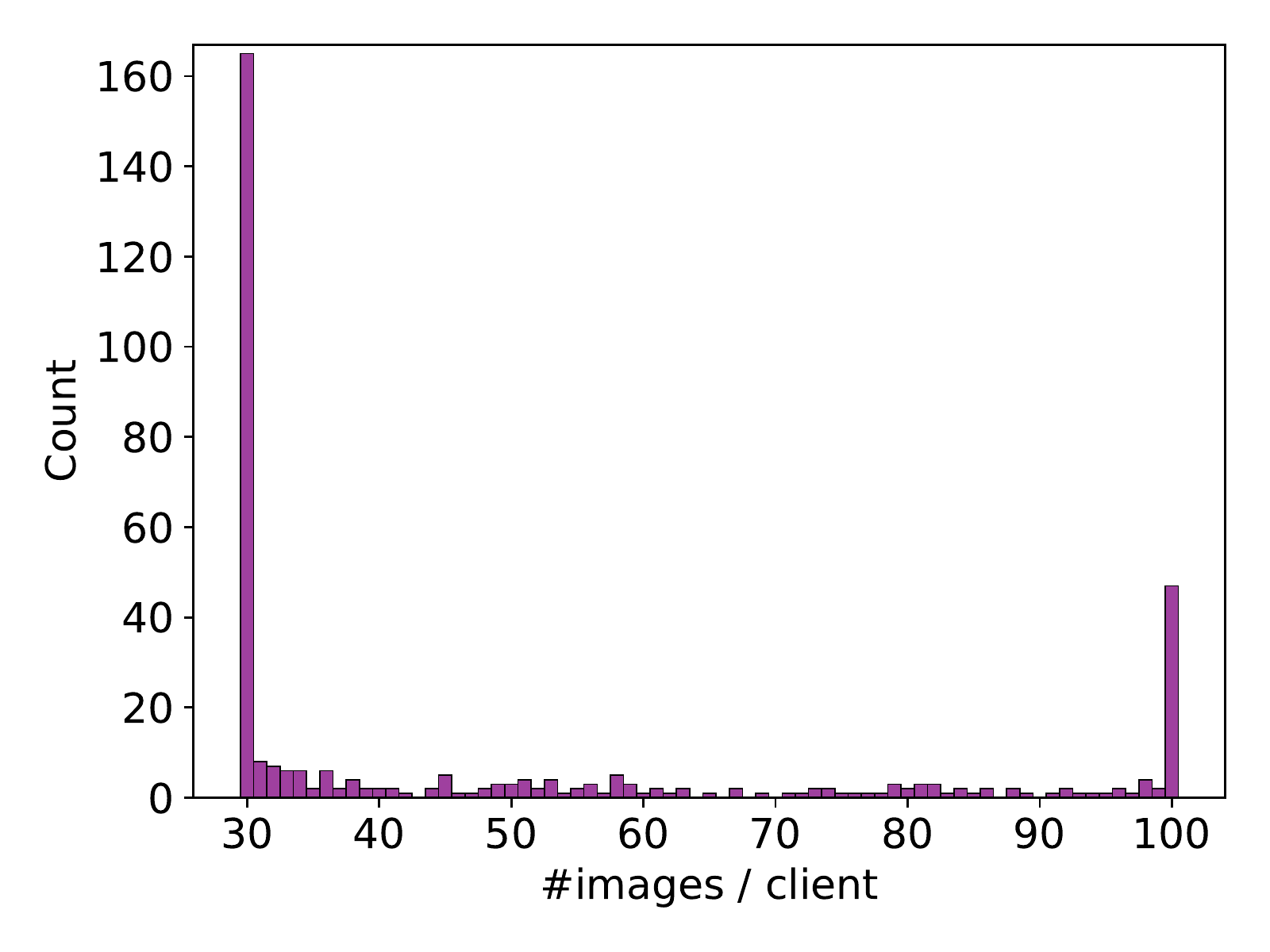}
        \end{subfigure}%
    \end{minipage}
    \caption{Histogram of images per clients in the proposed federated Mapillary Vistas split.}
    \label{fig:clients_mapillary}
\end{figure*}

\textbf{Mapillary.} We propose a novel split for the Mapillary Vistas \cite{mapillary} dataset via a clustering procedure based on the GPS coordinates of the images. We started from the original training set of $18000$ images and discarded $31$ of them missing the GPS coordinates. Then, we run the k-Means algorithm over the GPS coordinates six times, one per continent. The k-Means algorithm is constrained to assign every client a random number of images in the range $16$ and $100$.
The procedure resulted in $357$ clients, where each client observed samples from only one continent. 
The final distributions of the number of images per client are shown in Figure~\ref{fig:clients_mapillary}. Unlike the other scenarios, we observe a large variability across the distributions obtained in different continents due to the highly imbalanced nature of the dataset. Also, note that the two entries with higher values, $16$ and $100$, correspond to the extreme values of the constrained k-Means process.
\section{Additional Details on the Style-Based Client Clustering}

In a realistic FL setting, different clients may observe similar samples, \eg self-driving cars in the same region are likely to collect similar images, thus they are not subject to statistical heterogeneity during the server aggregation. 
Therefore, we proposed a style-driven client clustering as one of the foundational parts of our algorithm.
During the FL optimization stage, we employed the identified communities in a clustered and layer-aware aggregation policy on the server side.

First of all, we remark that the four clusters identified by the styles extracted from the images contain mostly clients belonging to one single geographical location (\ie, city). Table~\ref{tab:crosscity_clusters_bycity} shows the number of clients belonging to a specific city assigned to each cluster for the federated CrossCity dataset. Overall, the clustering accuracy, considering each cluster a city, is equal to 68\%. Therefore, there is not a one-to-one correspondence of the clusters with the cities.

\begin{table}[ht]
    \caption{Number of clients belonging to a specific city assigned to each cluster for the federated CrossCity split.}
    \setlength{\tabcolsep}{3.5pt}
    \label{tab:crosscity_clusters_bycity}
    \centering
    \footnotesize
        \begin{tabular}{rcccc}
        \toprule
        & \textbf{Cluster 1} & \textbf{Cluster 2} & \textbf{Cluster 3} & \textbf{Cluster 4} \\
        \midrule
        \textbf{Rio} & 7 & 1 & \textbf{70} & 38 \\
        \textbf{Rome} & \textbf{76} & 6 & 22 & 17 \\
        \textbf{Taipei} & 6 & \textbf{103} & 0 & 9 \\
        \textbf{Tokyo} & 26 & 8 & 10 & \textbf{73} \\
        \bottomrule
        \end{tabular}
    \vspace{-5pt}
\end{table}

\begin{figure*}[ht]
    \newcolumntype{Y}{>{\centering\arraybackslash}X}
    \centering
    \includegraphics[width=\textwidth]{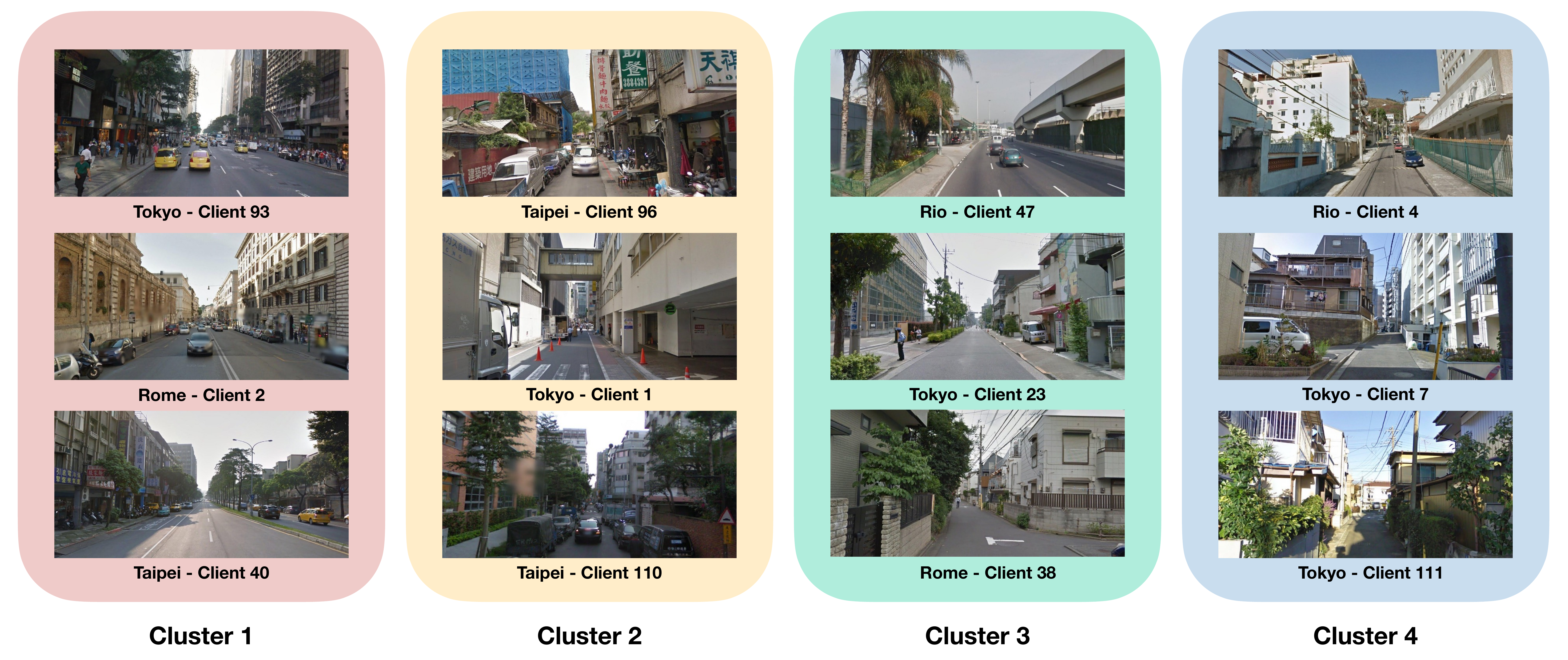}
    \caption{Sample images in each cluster for the federated CrossCity split.}
    \label{fig:crosscity_clusters}
\end{figure*}

\begin{figure*}
    \newcolumntype{Y}{>{\centering\arraybackslash}X}
    \centering
    \includegraphics[width=\textwidth]{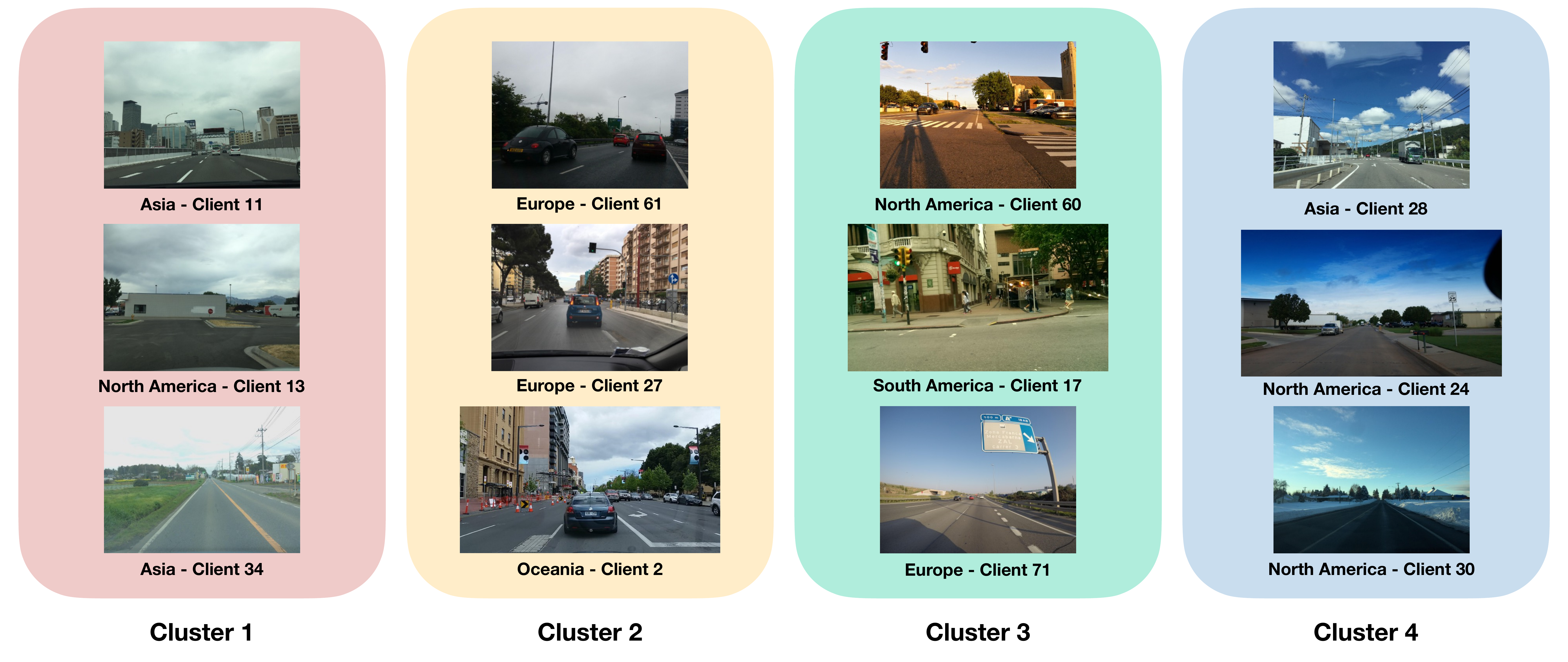}
    \caption{Sample images in some clusters for the federated Mapillary split.}
    \label{fig:mapillary_clusters}
\end{figure*}

To investigate this aspect, we show in Figure~\ref{fig:crosscity_clusters} some samples taken from the clients belonging to each of the four clusters in the federated CrossCity dataset. Here, we observe an interesting finding: despite being generated via style information only, the clusters tend to show scenes with similar semantics. For instance, \textit{Cluster 1} contains clients having images of large and trafficked streets, and grayish sky. \textit{Cluster 2} contains clients having images of narrow streets with little to no vegetation, many buildings, a few parked cars and whitish sky. \textit{Cluster 3} contains clients having images of empty roads with green surrounding vegetation. \textit{Cluster 4} contains clients having images from sunny weather and blue sky, narrow streets with no traffic and green vegetation.

Finally, we show in Figure~\ref{fig:mapillary_clusters} some samples taken from the clients belonging to each of four clusters in the federated Mapillary dataset. Unlike as for CrossCity, here we do not appreciate a clear assignment as the number of clusters is different from the number of towns or continents. Therefore, we observe that here the clustering is much more appearance-related, according to the style of the images.

For instance, \textit{Cluster 1} contains clients having cloudy and foggy images where the visual appearance is grayish.
\textit{Cluster 2} contains clients having grayish sky and yellowish buildings with some similar semantics across clients.
\textit{Cluster 3} contains clients having images at the sunset or sunrise where the light scatters yellow shadows. \textit{Cluster 4} contains clients having images with predominant blue colors in the sky.

\section{Implementation Details}
The proposed method is implemented in PyTorch, the code and federated splits are available at \url{https://github.com/Erosinho13/LADD}.

The semantic segmentation network used is DeepLab-V3 \cite{deeplabv3} with Mobilenet-V2 \cite{mobilenetv2} as the backbone and width multiplier equal to 1, representing a good compromise in terms of performance and lightness, important aspects to consider for real-world applications, such as self-driving cars. On each communication round, the selected clients are trained sequentially, allowing to perform the complete simulation and reproduce the results on a single GPU with 32GB of VRAM (we used a NVIDIA RTX 3090). 
 
\definecolor{road}{rgb}{.502,.251,.502}
\definecolor{sidewalk}{rgb}{.957,.137,.910}
\definecolor{building}{rgb}{.275,.275,.275}
\definecolor{wall}{rgb}{.4,.4,.612}
\definecolor{fence}{rgb}{.745,.6,.6}
\definecolor{pole}{rgb}{.6,.6,.6}
\definecolor{tlight}{rgb}{.980,.667,.118}
\definecolor{tsign}{rgb}{.863,.863,0}
\definecolor{vegetation}{rgb}{.420,.557,.137}
\definecolor{terrain}{rgb}{.596,.984,.596}
\definecolor{sky}{rgb}{0,.510,.706}
\definecolor{person}{rgb}{.863,.078,.235}
\definecolor{rider}{rgb}{1,0,0}
\definecolor{car}{rgb}{0,0,.557}
\definecolor{truck}{rgb}{0,0,.275}
\definecolor{bus}{rgb}{0,.235,.392}
\definecolor{train}{rgb}{0,.314,.392}
\definecolor{motorbike}{rgb}{0,0,.902}
\definecolor{bicycle}{rgb}{.467,.043,.125}
\definecolor{unlabelled}{rgb}{0,0,0}

\begin{figure*}[ht]
    \newcolumntype{Y}{>{\centering\arraybackslash}X}
    \centering
    \begin{subfigure}{\textwidth}
    \hspace*{.1em}%
    \begin{subfigure}{\imgWidth}
        \caption{\scriptsize{RGB}}
        \includegraphics[width=\textwidth]{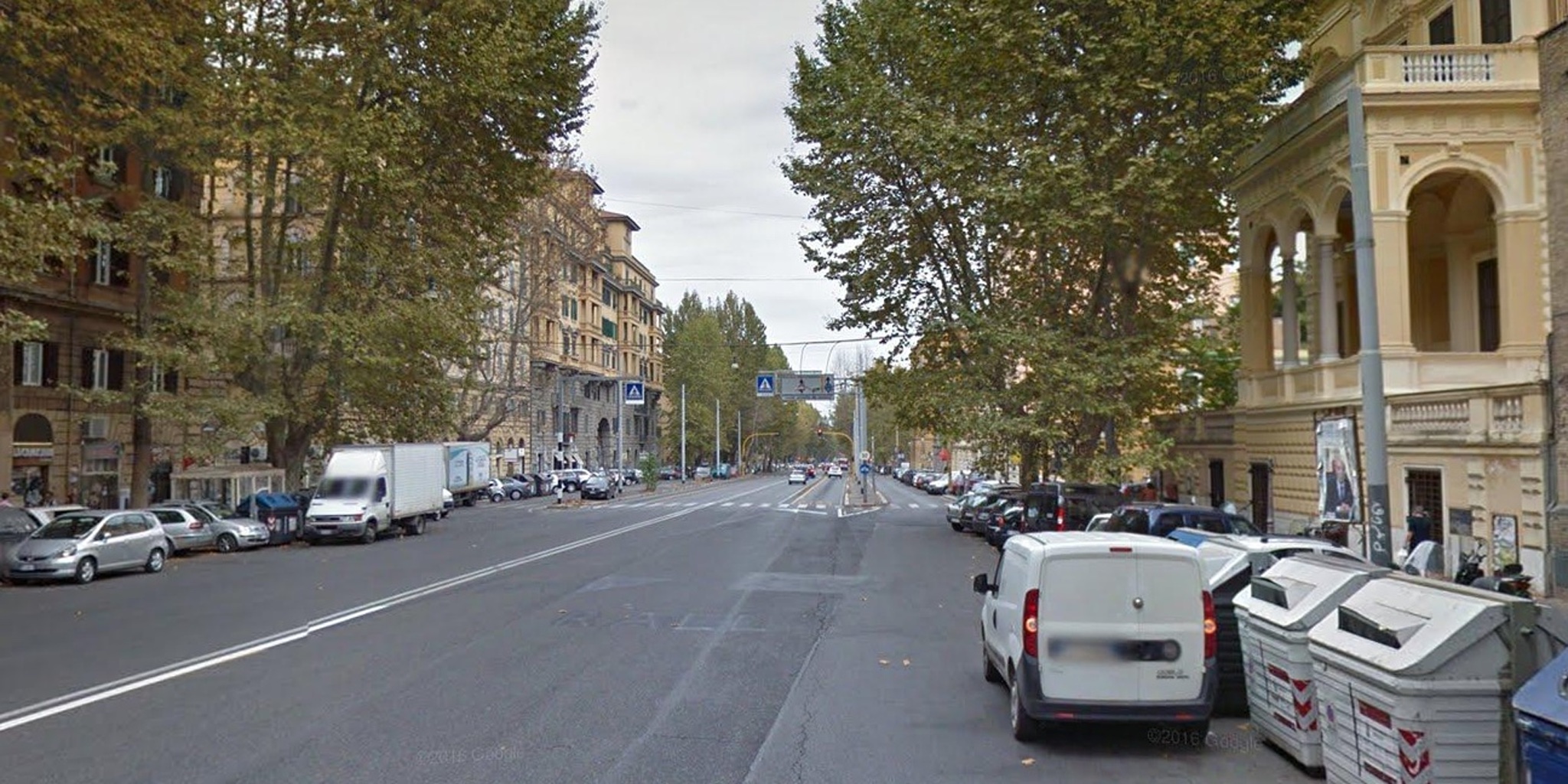}
    \end{subfigure}%
    \begin{subfigure}{\imgWidth}
        \caption{\scriptsize{GT}}
        \includegraphics[width=\textwidth]{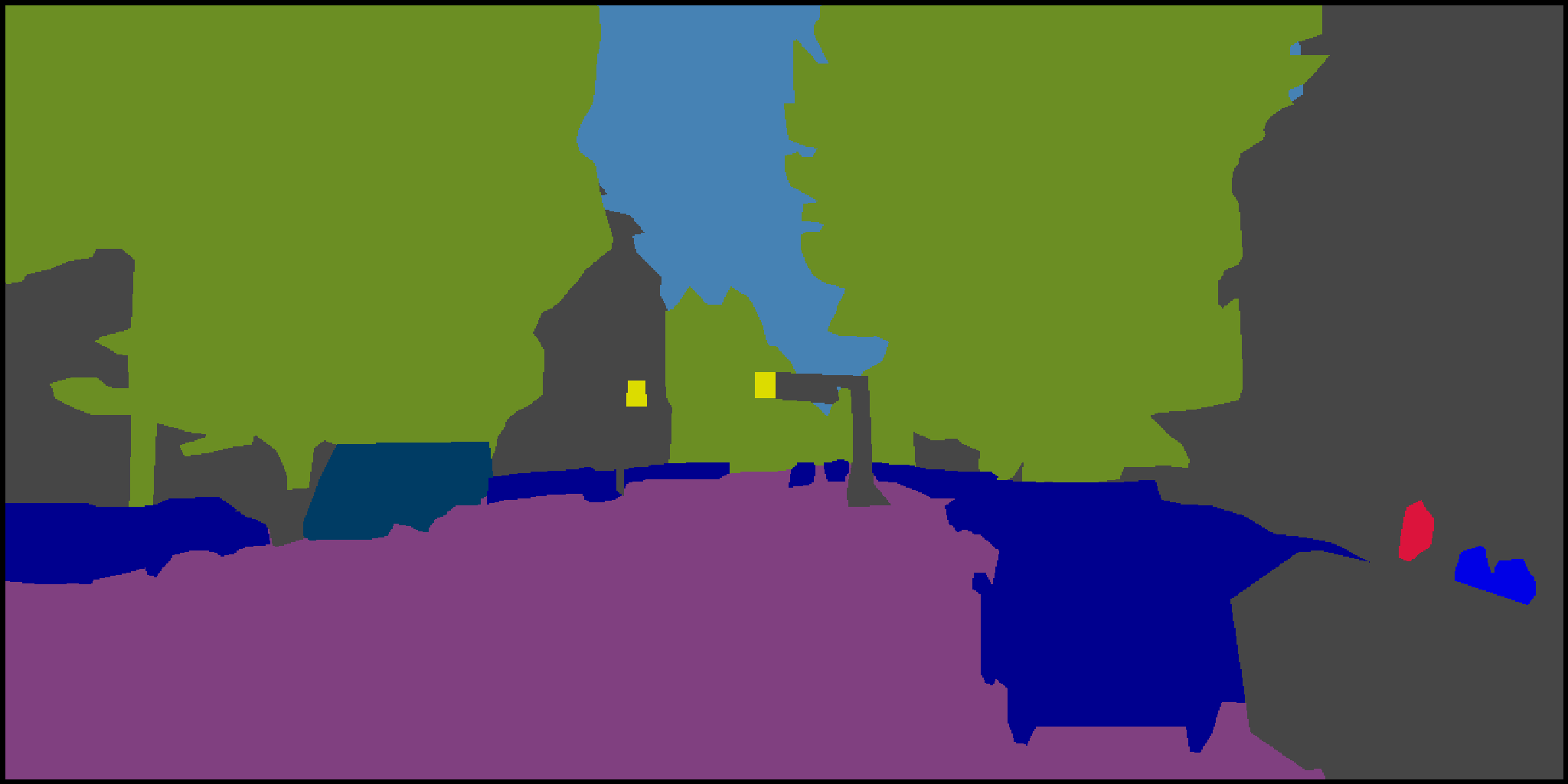}
    \end{subfigure}%
    \begin{subfigure}{\imgWidth}
        \caption{\scriptsize{Source Only}}
        \includegraphics[width=\textwidth]{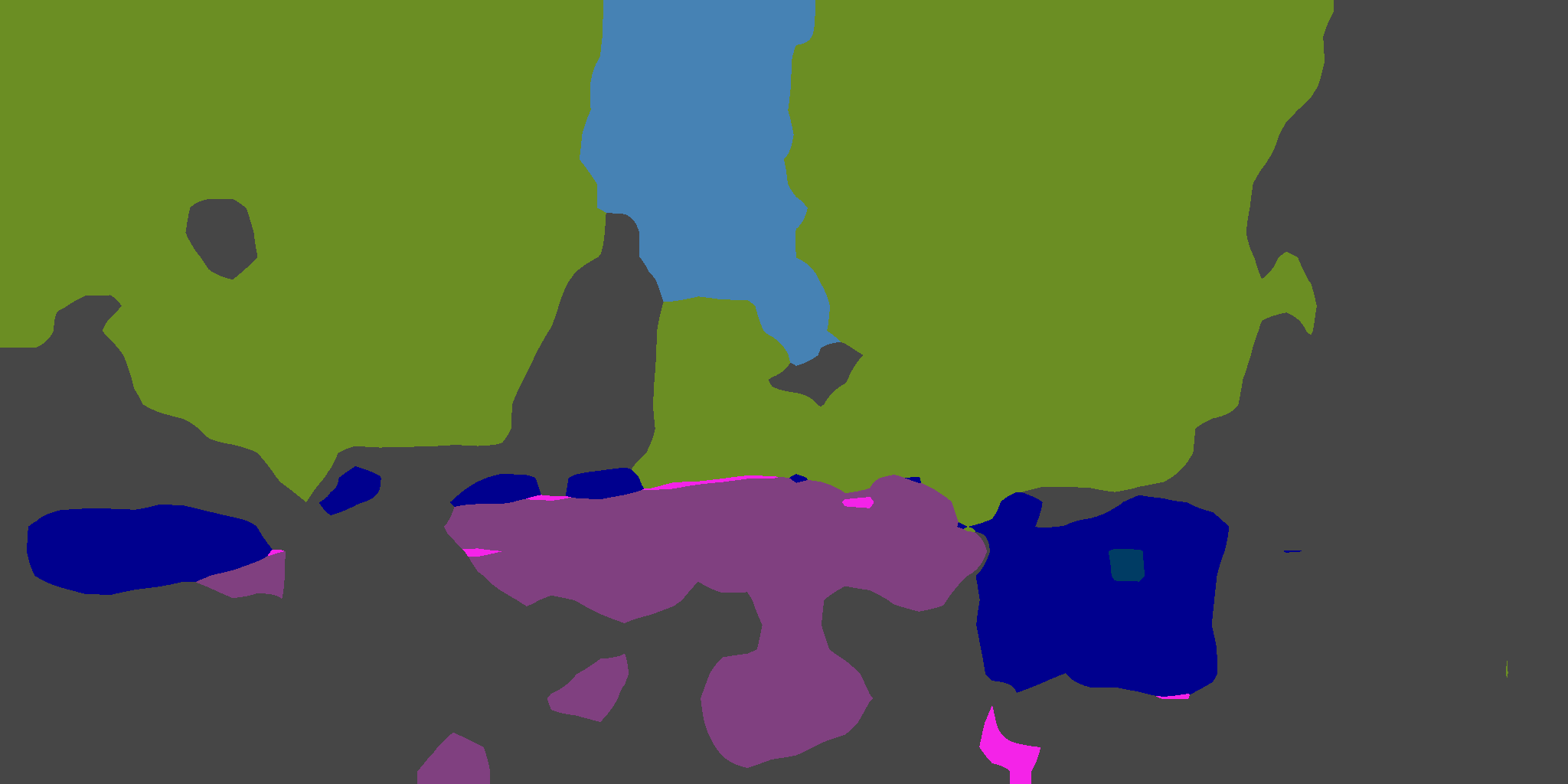}
    \end{subfigure}%
    \begin{subfigure}{\imgWidth}
        \caption{\scriptsize{FedAvg \cite{fedavg}  + Self-Tr.}}
        \includegraphics[width=\textwidth]{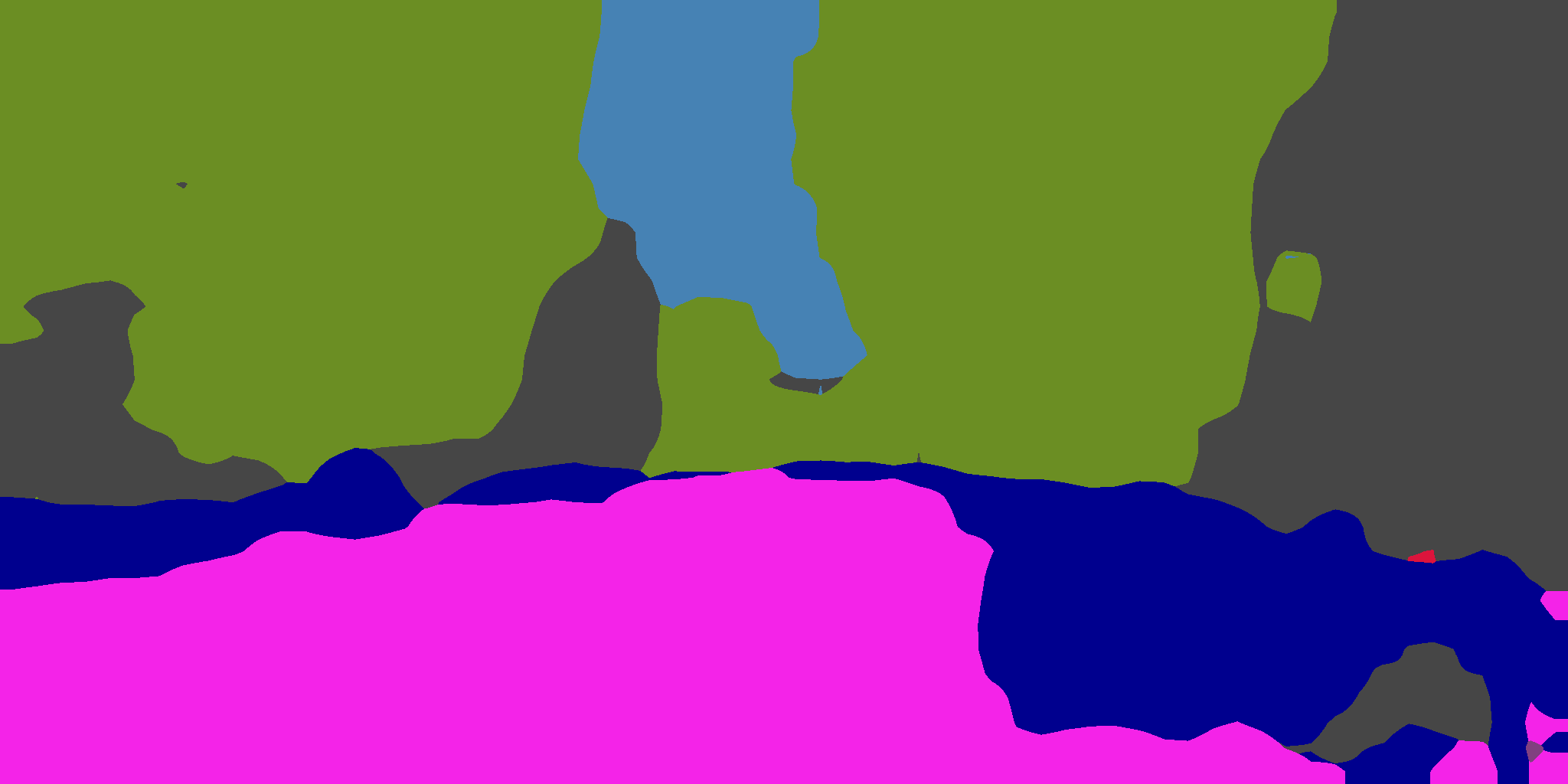}
    \end{subfigure}%
    \begin{subfigure}{\imgWidth}
        \caption{\scriptsize{LADD (all)}}
        \includegraphics[width=\textwidth]{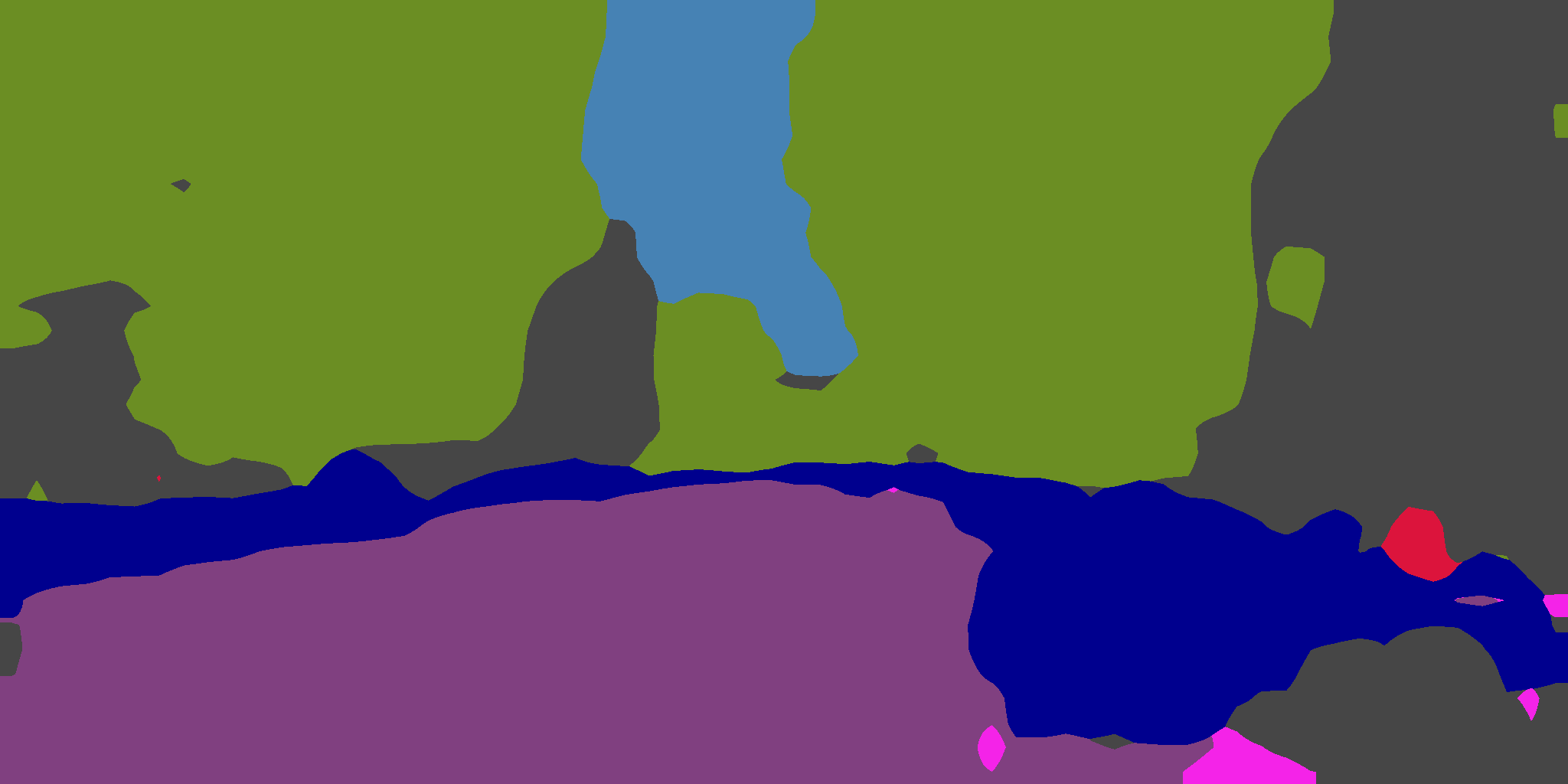}
    \end{subfigure}%
    \end{subfigure}
    \begin{subfigure}{\textwidth}
    \hspace*{.1em}%
    \begin{subfigure}{\imgWidth}
        \includegraphics[width=\textwidth]{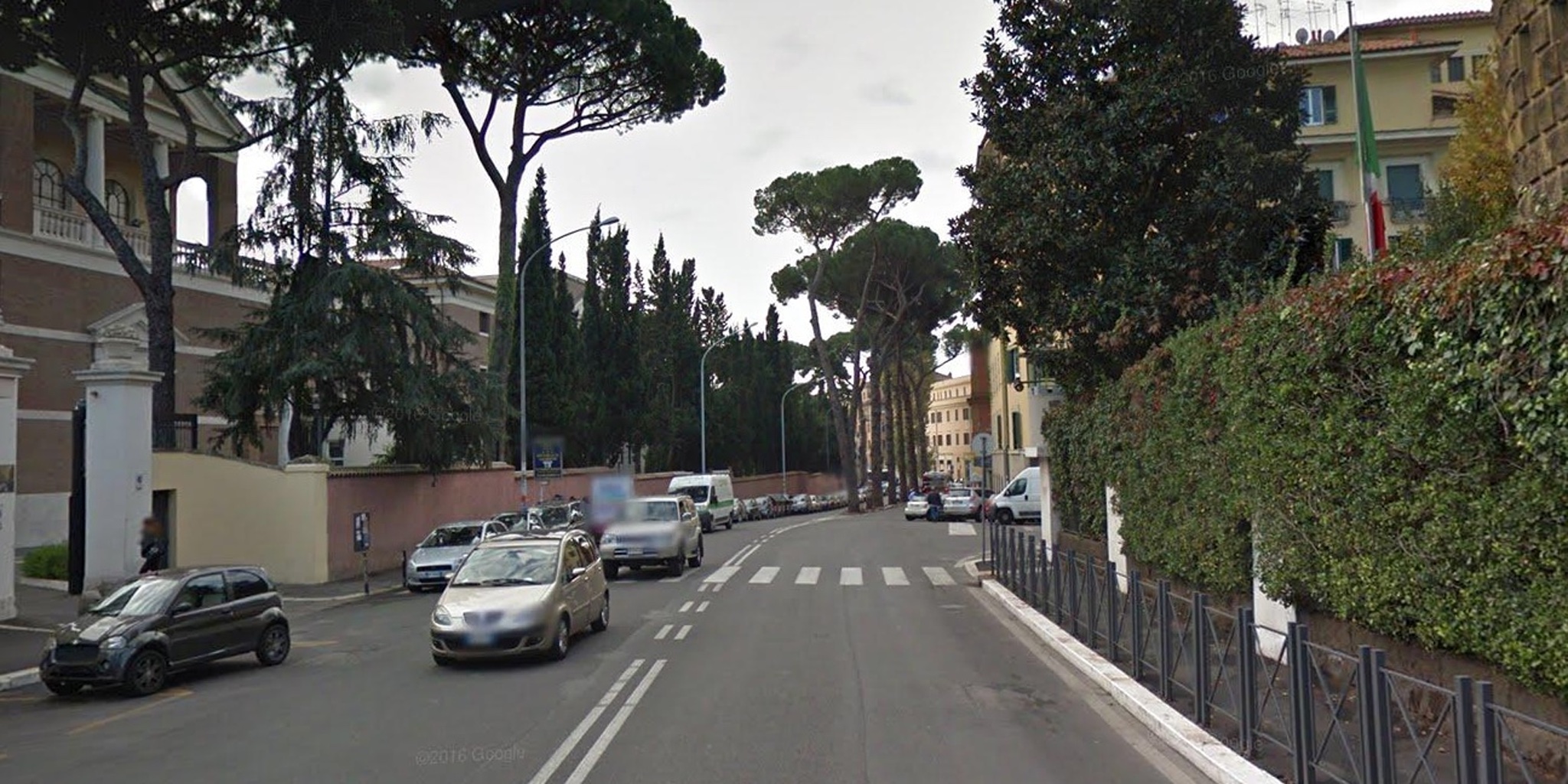}
    \end{subfigure}%
    \begin{subfigure}{\imgWidth}
        \includegraphics[width=\textwidth]{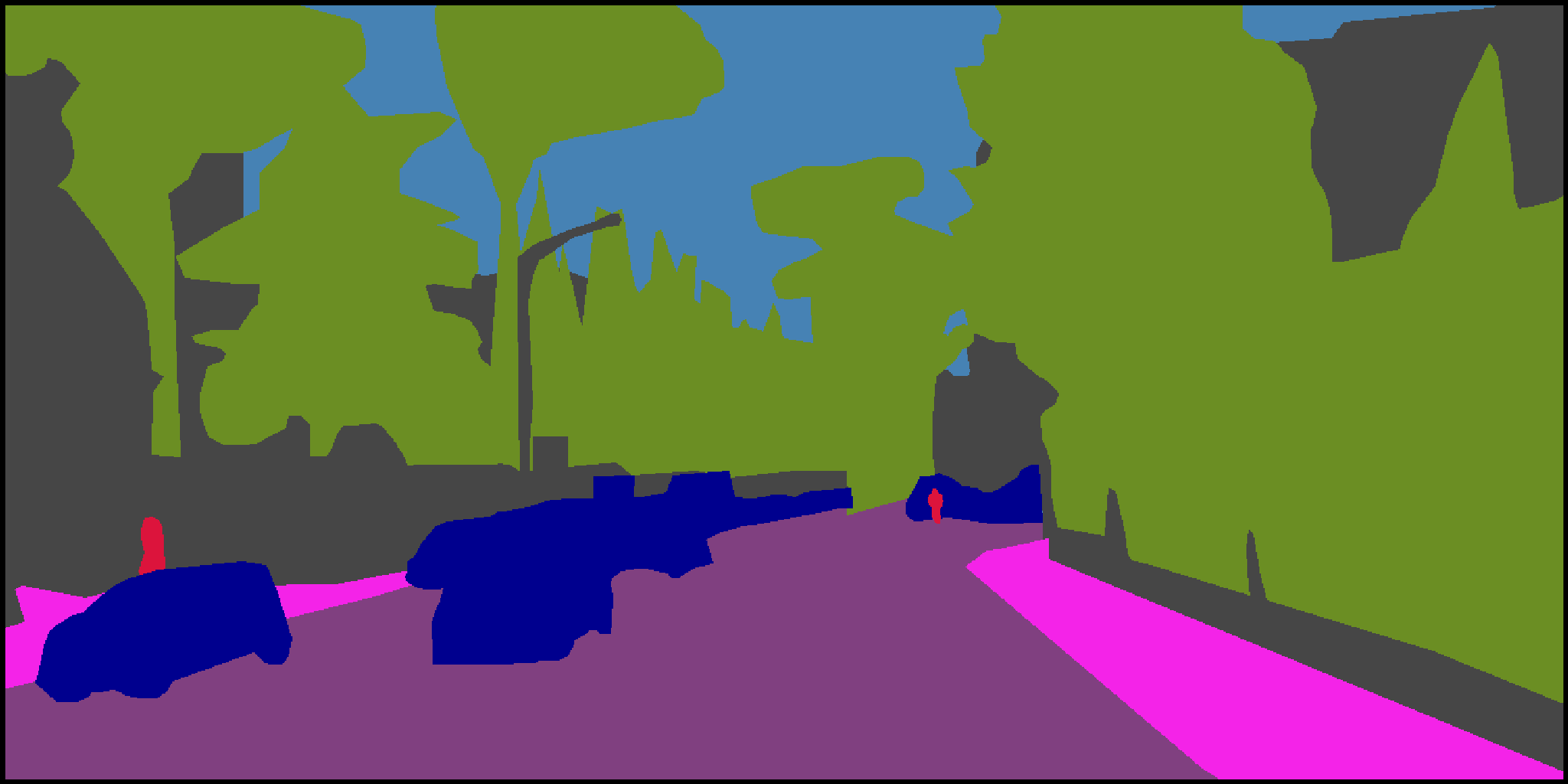}
    \end{subfigure}%
    \begin{subfigure}{\imgWidth}
        \includegraphics[width=\textwidth]{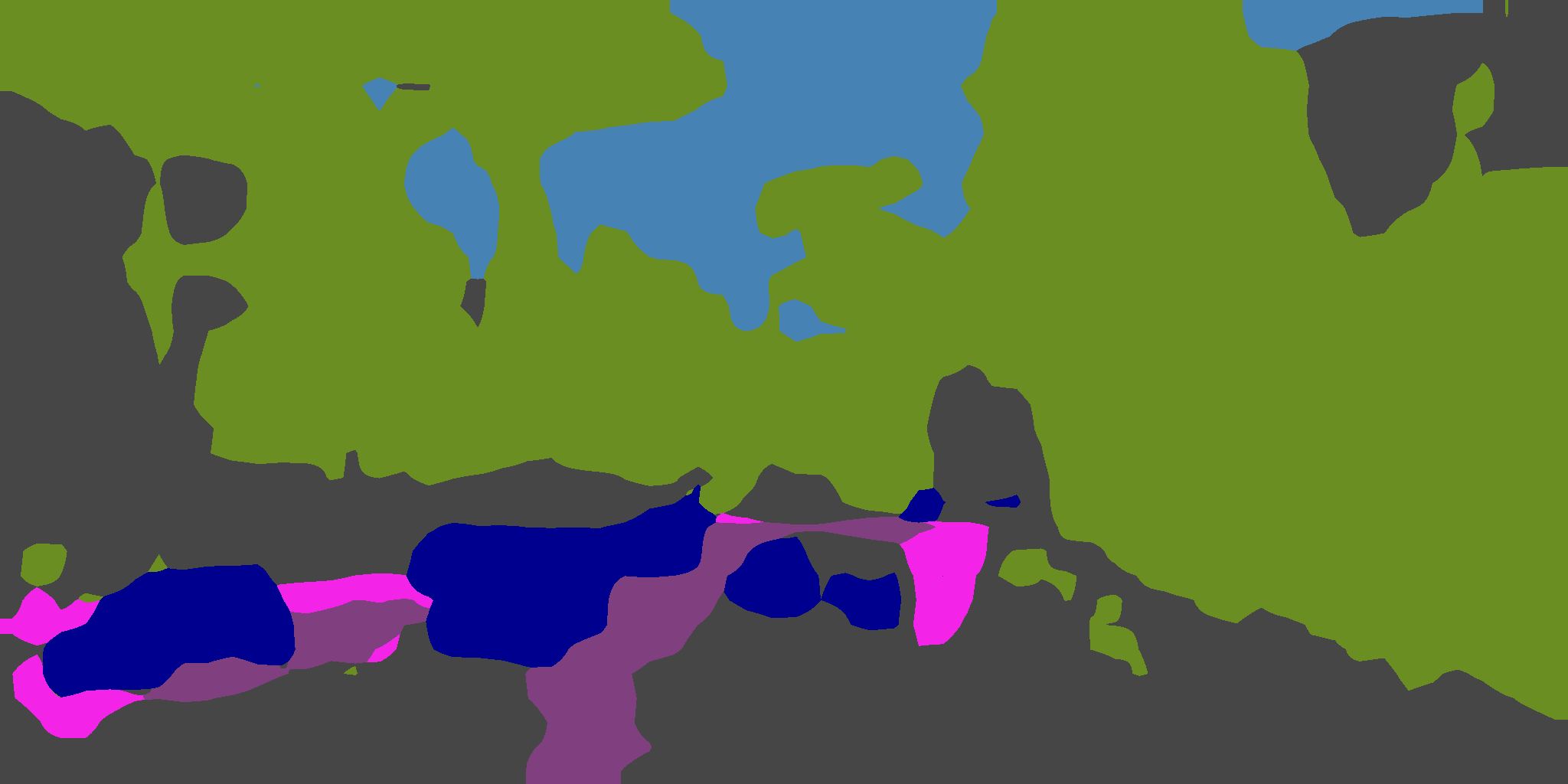}
    \end{subfigure}%
    \begin{subfigure}{\imgWidth}
        \includegraphics[width=\textwidth]{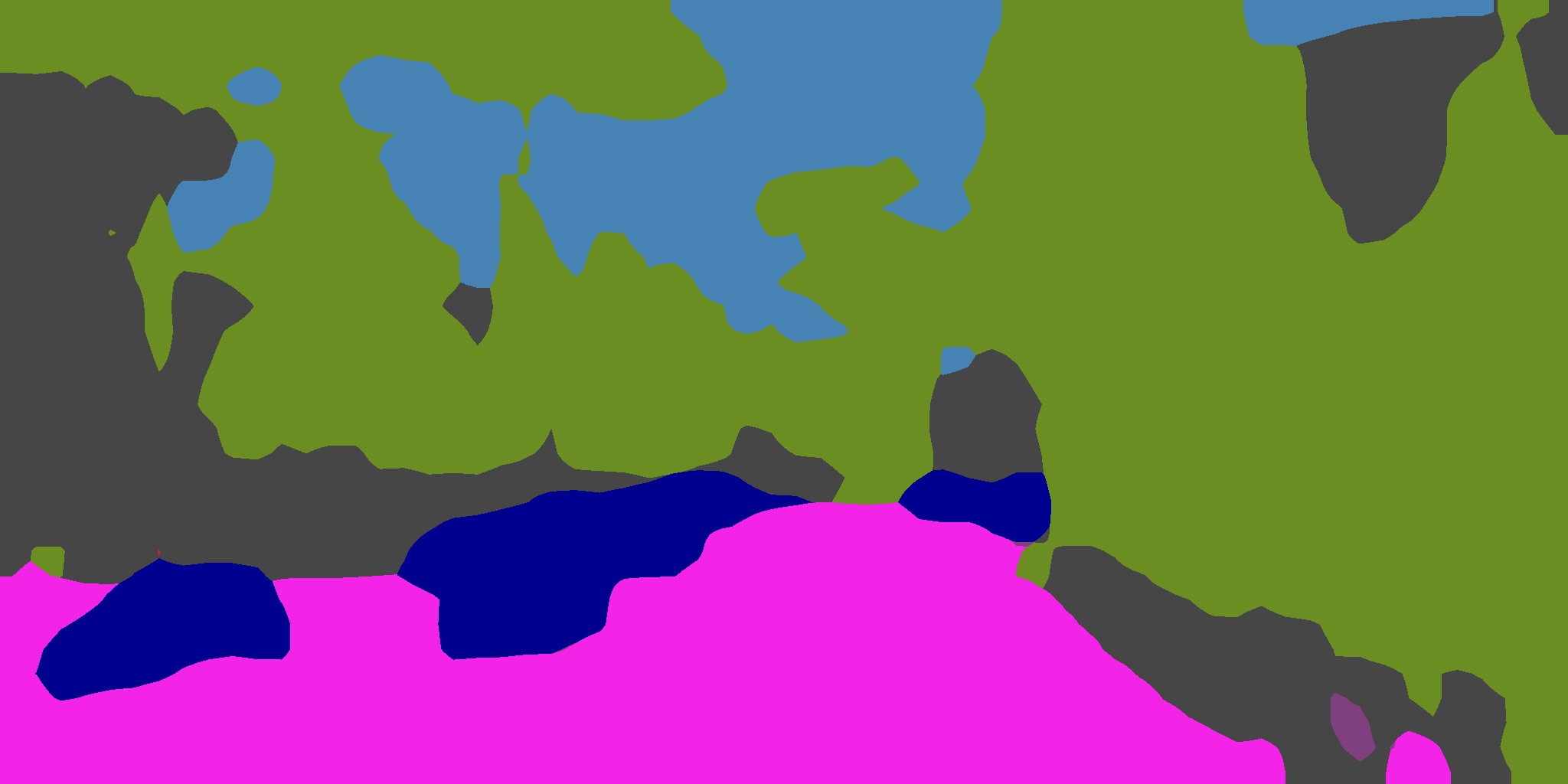}
    \end{subfigure}%
    \begin{subfigure}{\imgWidth}
        \includegraphics[width=\textwidth]{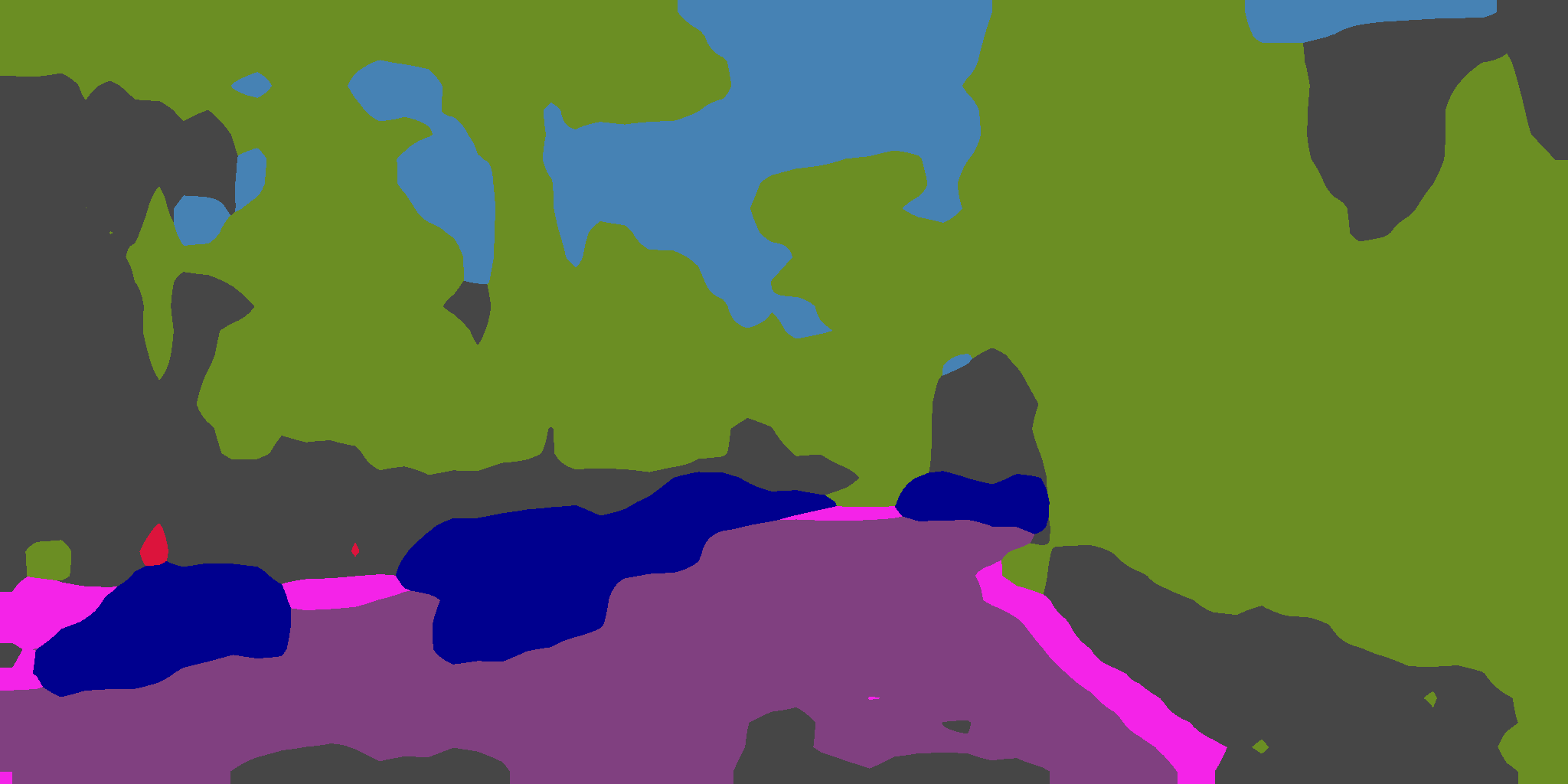}
    \end{subfigure}%
    \end{subfigure}
    \begin{subfigure}{\textwidth}
    \hspace*{.1em}%
    \begin{subfigure}{\imgWidth}
        \includegraphics[width=\textwidth]{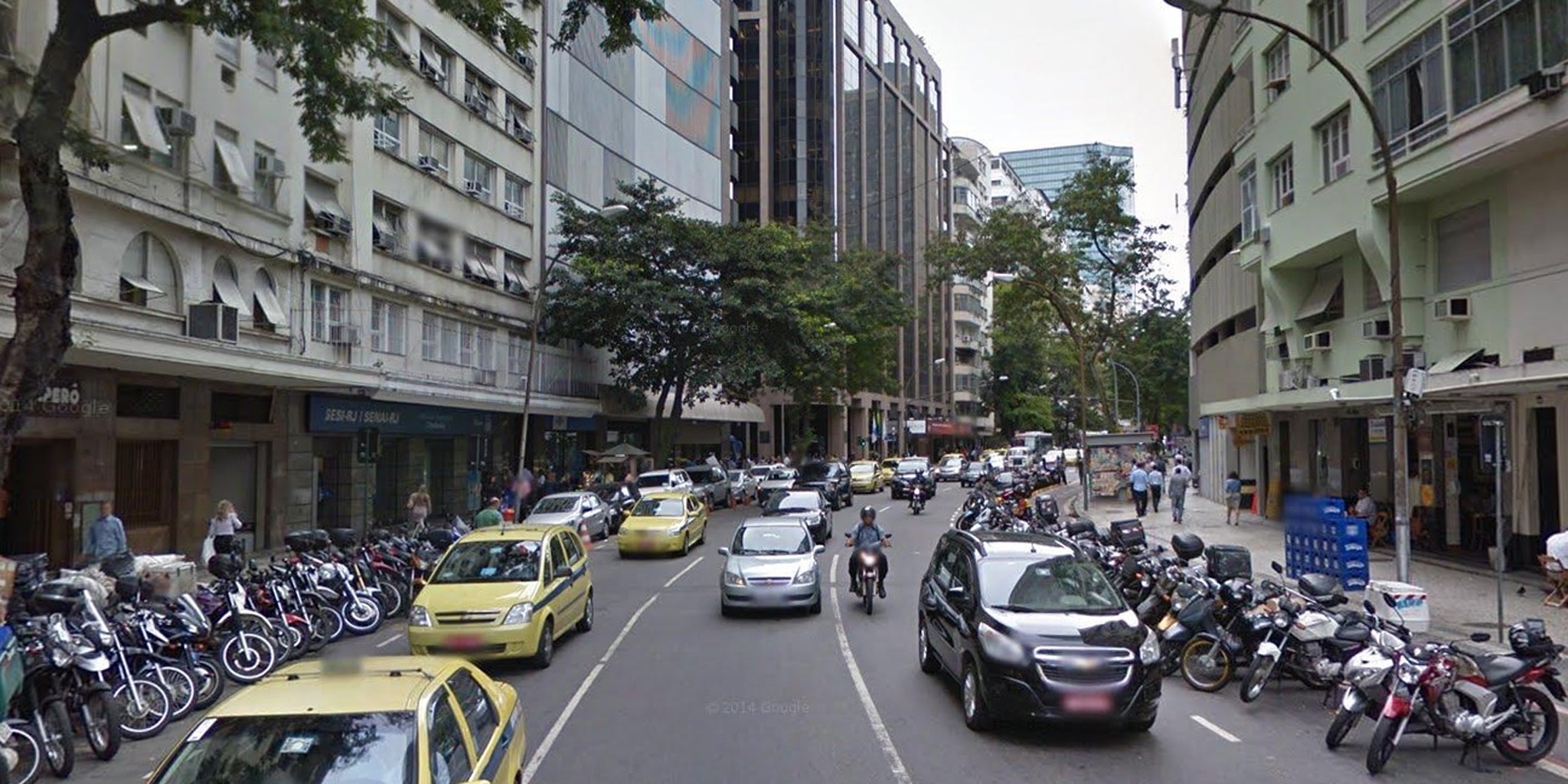}
    \end{subfigure}%
    \begin{subfigure}{\imgWidth}
        \includegraphics[width=\textwidth]{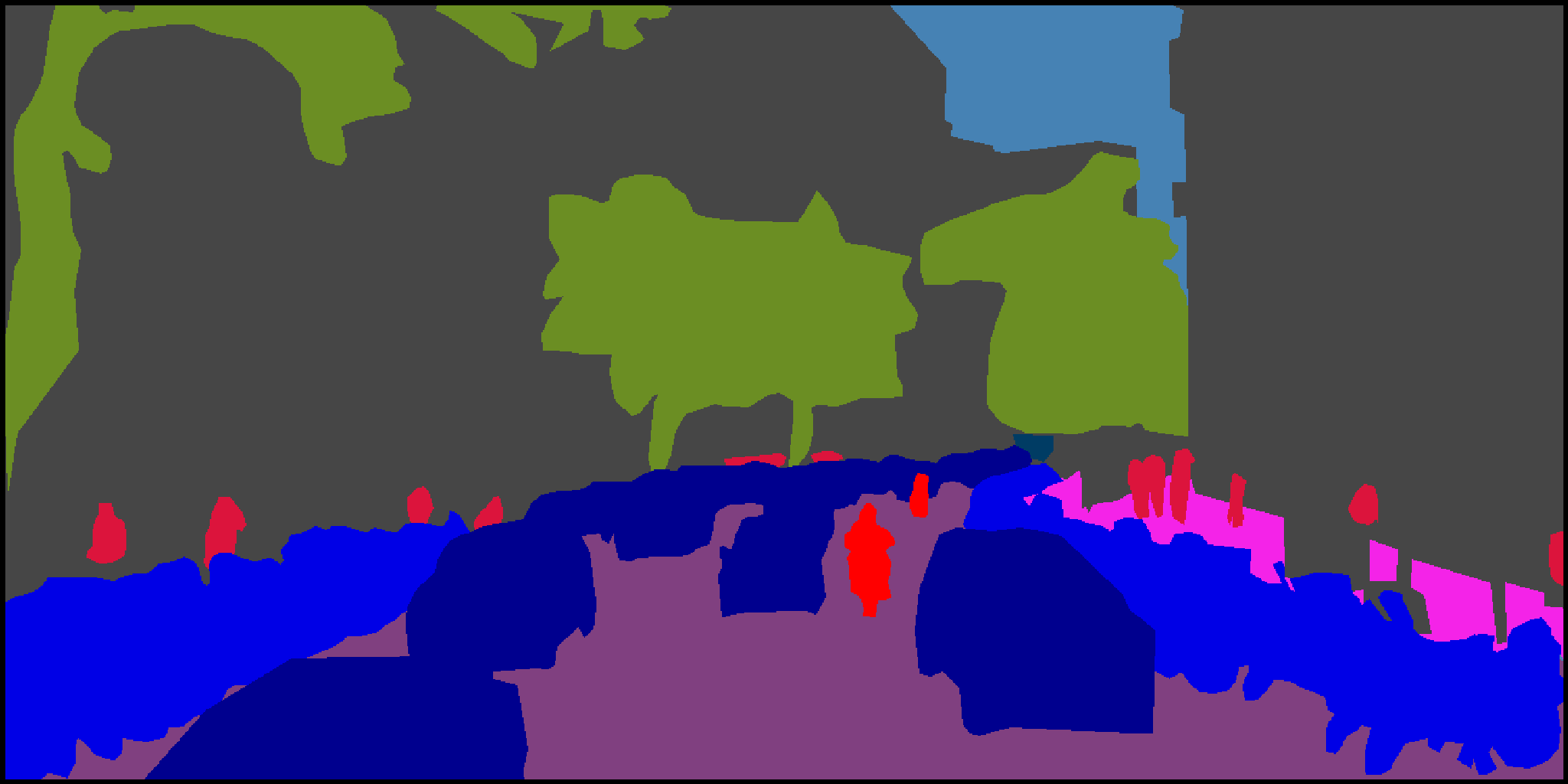}
    \end{subfigure}%
    \begin{subfigure}{\imgWidth}
        \includegraphics[width=\textwidth]{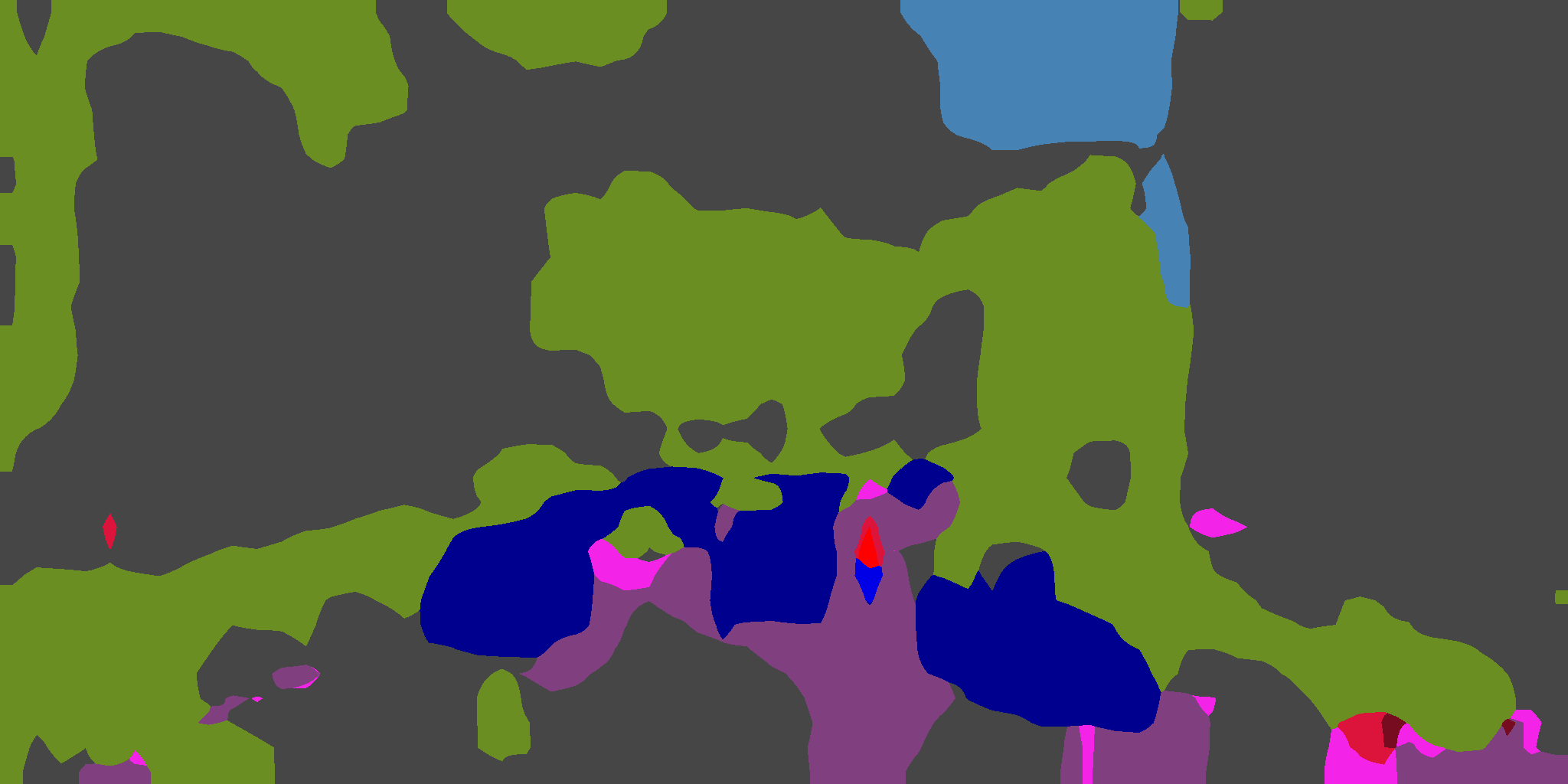}
    \end{subfigure}%
    \begin{subfigure}{\imgWidth}
        \includegraphics[width=\textwidth]{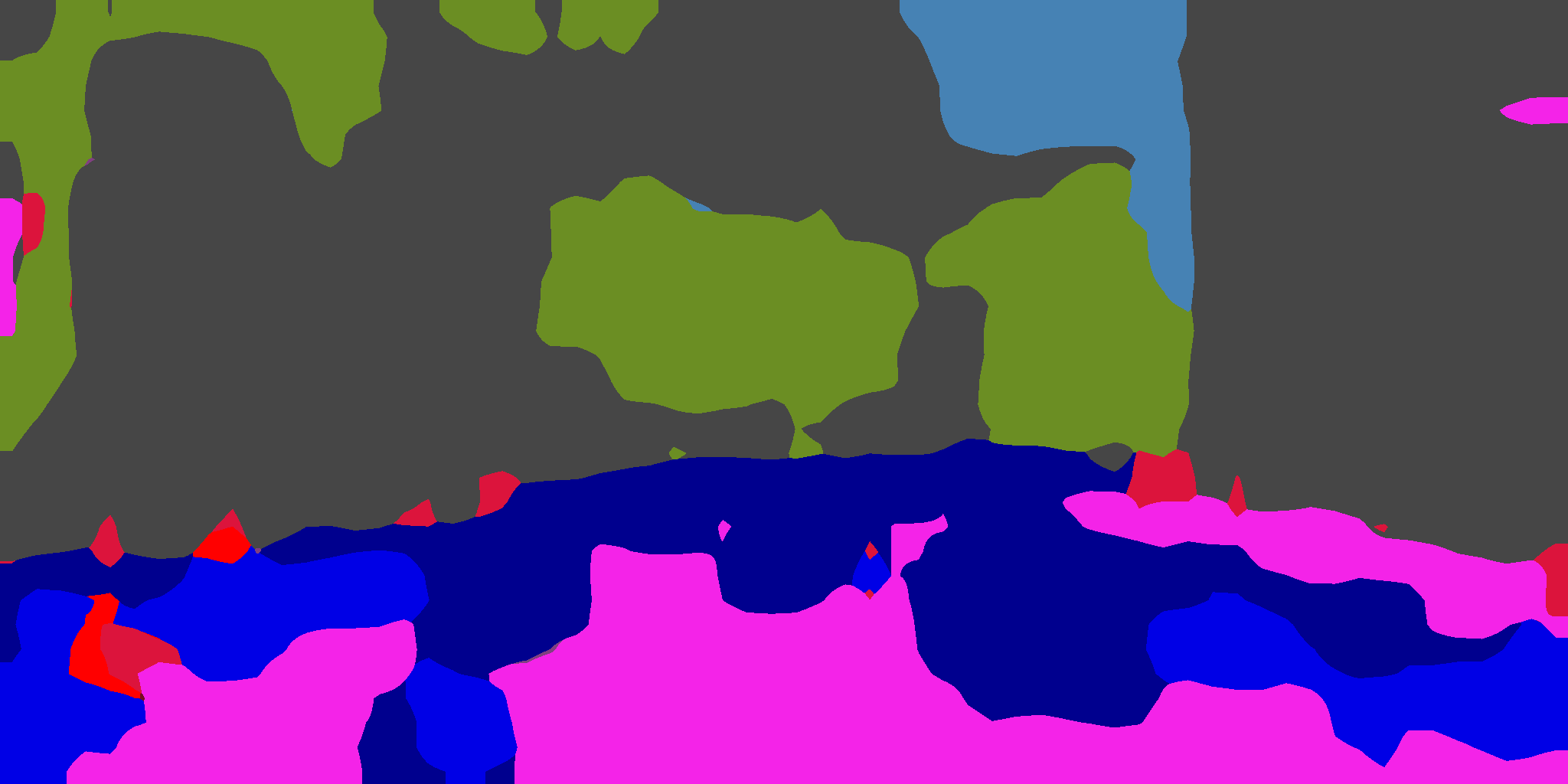}
    \end{subfigure}%
    \begin{subfigure}{\imgWidth}
        \includegraphics[width=\textwidth]{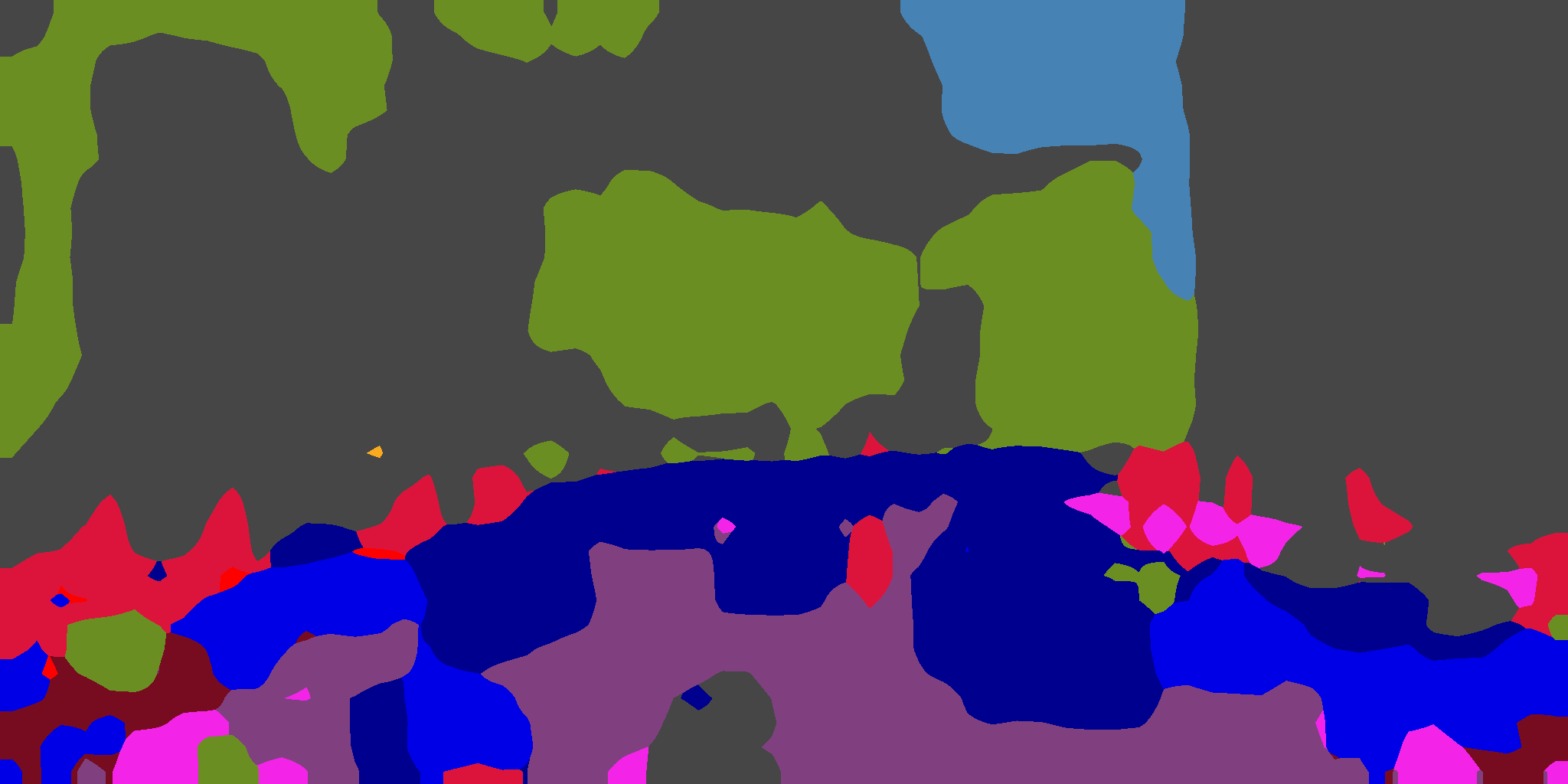}
    \end{subfigure}%
    \end{subfigure}
    \begin{subfigure}{\textwidth}
        \tiny
        \begin{tabularx}{\textwidth}{YYYYYYYYYY}
            \cellcolor{road} \textcolor{white}{Road} & \cellcolor{sidewalk} Sidewalk & \cellcolor{building} \textcolor{white}{Building} & \cellcolor{tlight} T. Light & \cellcolor{tsign} T. Sign & \cellcolor{vegetation} \textcolor{white}{Vegetation} &
            \cellcolor{sky} Sky \\ 
            \cellcolor{person} \textcolor{white}{Person} & \cellcolor{rider} \textcolor{white}{Rider} & \cellcolor{car} \textcolor{white}{Car} & \cellcolor{bus} \textcolor{white}{Bus} &   \cellcolor{bicycle} \textcolor{white}{Bicycle} & \cellcolor{unlabelled} \textcolor{white}{Unlabeled}
        \end{tabularx}
    \end{subfigure}
    \caption{GTA5$\rightarrow$CrossCity qualitative results.}
    \label{fig:quali_cross}
\end{figure*}

\begin{figure*}[ht]
    \newcolumntype{Y}{>{\centering\arraybackslash}X}
    \centering
    \begin{subfigure}{\textwidth}
    \hspace*{.1em}%
    \begin{subfigure}{\imgWidth}
        \caption{\scriptsize{RGB}}
        \includegraphics[width=\textwidth]{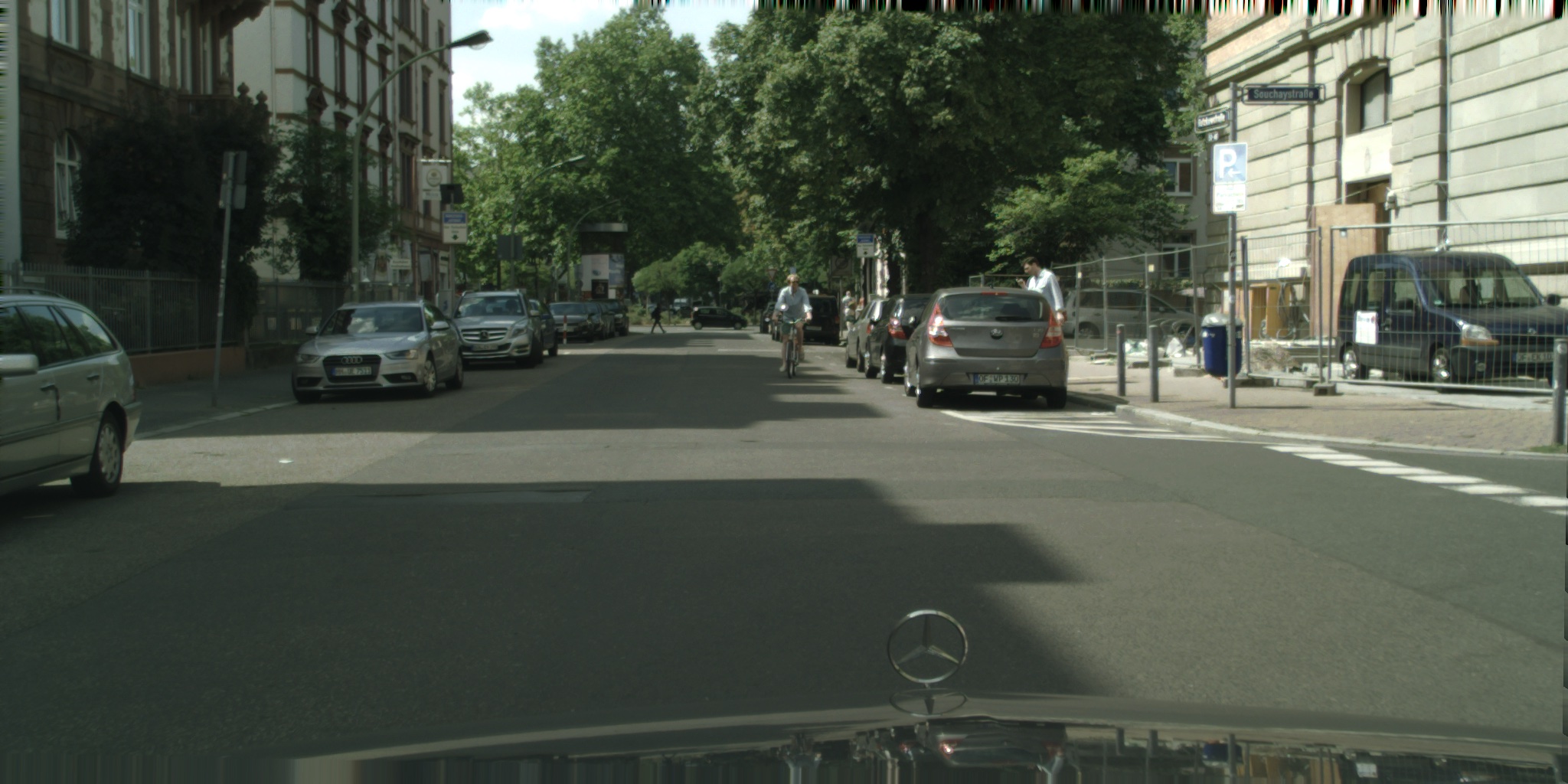}
    \end{subfigure}%
    \begin{subfigure}{\imgWidth}
        \caption{\scriptsize{GT}}
        \includegraphics[width=\textwidth]{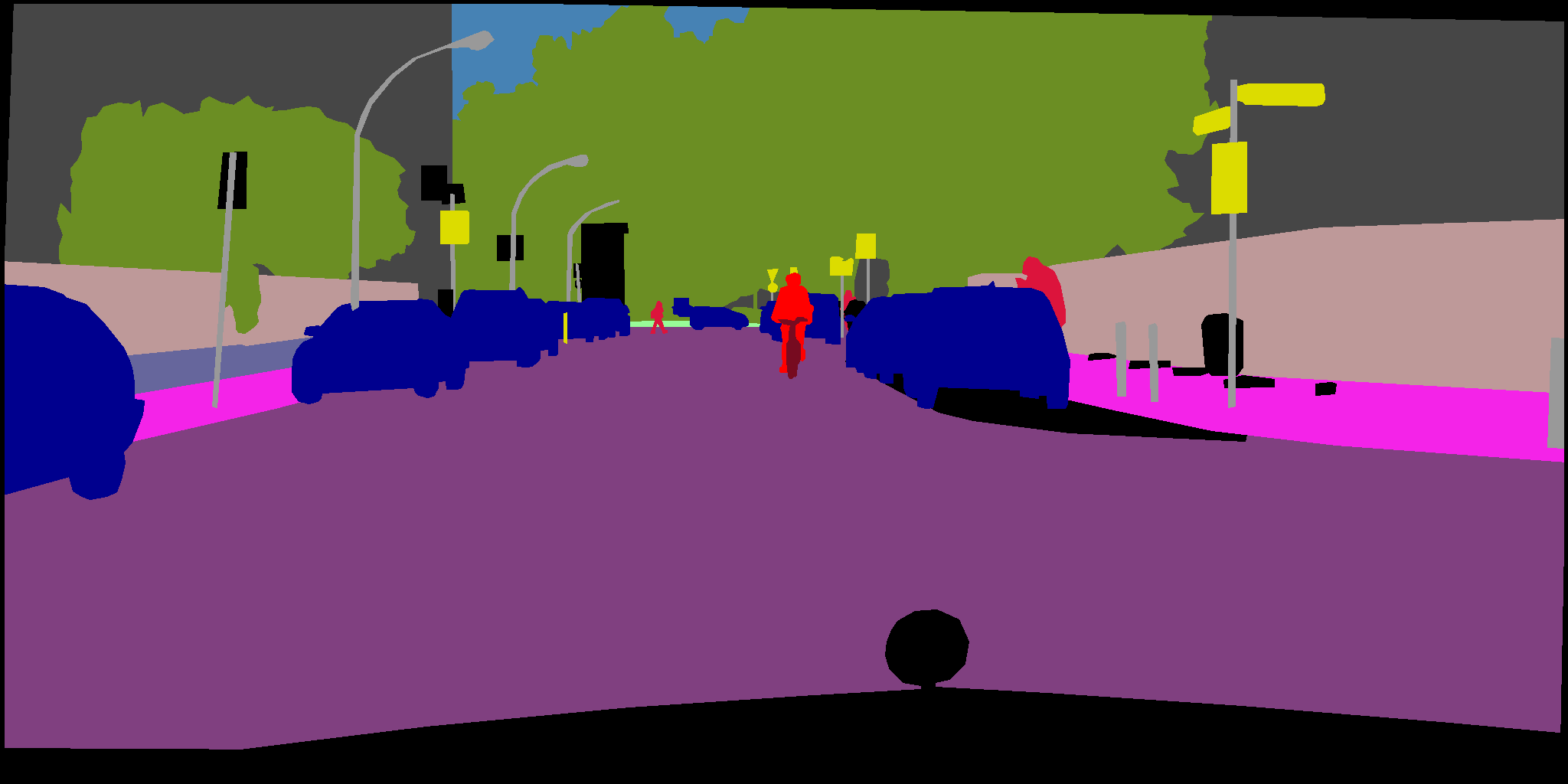}
    \end{subfigure}%
    \begin{subfigure}{\imgWidth}
        \caption{\scriptsize{Source Only}}
        \includegraphics[width=\textwidth]{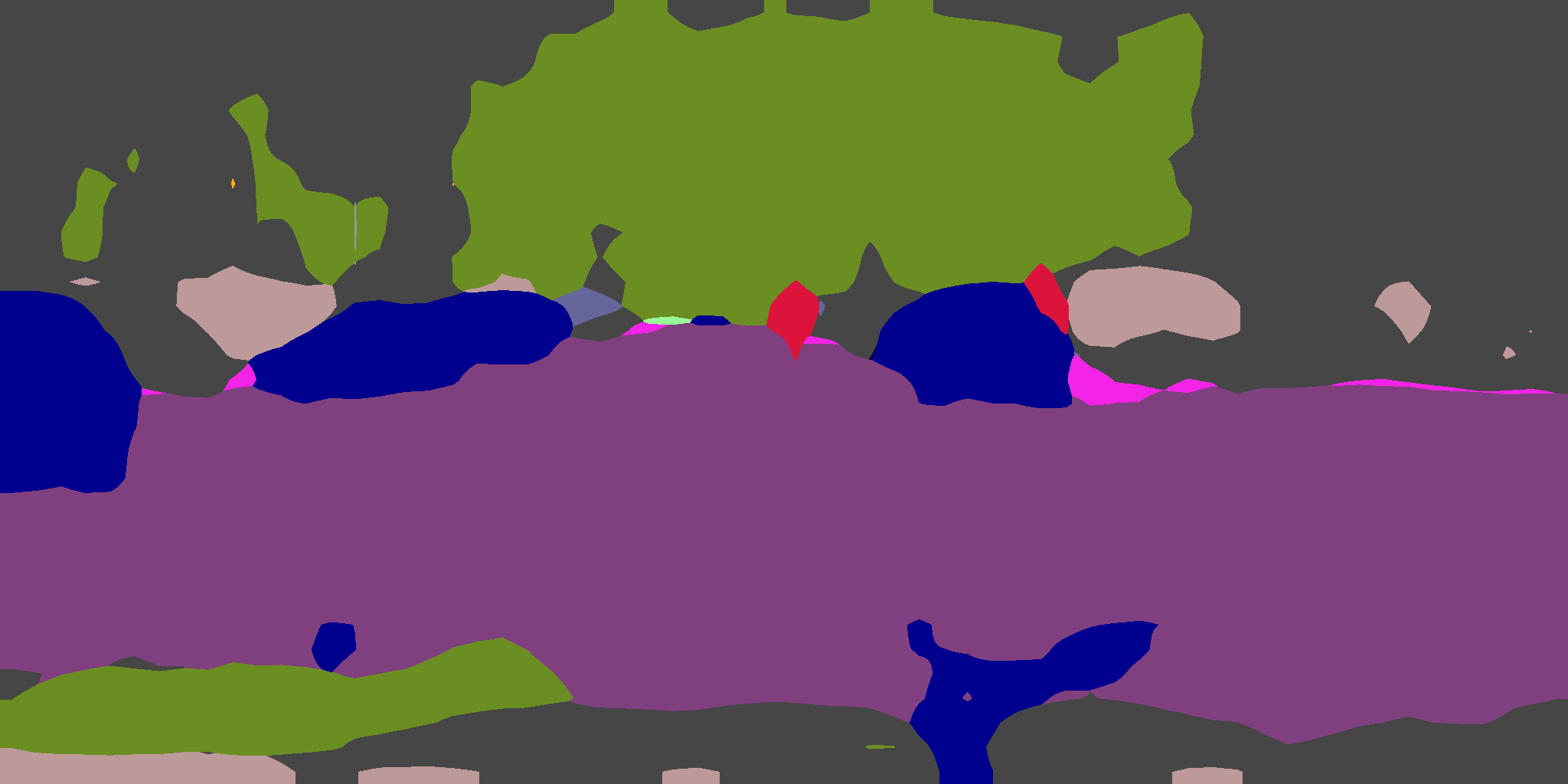}
    \end{subfigure}%
    \begin{subfigure}{\imgWidth}
        \caption{\scriptsize{FedAvg  \cite{fedavg}  + Self-Tr.}}
        \includegraphics[width=\textwidth]{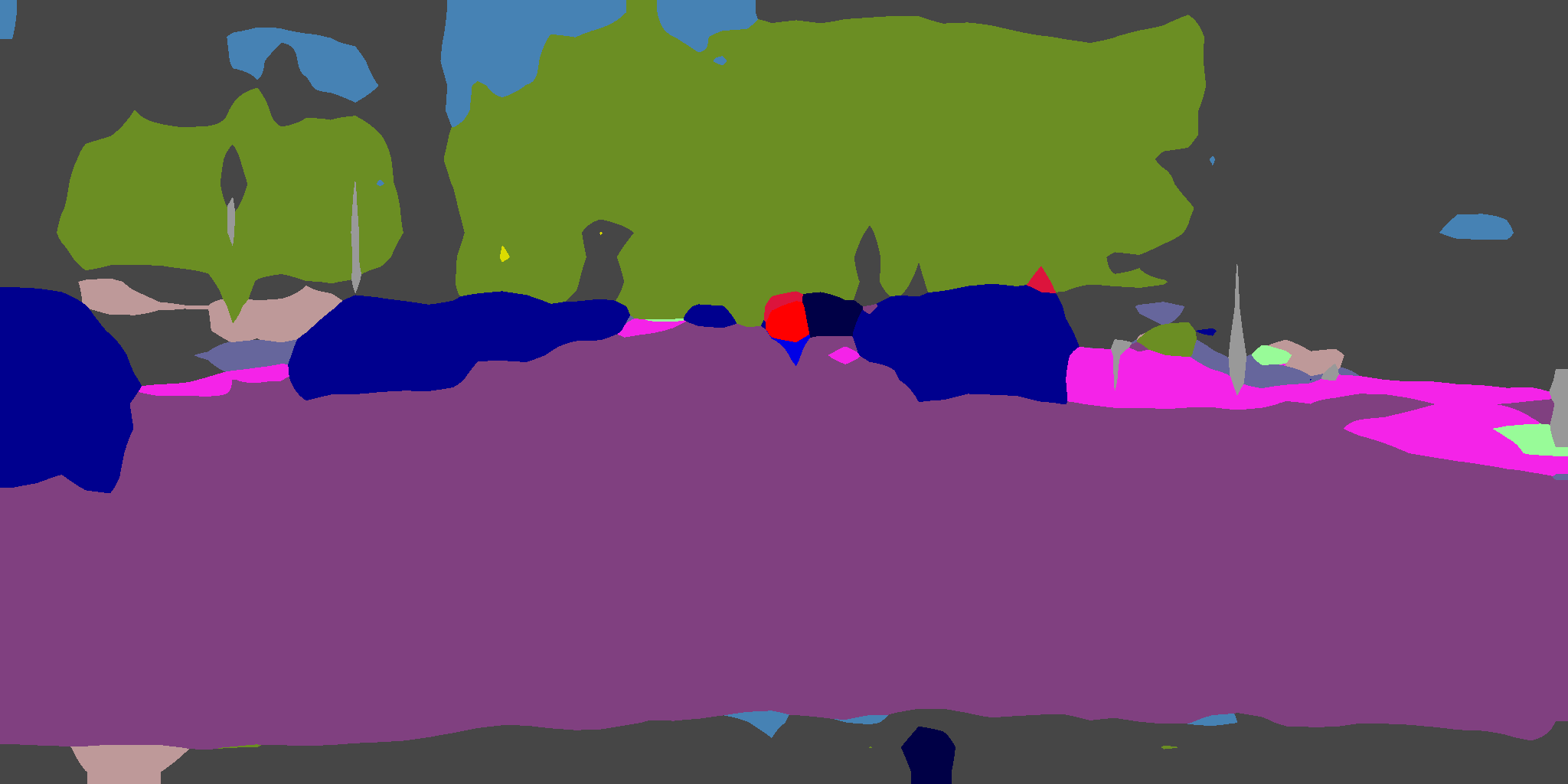}
    \end{subfigure}%
    \begin{subfigure}{\imgWidth}
        \caption{\scriptsize{LADD (all)}}
        \includegraphics[width=\textwidth]{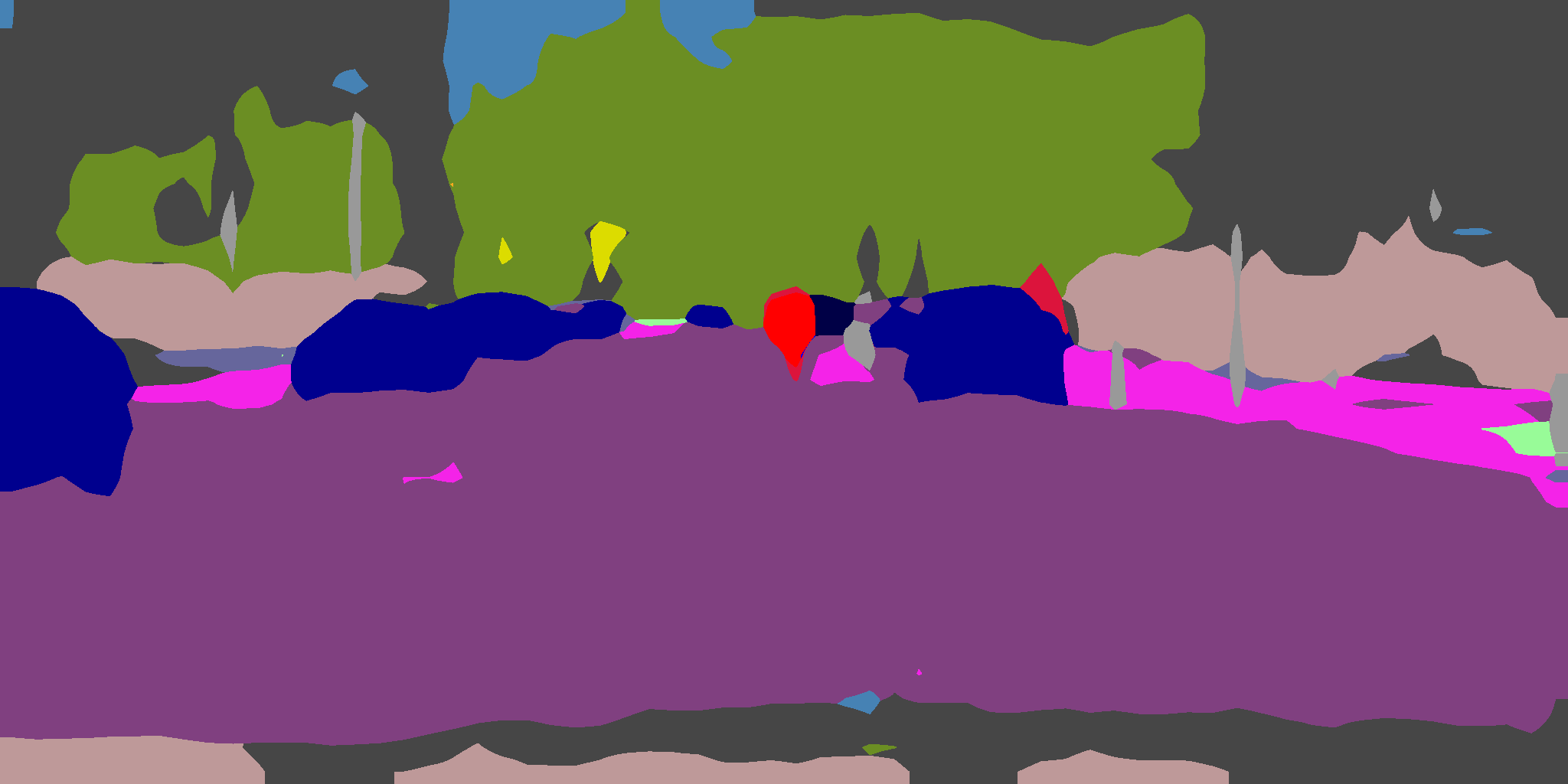}
    \end{subfigure}%
    \end{subfigure}
        \begin{subfigure}{\textwidth}
    \hspace*{.1em}%
    \begin{subfigure}{\imgWidth}
        \includegraphics[width=\textwidth]{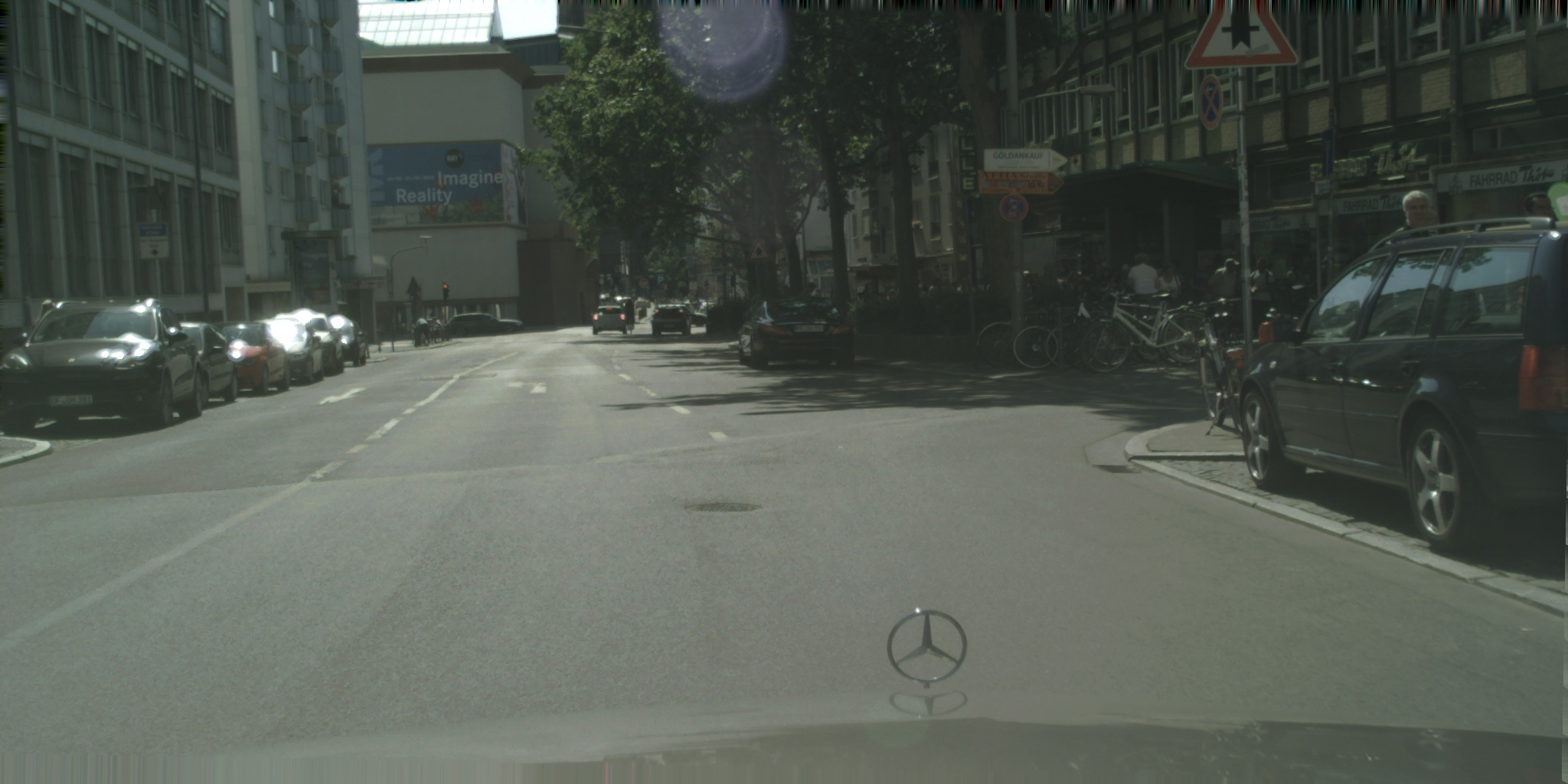}
    \end{subfigure}%
    \begin{subfigure}{\imgWidth}
        \includegraphics[width=\textwidth]{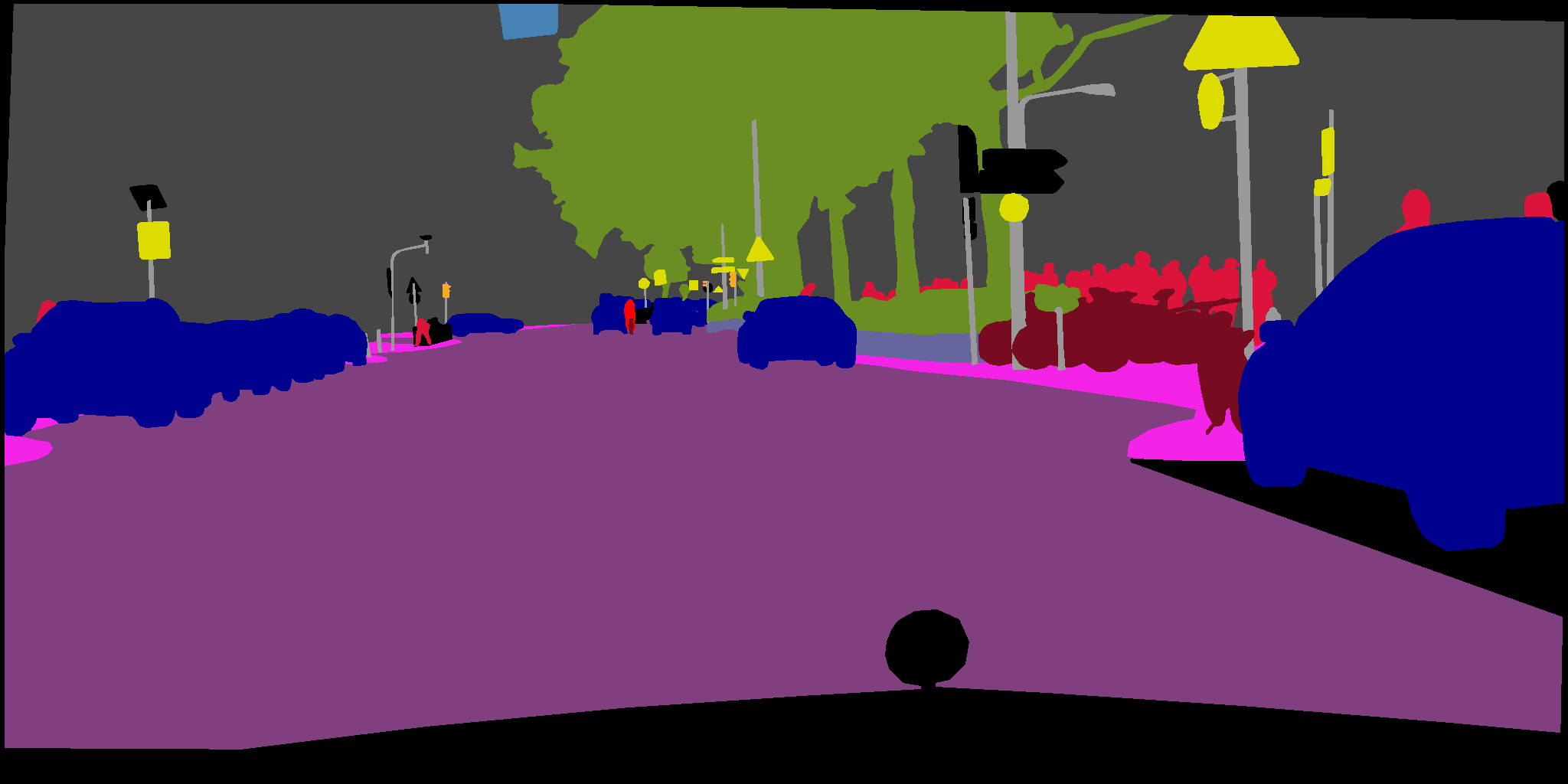}
    \end{subfigure}%
    \begin{subfigure}{\imgWidth}
        \includegraphics[width=\textwidth]{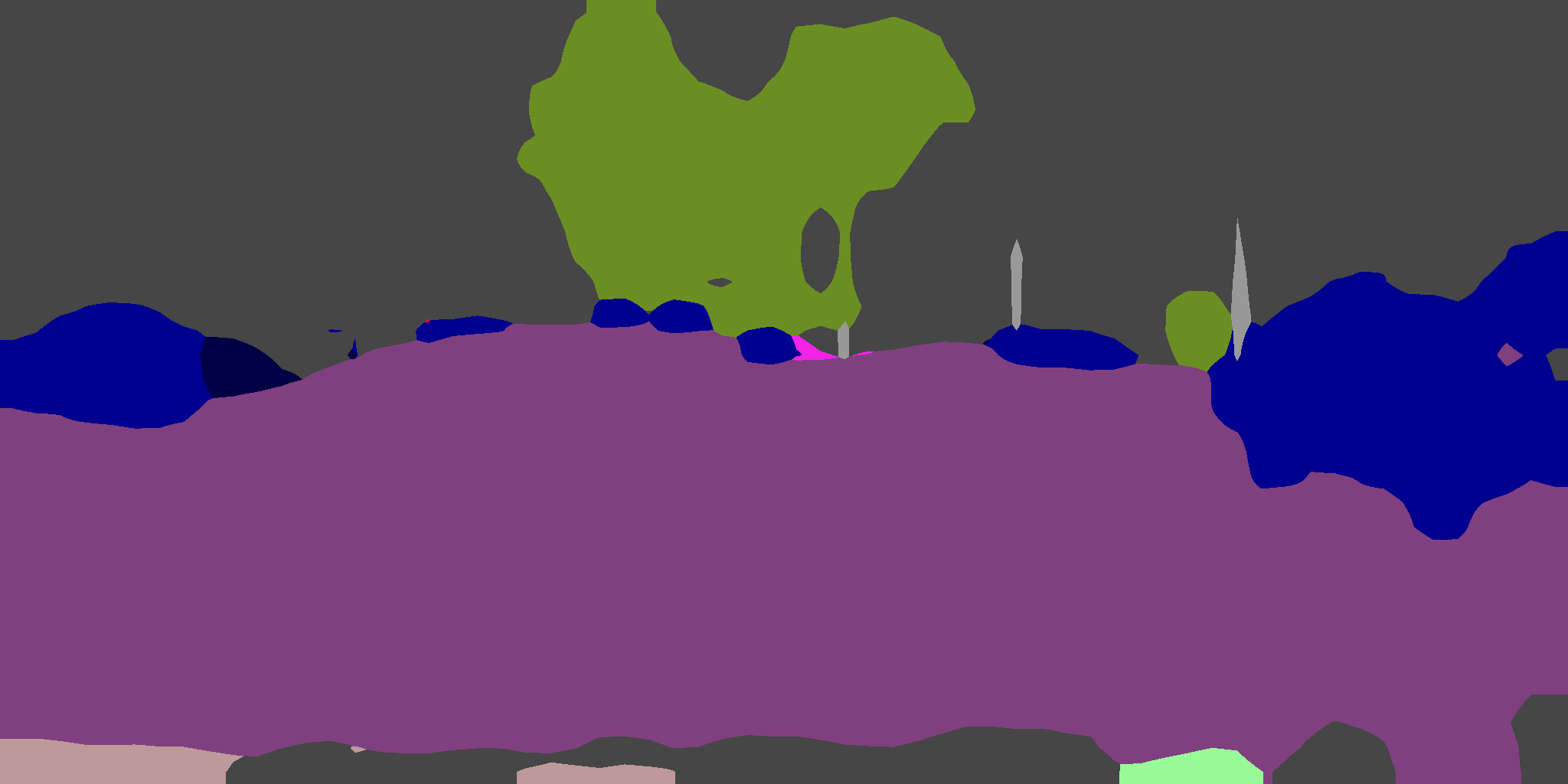}
    \end{subfigure}%
    \begin{subfigure}{\imgWidth}
        \includegraphics[width=\textwidth]{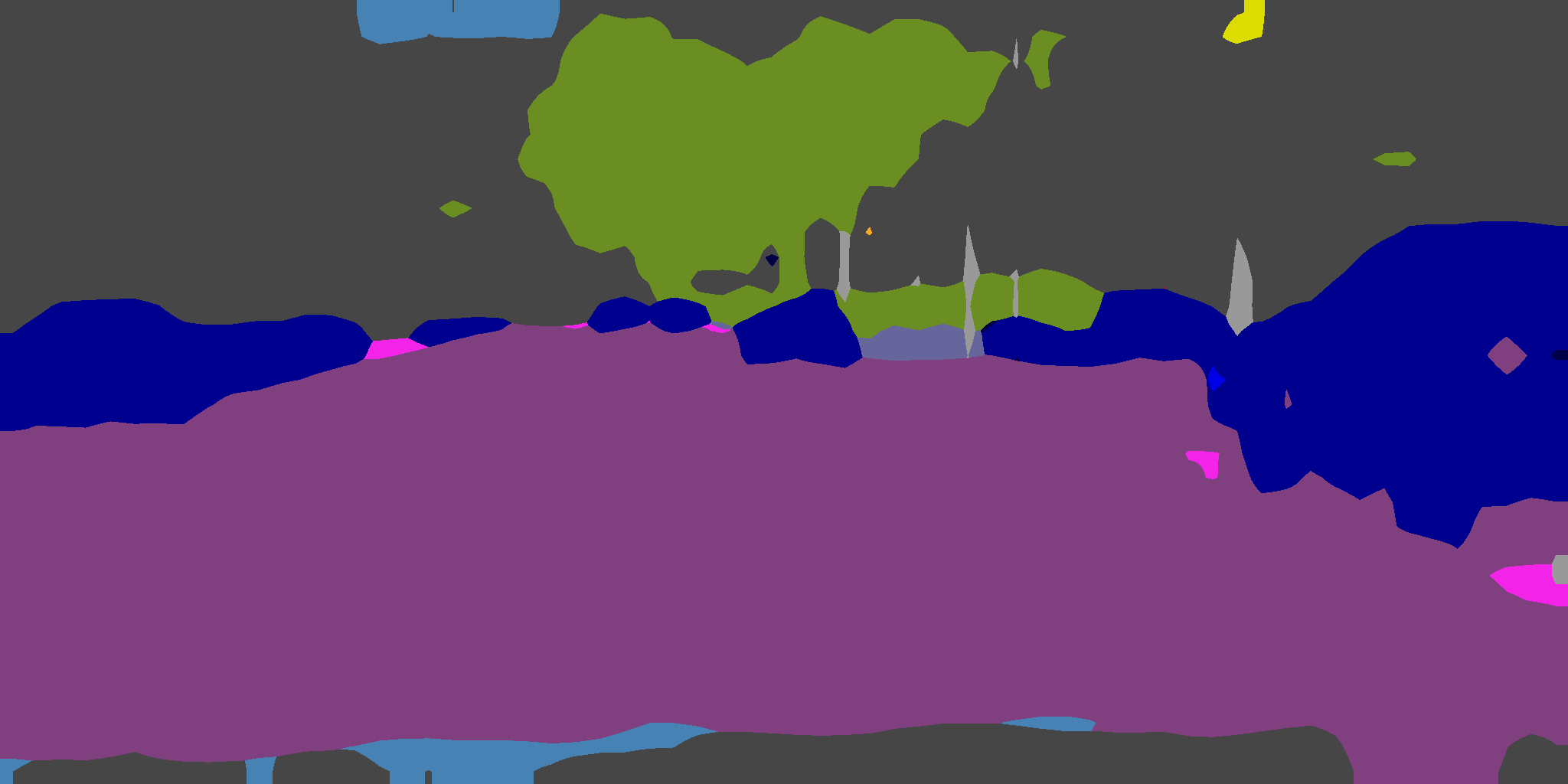}
    \end{subfigure}%
    \begin{subfigure}{\imgWidth}
        \includegraphics[width=\textwidth]{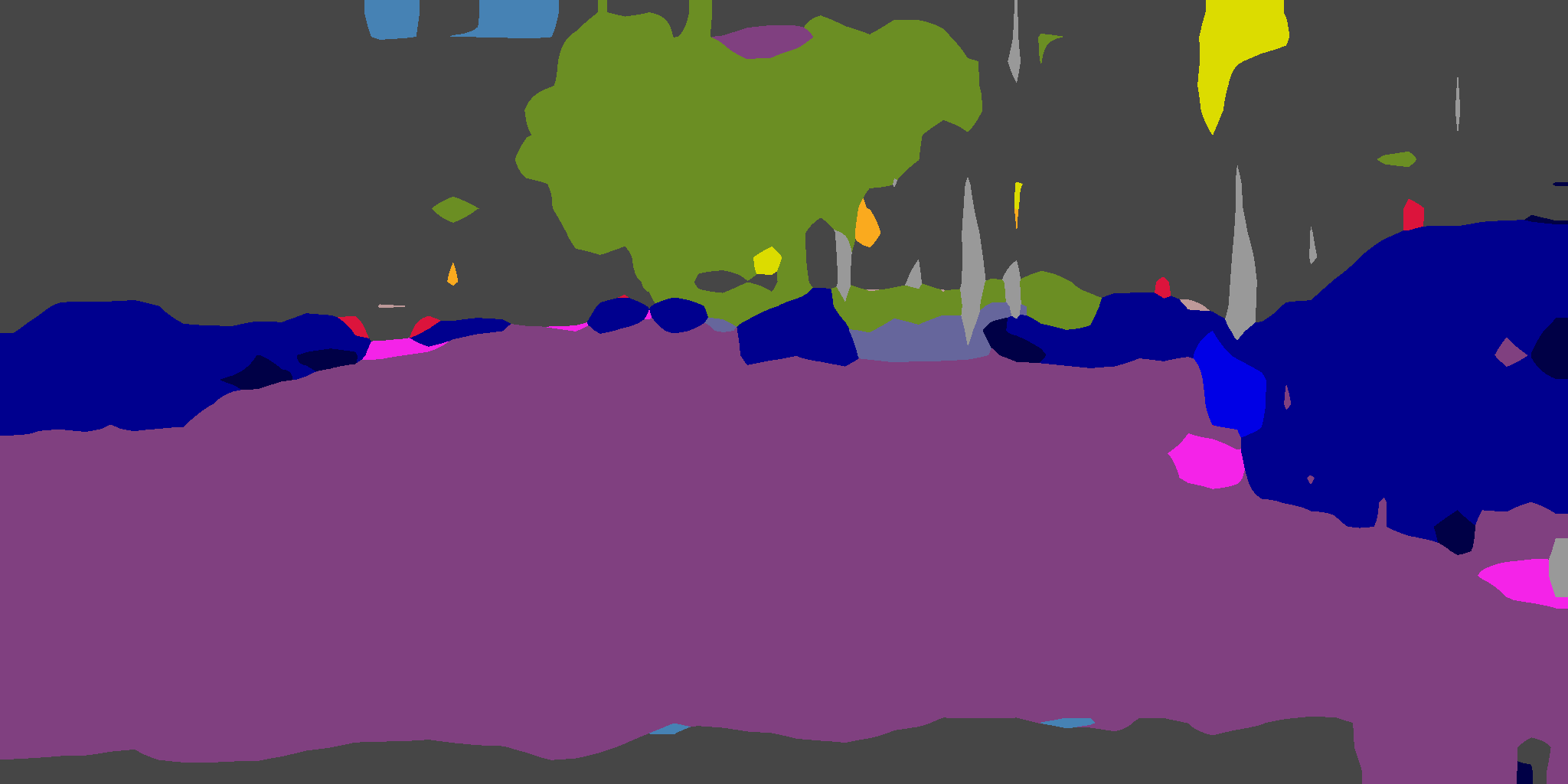}
    \end{subfigure}%
    \end{subfigure}
            \begin{subfigure}{\textwidth}
    \hspace*{.1em}%
    \begin{subfigure}{\imgWidth}
        \includegraphics[width=\textwidth]{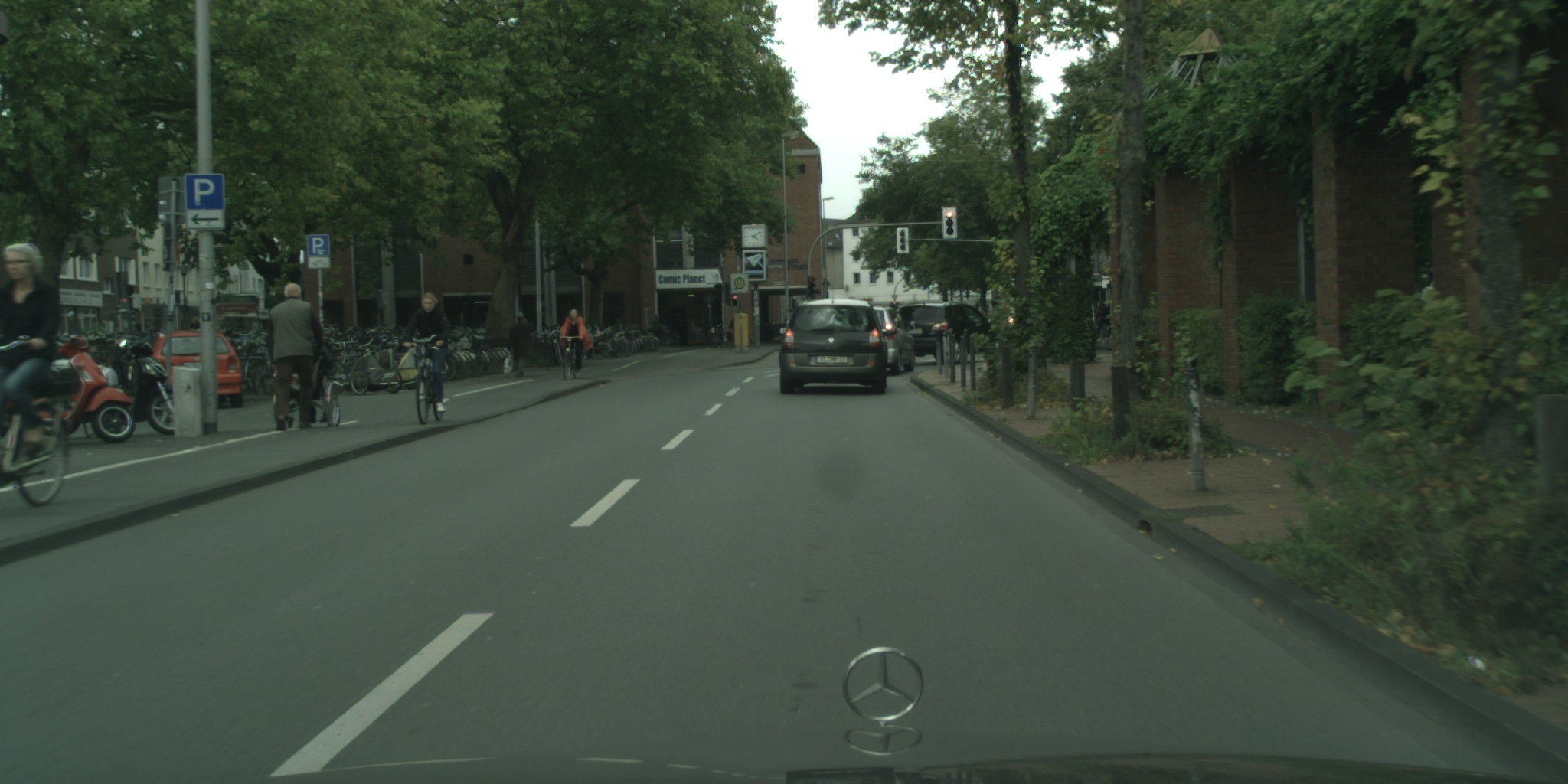}
    \end{subfigure}%
    \begin{subfigure}{\imgWidth}
        \includegraphics[width=\textwidth]{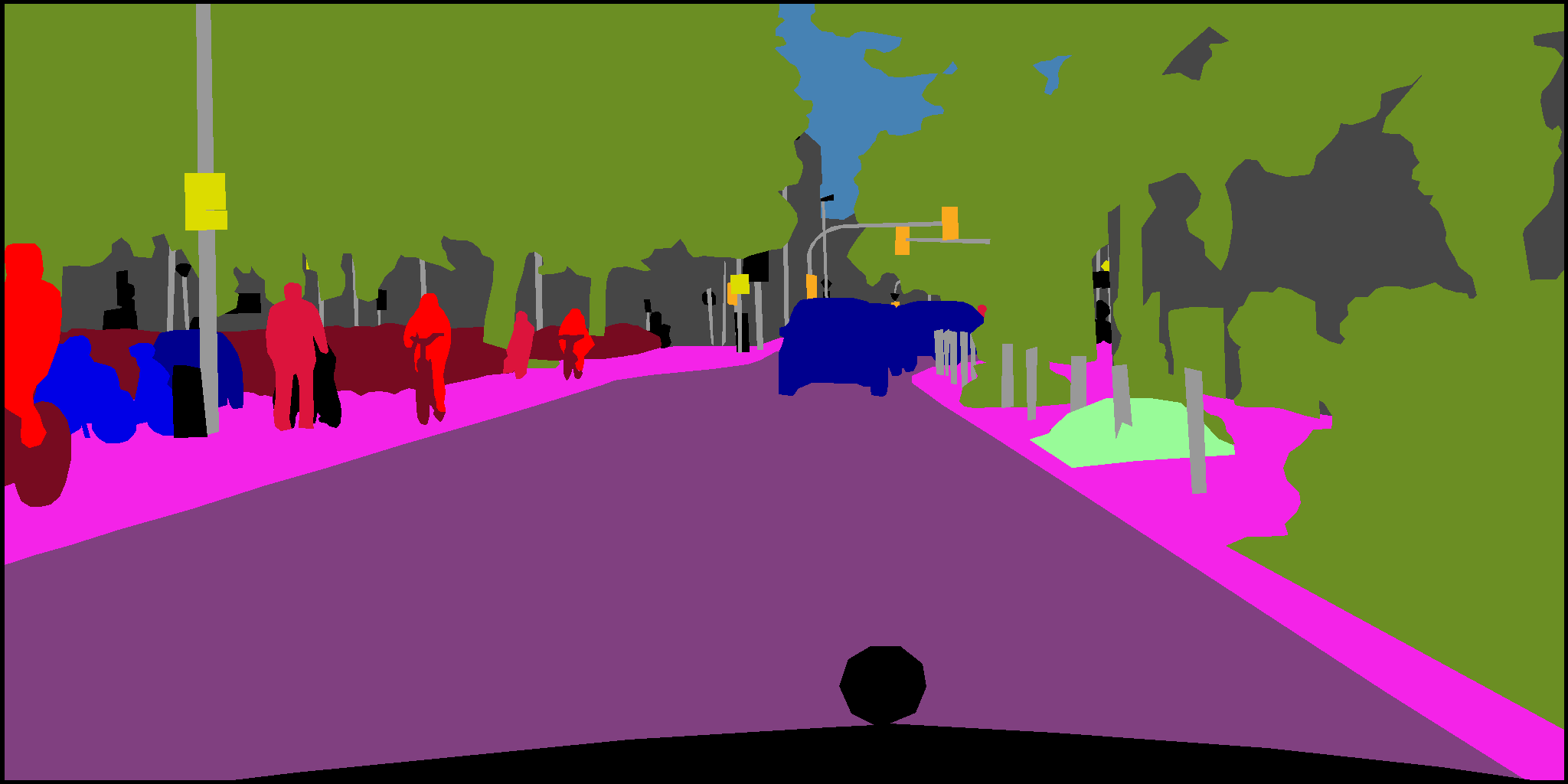}
    \end{subfigure}%
    \begin{subfigure}{\imgWidth}
        \includegraphics[width=\textwidth]{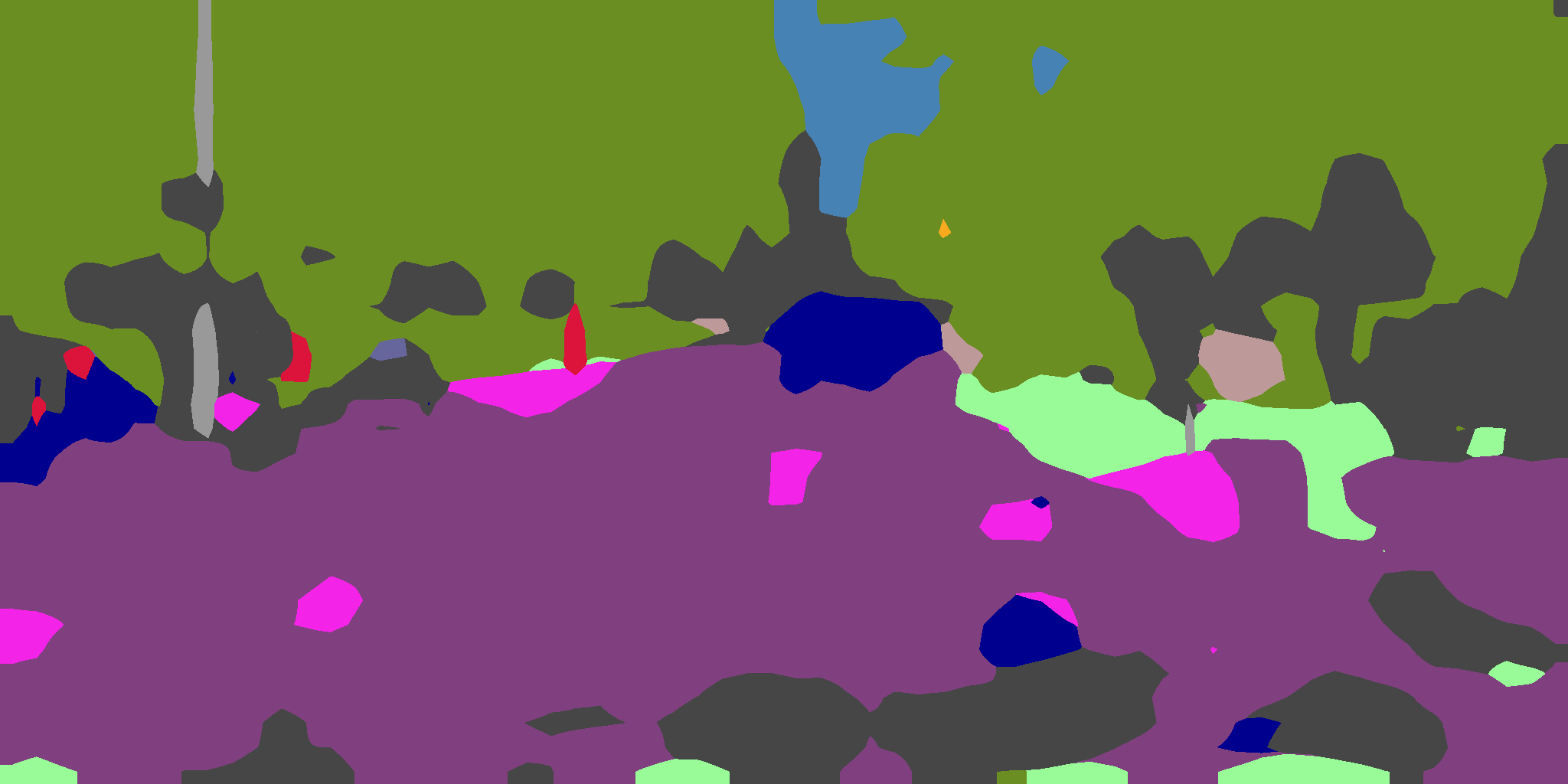}
    \end{subfigure}%
    \begin{subfigure}{\imgWidth}
        \includegraphics[width=\textwidth]{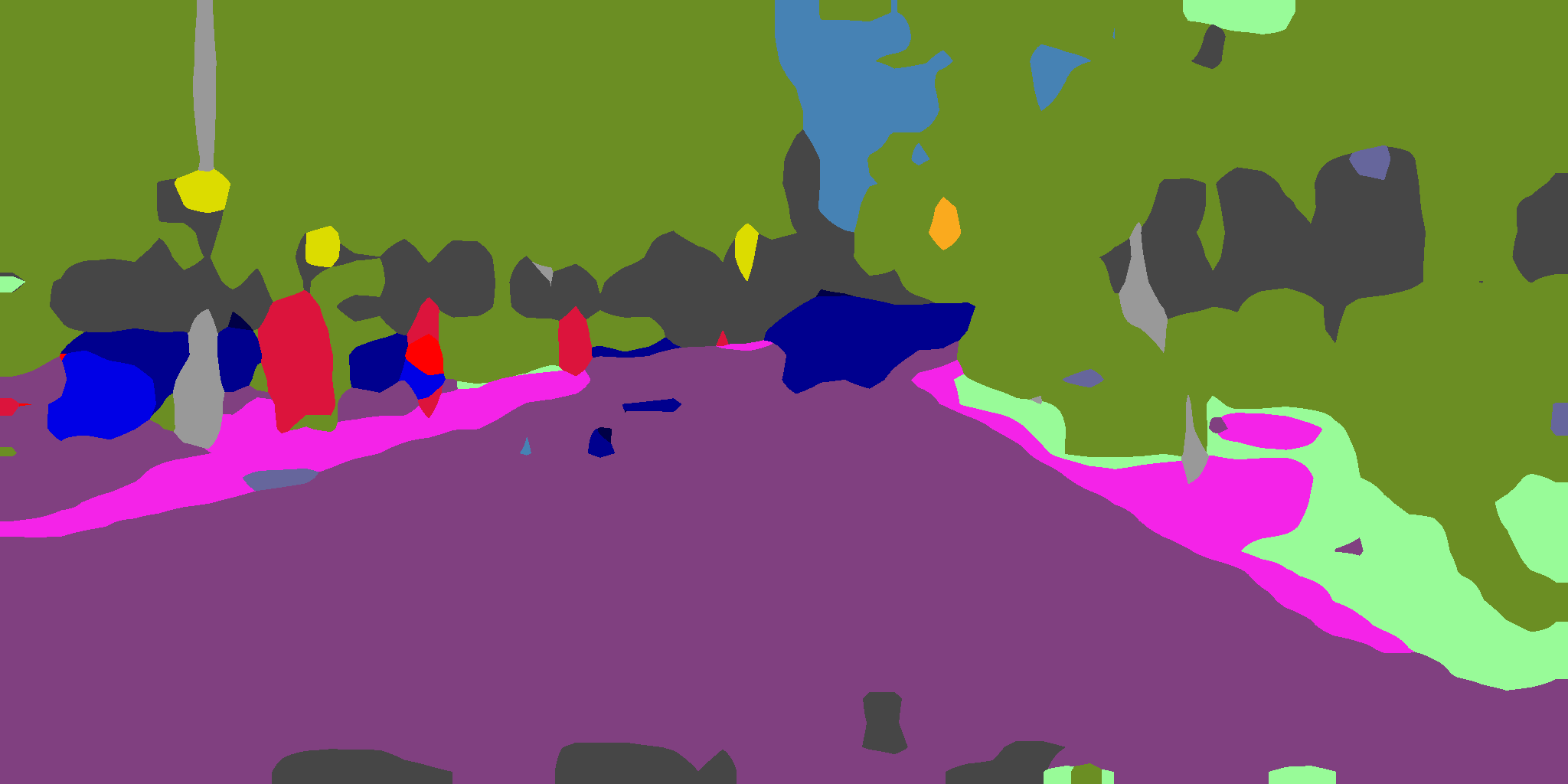}
    \end{subfigure}%
    \begin{subfigure}{\imgWidth}
        \includegraphics[width=\textwidth]{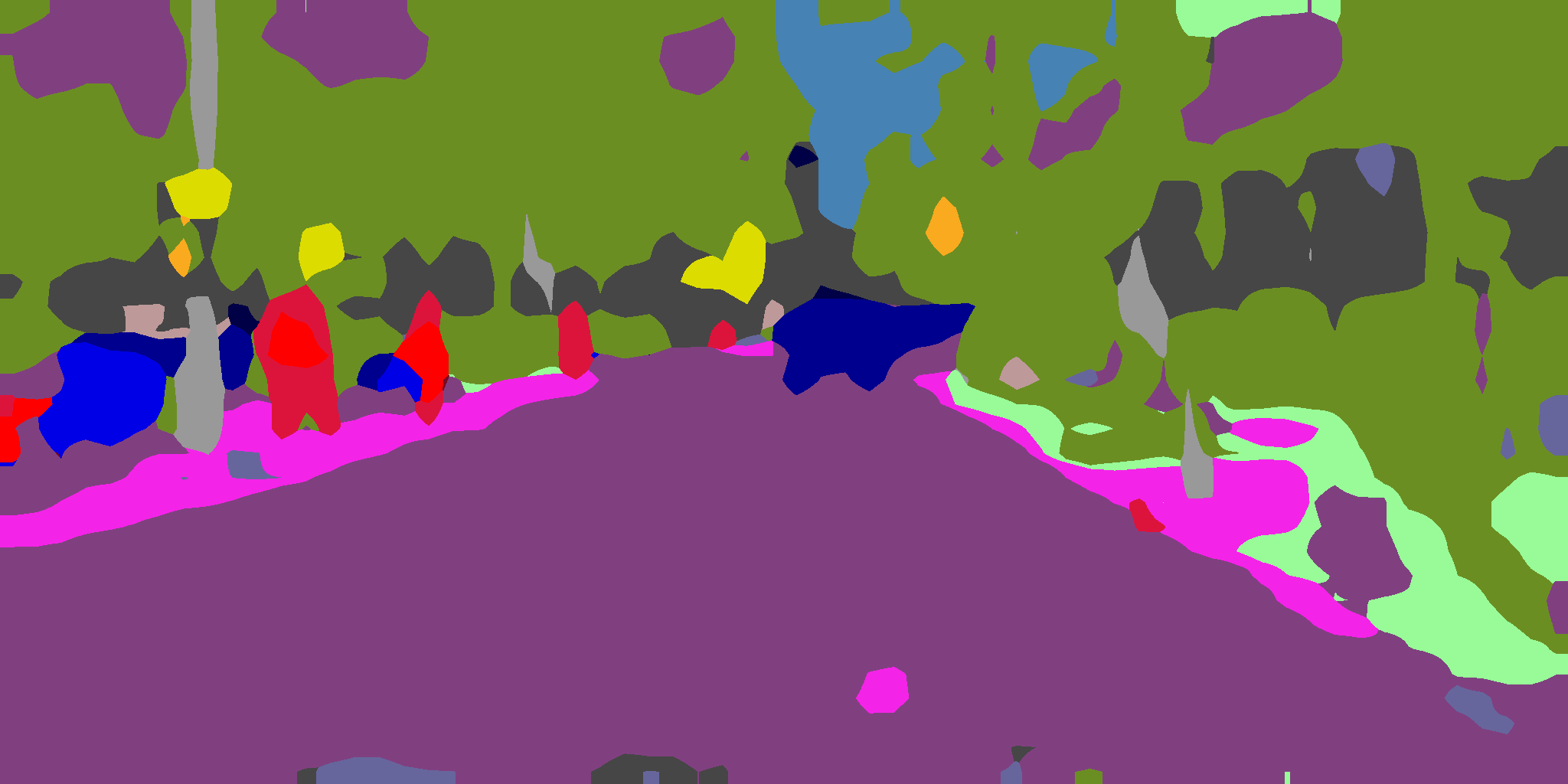}
    \end{subfigure}%
    \end{subfigure}
    \begin{subfigure}{\textwidth}
        \tiny
        \begin{tabularx}{\textwidth}{YYYYYYYYYY}
            \cellcolor{road} \textcolor{white}{Road} & \cellcolor{sidewalk} Sidewalk & \cellcolor{building} \textcolor{white}{Building} & \cellcolor{wall} \textcolor{white}{Wall} & \cellcolor{fence} Fence & \cellcolor{pole} Pole & \cellcolor{tlight} T. Light & \cellcolor{tsign} T. Sign & \cellcolor{vegetation} \textcolor{white}{Vegetation} & \cellcolor{terrain} Terrain \\
            \cellcolor{sky} Sky & \cellcolor{person} \textcolor{white}{Person} & \cellcolor{rider} \textcolor{white}{Rider} & \cellcolor{car} \textcolor{white}{Car} & \cellcolor{truck} \textcolor{white}{Truck} & \cellcolor{bus} \textcolor{white}{Bus} &  \cellcolor{train} \textcolor{white}{Train} & \cellcolor{motorbike} \textcolor{white}{Motorbike} & \cellcolor{bicycle} \textcolor{white}{Bicycle} & \cellcolor{unlabelled} \textcolor{white}{Unlabeled}
        \end{tabularx}
    \end{subfigure}
    \caption{GTA5$\rightarrow$Cityscapes qualitative results.}
    \label{fig:quali_cityscapes}
\end{figure*}

\begin{figure*}[ht]
    \newcolumntype{Y}{>{\centering\arraybackslash}X}
    \centering
    \begin{subfigure}{\textwidth}
    \hspace*{.1em}%
    \begin{subfigure}{\imgWidth}
        \caption{\scriptsize{RGB}}
        \includegraphics[width=\textwidth]{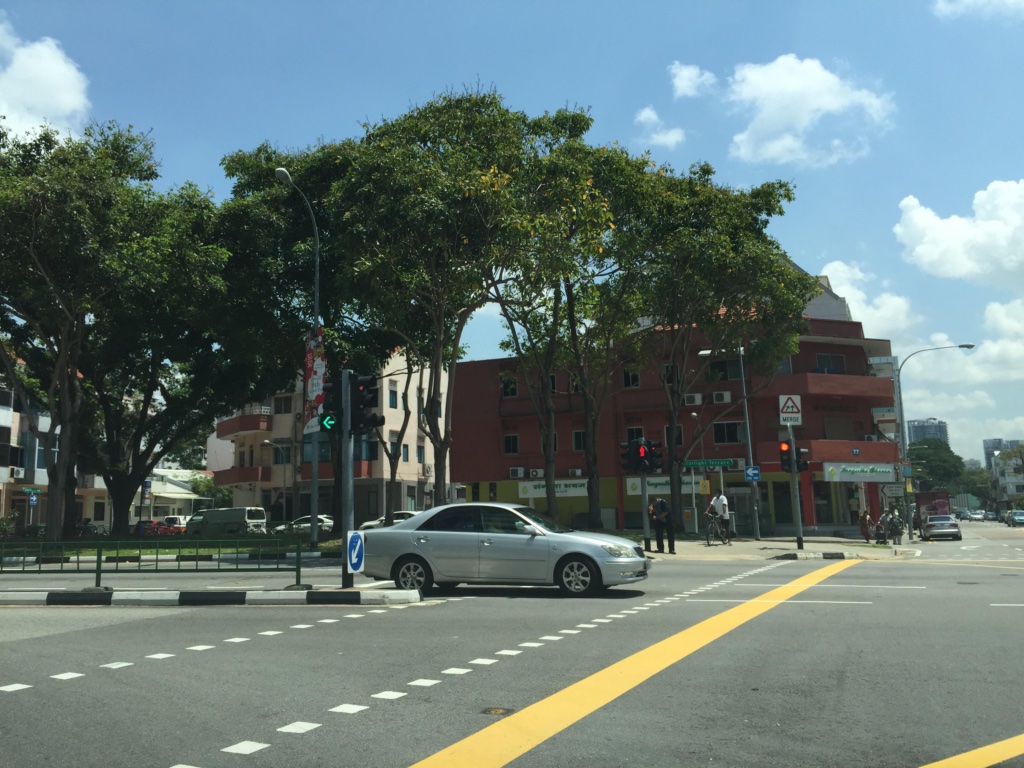}
    \end{subfigure}%
    \begin{subfigure}{\imgWidth}
        \caption{\scriptsize{GT}}
        \includegraphics[width=\textwidth]{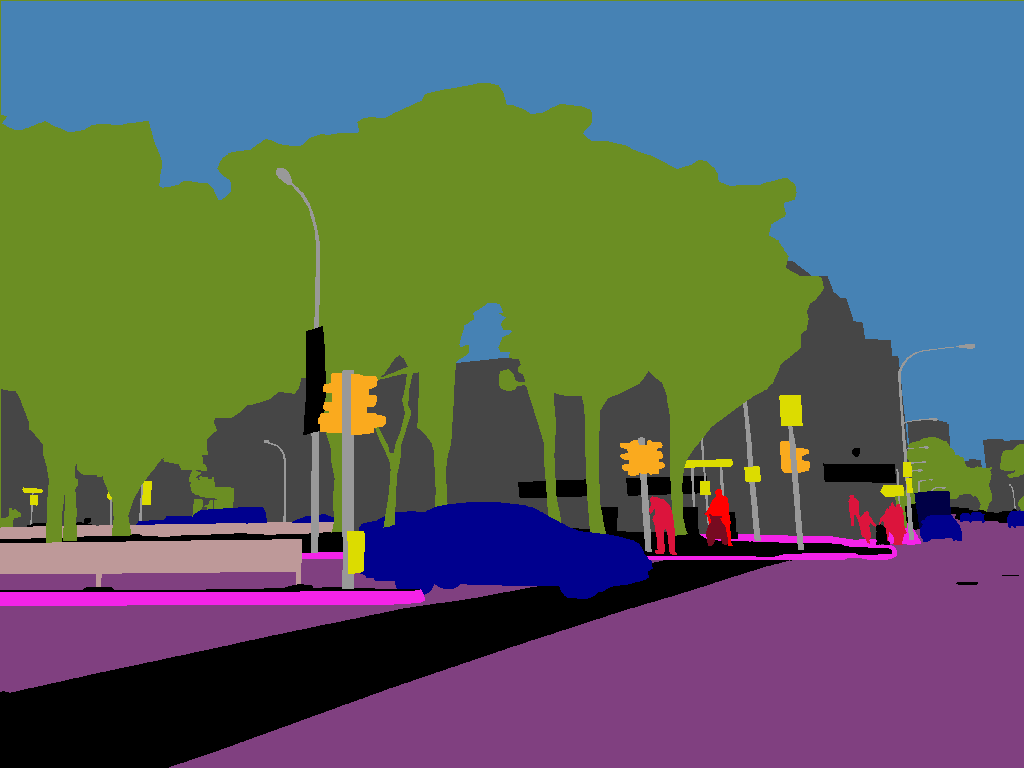}
    \end{subfigure}%
    \begin{subfigure}{\imgWidth}
        \caption{\scriptsize{Source Only}}
        \includegraphics[width=\textwidth]{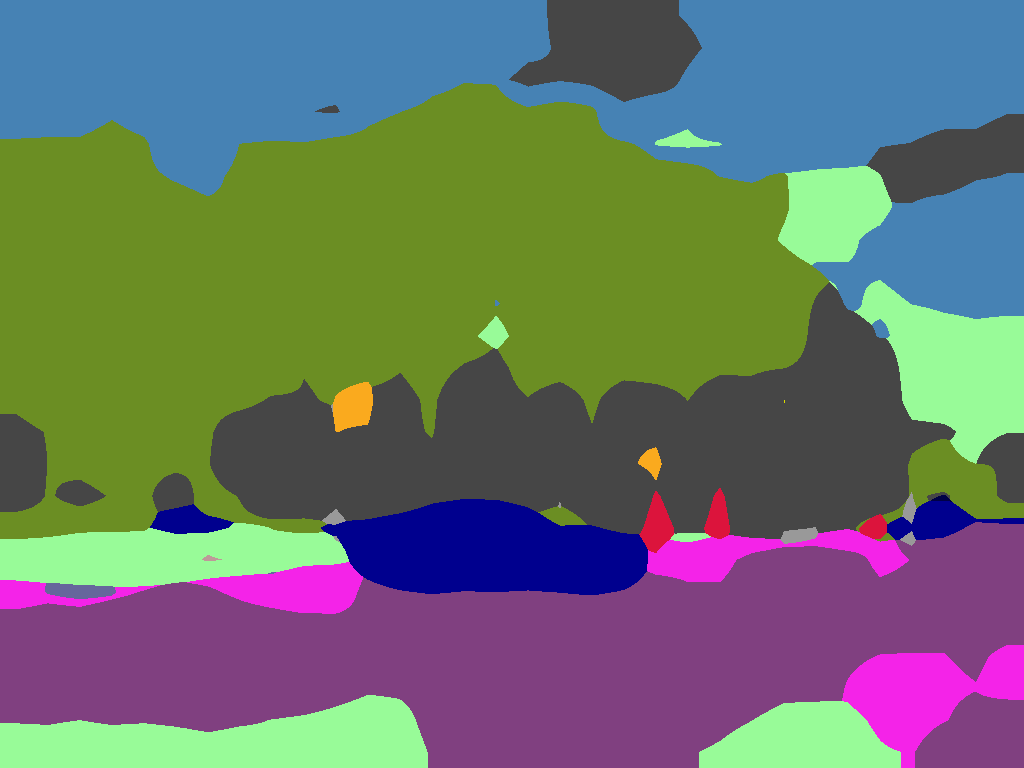}
    \end{subfigure}%
    \begin{subfigure}{\imgWidth}
        \caption{\scriptsize{FedAvg  \cite{fedavg}  + Self-Tr.}}
        \includegraphics[width=\textwidth]{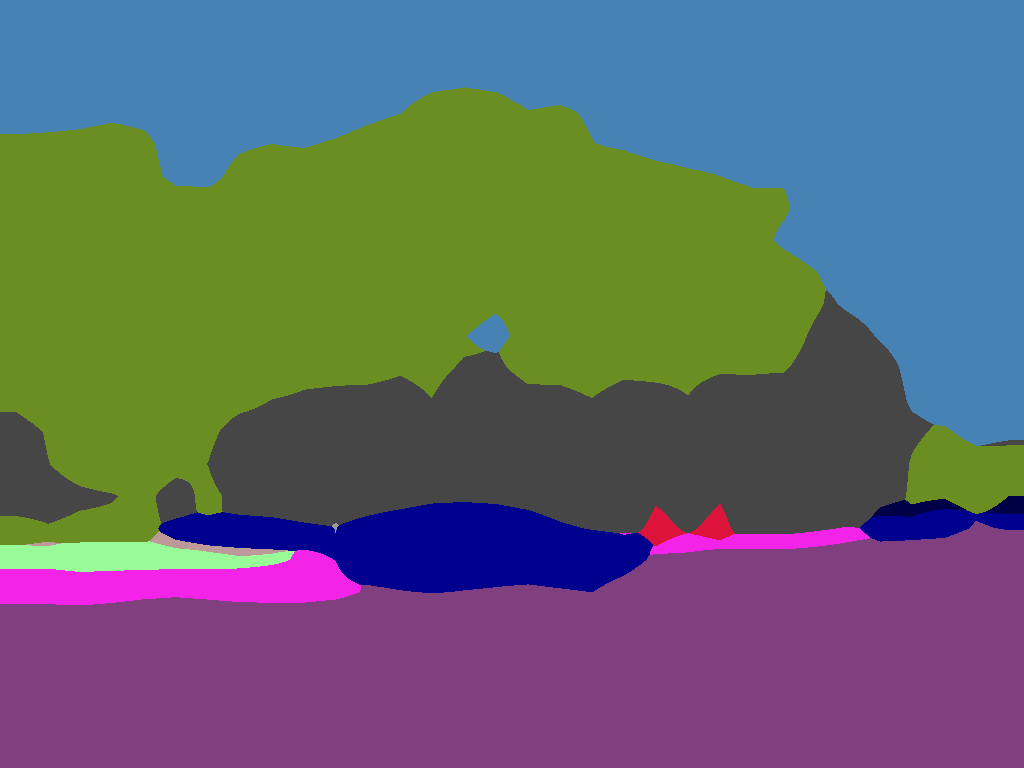}
    \end{subfigure}%
    \begin{subfigure}{\imgWidth}
        \caption{\scriptsize{LADD (all)}}
        \includegraphics[width=\textwidth]{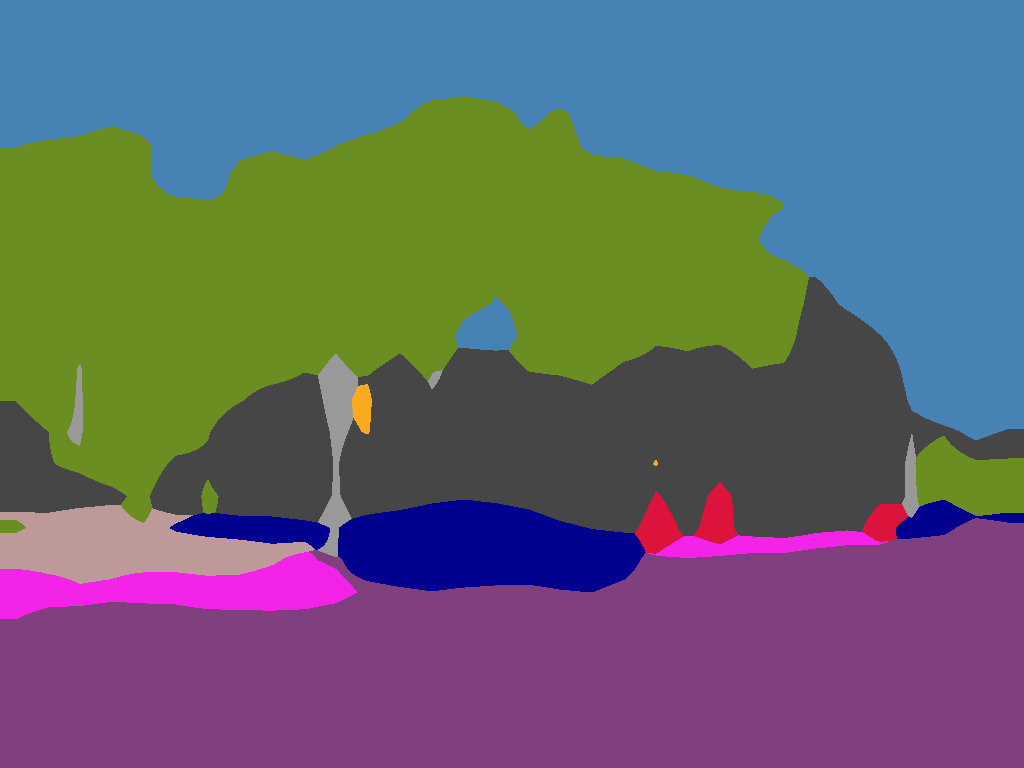}
    \end{subfigure}%
    \end{subfigure}
    \begin{subfigure}{\textwidth}
    \hspace*{.1em}%
    \begin{subfigure}{\imgWidth}
        \includegraphics[width=\textwidth]{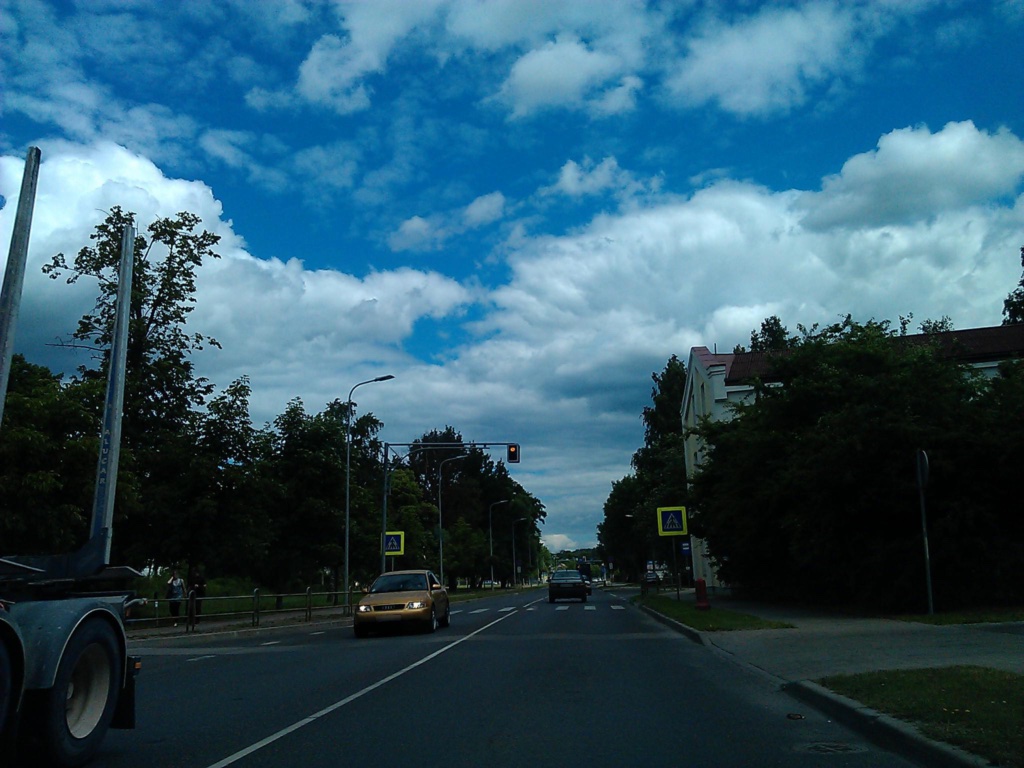}
    \end{subfigure}%
    \begin{subfigure}{\imgWidth}
        \includegraphics[width=\textwidth]{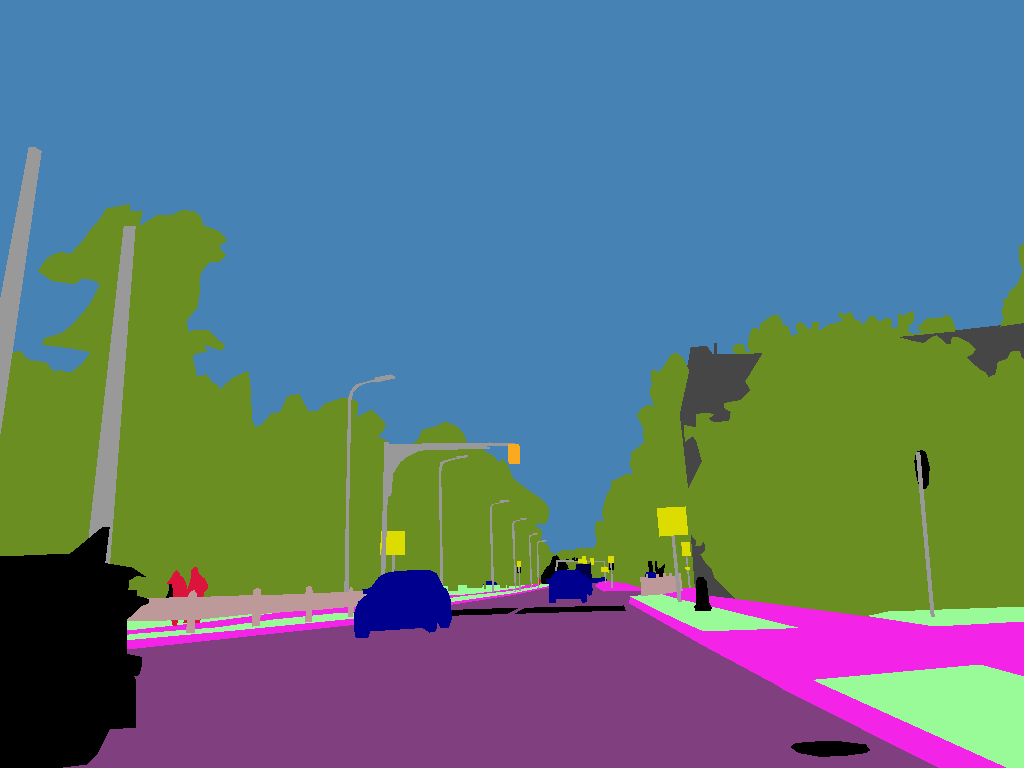}
    \end{subfigure}%
    \begin{subfigure}{\imgWidth}
        \includegraphics[width=\textwidth]{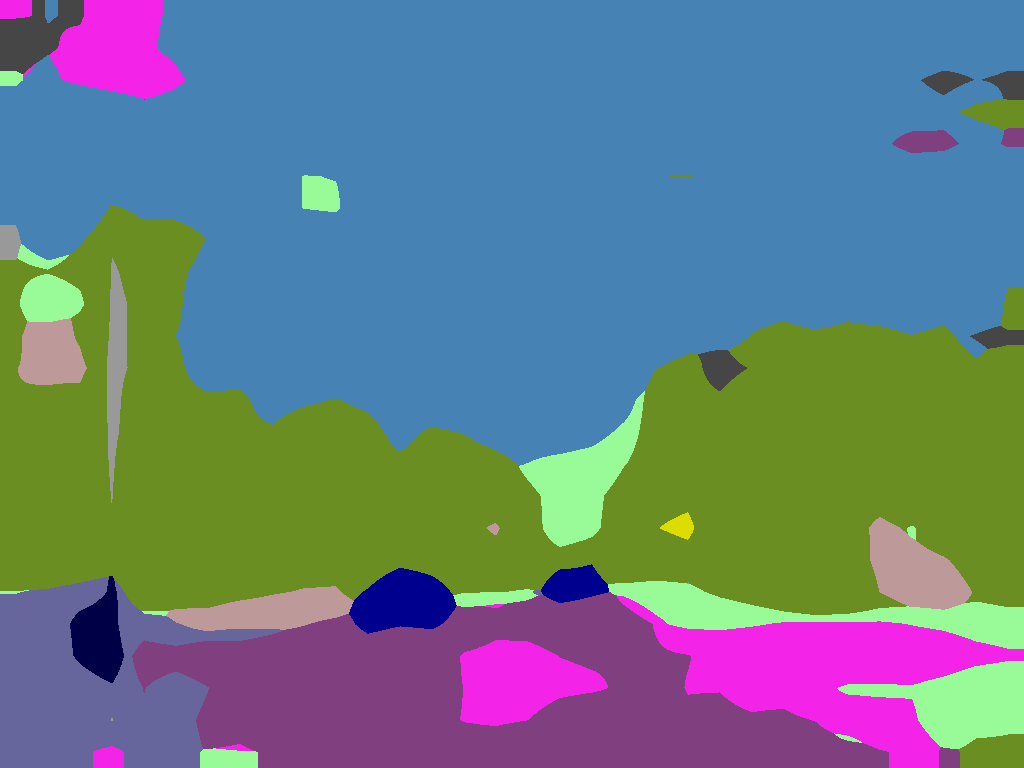}
    \end{subfigure}%
    \begin{subfigure}{\imgWidth}
        \includegraphics[width=\textwidth]{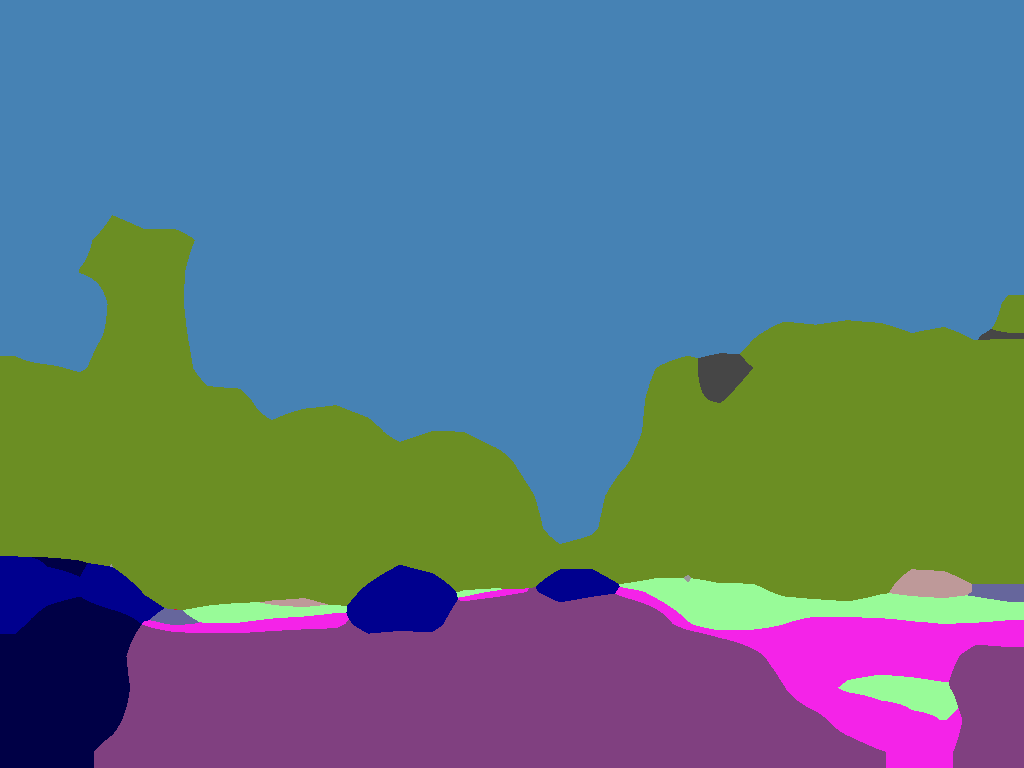}
    \end{subfigure}%
    \begin{subfigure}{\imgWidth}
        \includegraphics[width=\textwidth]{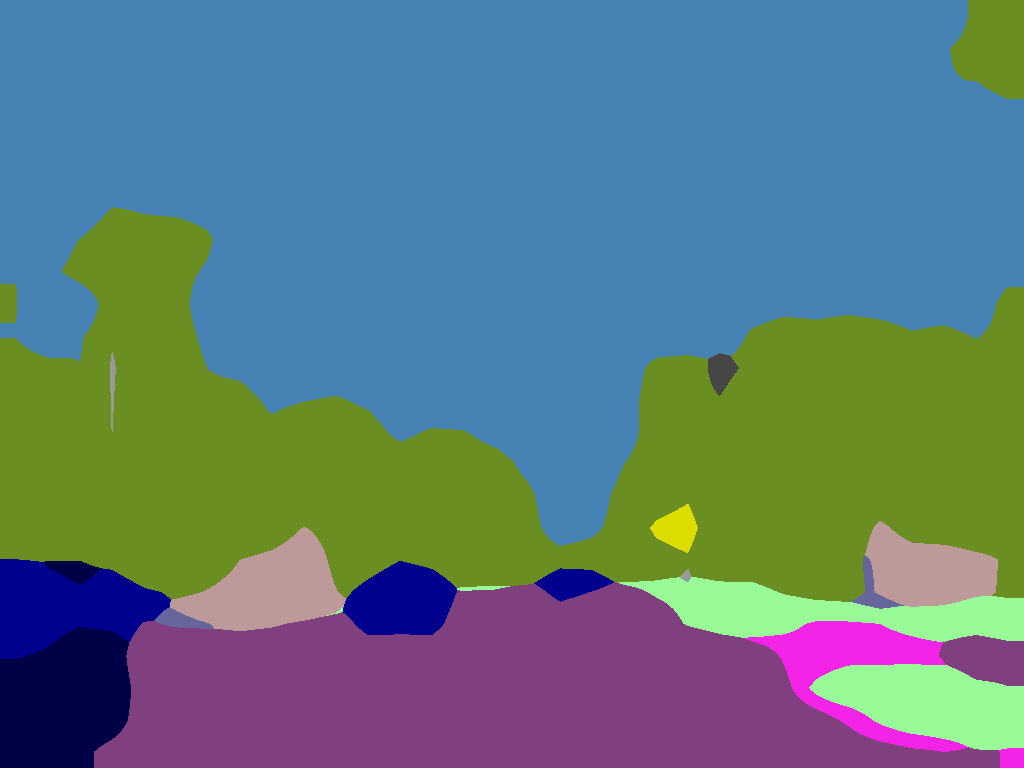}
    \end{subfigure}%
    \end{subfigure}
        \begin{subfigure}{\textwidth}
    \hspace*{.1em}%
    \begin{subfigure}{\imgWidth}
        \includegraphics[width=\textwidth]{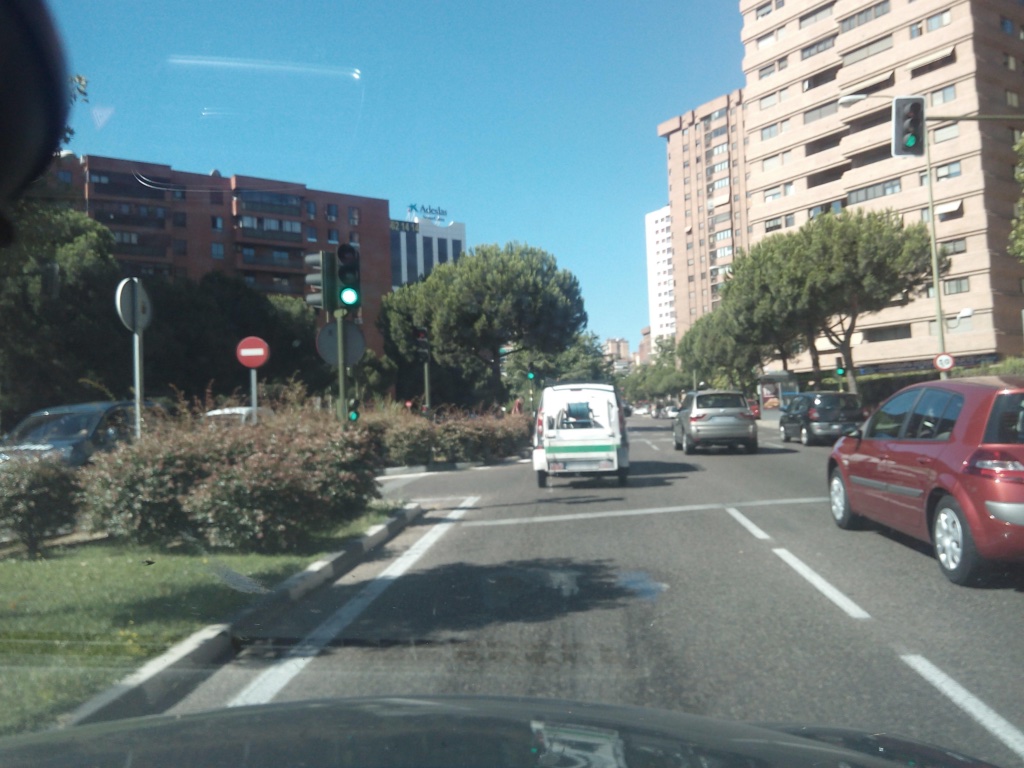}
    \end{subfigure}%
    \begin{subfigure}{\imgWidth}
        \includegraphics[width=\textwidth]{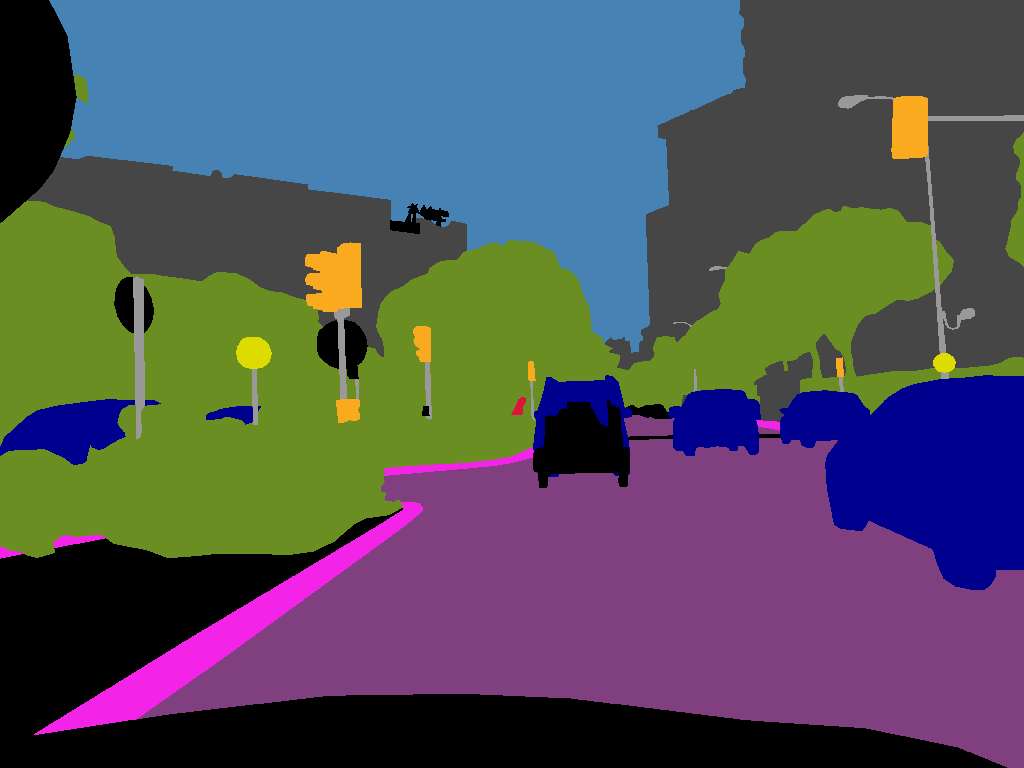}
    \end{subfigure}%
    \begin{subfigure}{\imgWidth}
        \includegraphics[width=\textwidth]{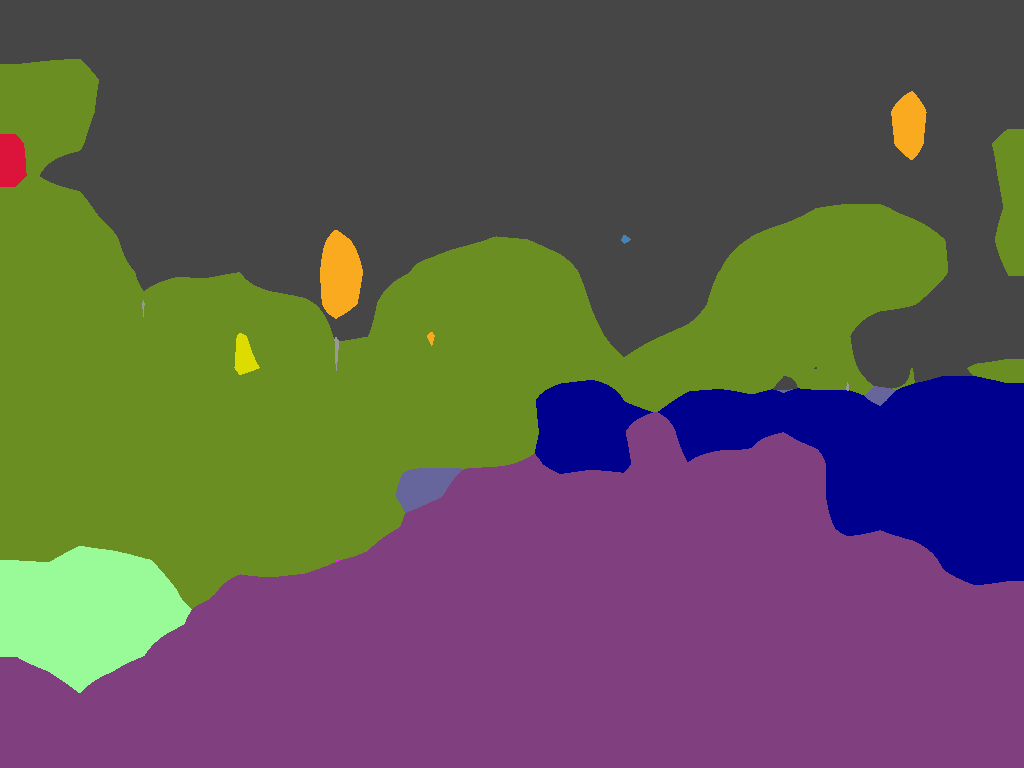}
    \end{subfigure}%
    \begin{subfigure}{\imgWidth}
        \includegraphics[width=\textwidth]{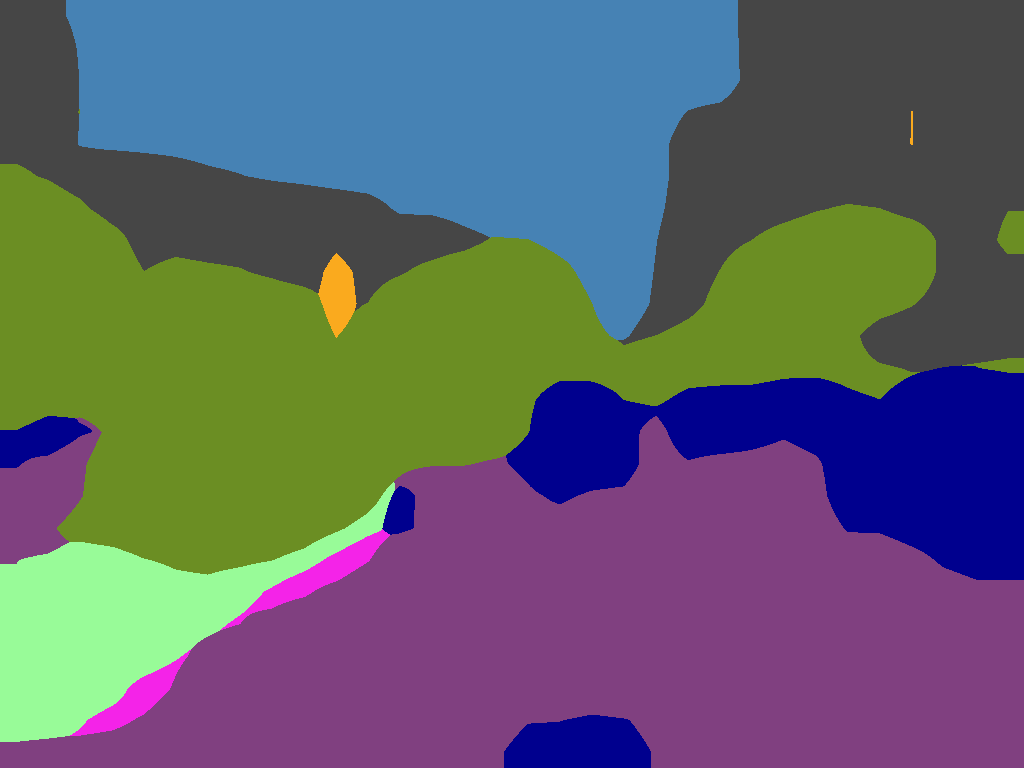}
    \end{subfigure}%
    \begin{subfigure}{\imgWidth}
        \includegraphics[width=\textwidth]{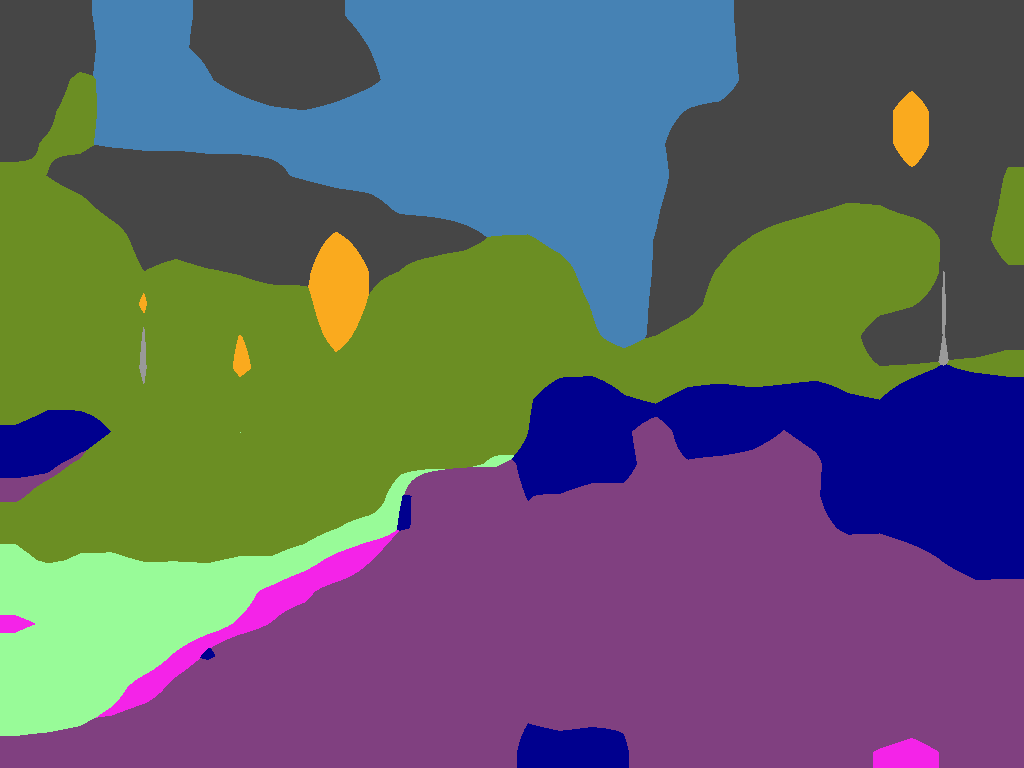}
    \end{subfigure}%
    \end{subfigure}
    \begin{subfigure}{\textwidth}
        \tiny
        \begin{tabularx}{\textwidth}{YYYYYYYYYY}
            \cellcolor{road} \textcolor{white}{Road} & \cellcolor{sidewalk} Sidewalk & \cellcolor{building} \textcolor{white}{Building} & \cellcolor{wall} \textcolor{white}{Wall} & \cellcolor{fence} Fence & \cellcolor{pole} Pole & \cellcolor{tlight} T. Light & \cellcolor{tsign} T. Sign & \cellcolor{vegetation} \textcolor{white}{Vegetation} & \cellcolor{terrain} Terrain \\
            \cellcolor{sky} Sky & \cellcolor{person} \textcolor{white}{Person} & \cellcolor{rider} \textcolor{white}{Rider} & \cellcolor{car} \textcolor{white}{Car} & \cellcolor{truck} \textcolor{white}{Truck} & \cellcolor{bus} \textcolor{white}{Bus} &  \cellcolor{train} \textcolor{white}{Train} & \cellcolor{motorbike} \textcolor{white}{Motorbike} & \cellcolor{bicycle} \textcolor{white}{Bicycle} & \cellcolor{unlabelled} \textcolor{white}{Unlabeled}
        \end{tabularx}
    \end{subfigure}
    \caption{GTA5$\rightarrow$Mapillary qualitative results.}
    \label{fig:quali_map}
\end{figure*}

\section{Qualitative Results} 
We provide some qualitative results in the form of segmentation maps of target images generated by the segmentation model subject to different adaptation schemes.
Figures \ref{fig:quali_cross}, \ref{fig:quali_cityscapes} and \ref{fig:quali_map} refer to the 3 adaptation setups chosen for experimental evaluations, with respectively CrossCity, Cityscapes and Mapillary as target  datasets.
We compare the na\"ive source only training (3rd columns in all the aforementioned figures) and the baseline federated adaptation strategy (4th columns), based on FedAvg\cite{fedavg} aggregation and local self-training, with the proposed LADD (when cluster-specific aggregation is extended to all the segmentation network layers) (last columns). For fair comparison we employ the same pretraining for FedAvg and LADD.
By inspecting the segmentation maps produced by the different adaptation strategies, we notice how the \textit{source only} maps show inconsistent and noisy predictions, where semantically similar classes are confused, such as \textit{sidewalk} and \textit{road} or \textit{terrain} in all the reported samples.
Local self-training and standard FedAvg aggregation at server-side partially mitigate the prediction accuracy drop caused by domain shift between source and target data. 
Nonetheless, we observe that the adapted model still tends to mistake semantically-similar classes such as sidewalk and road in the first sample of Figure~\ref{fig:quali_cross}.
The proposed regularized local training leads to more robust local optimization, which otherwise tends to suffer from unsteady behavior, due to the small amount of available training data and the lack of any form of supervision (even from the source domain) at the client side. 
This, along with the cluster-specific semantically aware aggregation mechanism, results into less noisy and more accurate predictions as we can see in the last columns of the figures. 

\section{Additional Quantitative Results}

Finally, we report additional results in the form of per-class IoUs achieved when different modules of our framework are enabled.
Once more, results are reported with CrossCity (Table \ref{tab:perclass_cross}), Cityscapes (Table \ref{tab:perclass_city})  and Mapillary (Table \ref{tab:perclass_map}) as target datasets, in terms of mean and standard deviation computed over the last $10\%$ rounds.

When enabled, we observe that each module improves the overall mIoU score, which is also generally shared by the individual IoU scores of the semantic classes in the different experimental setups.

\begin{figure*}[!ht]
    \newcolumntype{Y}{>{\centering\arraybackslash}X}
    \centering
    \includegraphics[width=0.68\textwidth]{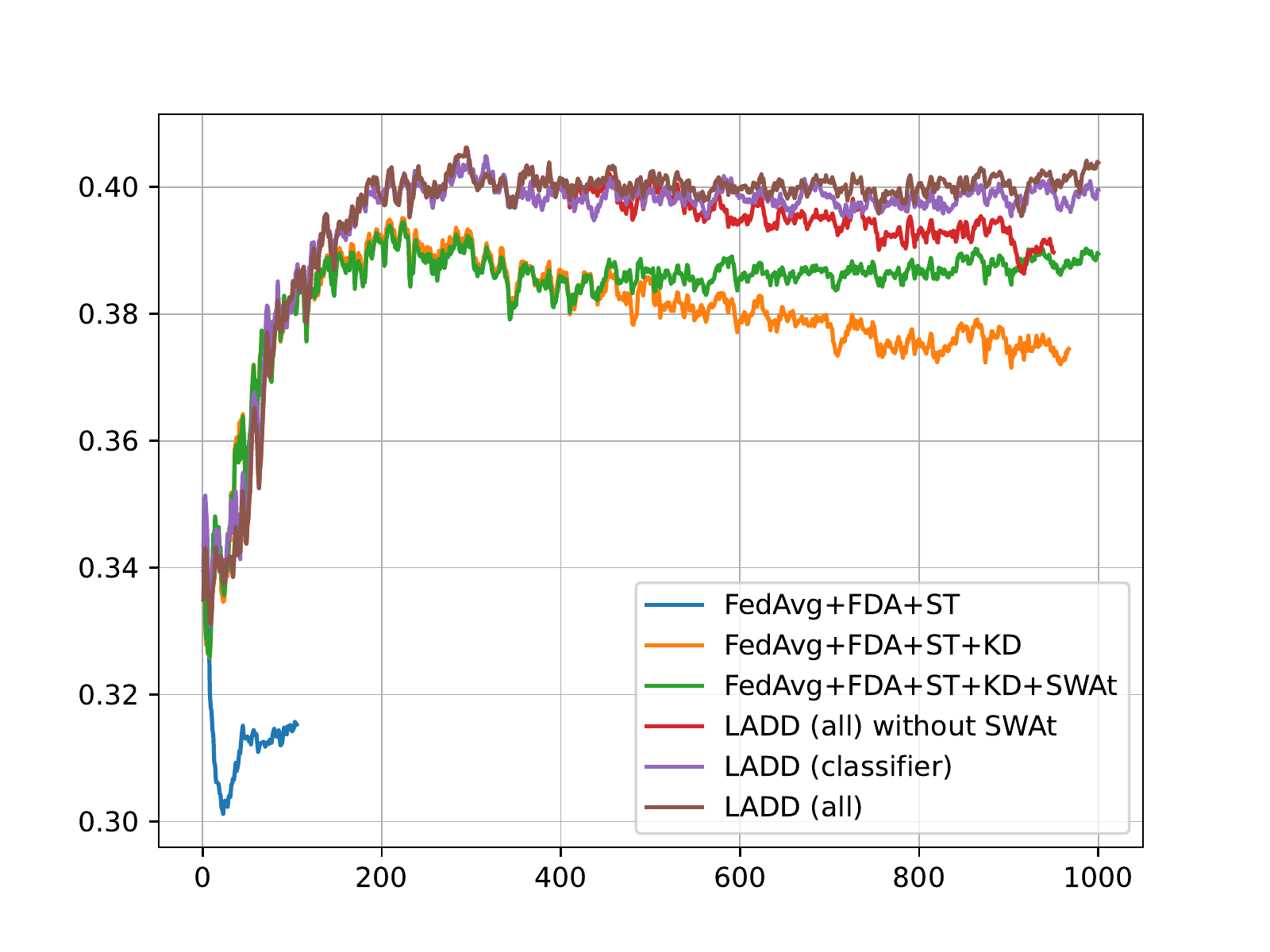}
    \caption{Comparison of learning curves in the CrossCity federated split.}

    \label{fig:runs_crosscity}
\end{figure*}

In addition, in Figure \ref{fig:runs_crosscity} we report the learning curves as a result of federated optimization under different configurations of the proposed LADD method in the GTA$\rightarrow$CrossCity setup.
When only ST is employed in the client-side optimization, the training is extremely unstable, showing a small initial burst of performance followed by a rapid decrease after few rounds.
When adding KD and then SWAt, the training curves become progressively more robust and stable, achieving the best results when KD and SWAt are joined by the cluster-specific aggregation, in either classifier-exclusive or full model configuration of cluster-specific parameters.
We finally remark how LADD in its complete configuration is characterized by steady and converging learning curves, unaffected by diverging phenomena.

\begin{table*}[t]
    \caption{CrossCity IoU by class and mIoU (\%).}
    \setlength{\tabcolsep}{3.5pt}
    \label{tab:iou_xc}
    \centering
    \footnotesize
    \begin{adjustbox}{width=1.0\linewidth}
        \begin{tabular}{ccccc|ccccccccccccc|c}
        \toprule
            \rotatebox[origin=c]{90}{\textbf{FDA}} & \rotatebox[origin=c]{90}{\textbf{ST}} & \rotatebox[origin=c]{90}{\textbf{KD}} & \rotatebox[origin=c]{90}{\textbf{SWAt}} & \rotatebox[origin=c]{90}{\textbf{Cl Aggr}} & \rotatebox[origin=c]{90}{road} & \rotatebox[origin=c]{90}{sidewalk} & \rotatebox[origin=c]{90}{building} & \rotatebox[origin=c]{90}{traffic light} & \rotatebox[origin=c]{90}{traffic sign} & \rotatebox[origin=c]{90}{vegetation} & \rotatebox[origin=c]{90}{sky} & \rotatebox[origin=c]{90}{person} & \rotatebox[origin=c]{90}{rider} & \rotatebox[origin=c]{90}{car} & \rotatebox[origin=c]{90}{bus} & \rotatebox[origin=c]{90}{motorcycle} & \rotatebox[origin=c]{90}{bicycle} & \rotatebox[origin=c]{90}{\textbf{mIoU}} \\
            \midrule
             & & & & & $25.6$ & $21.6$ & $65.9$ & $3.9$ & $8.6$ & $67.5$ & $73.5$ & $33.1$ & $2.1$ & $43.0$ & $6.6$ & $0.3$ & $0.2$ & $26.5 \pm 1.5$ \\
            \cmark & & & & & $38.2$ & $24.0$ & $74.8$ & $7.0$ & $8.9$ & $70.5$ & $80.9$ & $37.0$ & $4.0$ & $63.6$ & $12.0$ & $3.5$ & $0.0$ & $32.4 \pm 0.6$ \\
            \cmark & \cmark & & & & $21.9$ & $17.9$ & $81.3$ & $9.5$ & $14.5$ & $77.4$ & $85.2$ & $41.0$ & $2.3$ & $66.1$ & $10.7$ & $8.0$ & $0.9$ & $33.6 \pm 1.3$ \\
            \cmark & \cmark & \cmark & & & $49.5$ & $26.7$ & $81.2$ & $11.7$ & $12.4$ & $77.6$ & $87.0$ & $40.4$ & $1.0$ & $68.9$ & $16.3$ & $11.3$ & $3.4$ & $37.5 \pm 0.1$ \\
            \cmark & \cmark & \cmark & \cmark & & $53.3$ & $28.6$ & $81.1$ & $12.1$ & $12.0$ & $77.5$ & $87.1$ & $42.1$ & $1.9$ & $68.9$ & $17.2$ & $16.7$ & $4.9$ & $38.8 \pm 0.1$ \\
            \cmark & \cmark & \cmark & & \cmark & $63.4$ & $32.3$ & $81.5$ & $12.1$ & $12.2$ & $77.5$ & $86.9$ & $41.0$ & $1.1$ & $69.0$ & $16.0$ & $12.5$ & $3.8$ & $39.2 \pm 0.2$ \\
            \cmark & \cmark & \cmark & \cmark & \cmark & $64.3$ & $33.7$ & $81.0$ & $12.5$ & $14.4$ & $77.2$ & $86.8$ & $42.1$ & $1.4$ & $69.1$ & $18.1$ & $15.6$ & $4.8$ & $40.1 \pm 0.2$ \\
        \bottomrule
        \end{tabular}
    \end{adjustbox}
    \vspace{-5pt}
     \label{tab:perclass_cross}
\end{table*}

\begin{table*}[t]
    \caption{Cityscapes IoU by class and mIoU (\%).}
    \setlength{\tabcolsep}{3.5pt}
    \label{tab:iou_cts}
    \centering
    \footnotesize
    \begin{adjustbox}{width=1.0\linewidth}
        \begin{tabular}{ccccc|ccccccccccccccccccc|c}
        \toprule
            \rotatebox[origin=c]{90}{\textbf{FDA}} & \rotatebox[origin=c]{90}{\textbf{ST}} & \rotatebox[origin=c]{90}{\textbf{KD}} & \rotatebox[origin=c]{90}{\textbf{SWAt}} & \rotatebox[origin=c]{90}{\textbf{Cl Aggr}} & \rotatebox[origin=c]{90}{road} & \rotatebox[origin=c]{90}{sidewalk} & \rotatebox[origin=c]{90}{building} & \rotatebox[origin=c]{90}{wall} & \rotatebox[origin=c]{90}{fence} & \rotatebox[origin=c]{90}{pole} & \rotatebox[origin=c]{90}{traffic light} & \rotatebox[origin=c]{90}{traffic sign} & \rotatebox[origin=c]{90}{vegetation} & \rotatebox[origin=c]{90}{terrain} & \rotatebox[origin=c]{90}{sky} & \rotatebox[origin=c]{90}{person} & \rotatebox[origin=c]{90}{rider} & \rotatebox[origin=c]{90}{car} & \rotatebox[origin=c]{90}{truck} & \rotatebox[origin=c]{90}{bus} & \rotatebox[origin=c]{90}{train} & \rotatebox[origin=c]{90}{motorcycle} & \rotatebox[origin=c]{90}{bicycle} & \rotatebox[origin=c]{90}{\textbf{mIoU}} \\
            \midrule
            \cmark & \cmark & & & & $84.5$ & $36.8$ & $77.3$ & $23.9$ & $11.3$ & $20.5$ & $29.1$ & $22.6$ & $76.9$ & $26.5$ & $68.9$ & $53.4$ & $13.7$ & $79.0$ & $15.2$ & $14.0$ & $1.4$ & $11.0$ & $5.1$ & $35.1 \pm 0.7$ \\
            \cmark & \cmark & \cmark & & & $79.3$ & $34.0$ & $73.6$ & $22.0$ & $16.4$ & $24.6$ & $30.3$ & $31.3$ & $61.7$ & $23.2$ & $70.1$ & $51.2$ & $19.3$ & $73.7$ & $13.6$ & $17.9$ & $7.3$ & $12.1$ & $15.3$ & $35.6 \pm 0.1$ \\
            \cmark & \cmark & \cmark & \cmark & \cmark & $80.0$ & $36.1$ & $74.1$ & $22.8$ & $18.3$ & $26.3$ & $30.6$ & $33.0$ & $65.2$ & $25.4$ & $69.4$ & $52.3$ & $19.1$ & $74.5$ & $13.4$ & $18.0$ & $7.2$ & $12.6$ & $14.2$ & $36.5 \pm 0.1$ \\
        \bottomrule
        \end{tabular}
    \end{adjustbox}
    \vspace{-5pt}
     \label{tab:perclass_city}
\end{table*}

\begin{table*}[!t]
    \caption{Mapillary Vistas IoU by class and mIoU (\%).}
    \setlength{\tabcolsep}{3.5pt}
    \label{tab:iou_map}
    \centering
    \footnotesize
    \begin{adjustbox}{width=1.0\linewidth}
        \begin{tabular}{ccccc|ccccccccccccccccccc|c}
        \toprule
            \rotatebox[origin=c]{90}{\textbf{FDA}} & \rotatebox[origin=c]{90}{\textbf{ST}} & \rotatebox[origin=c]{90}{\textbf{KD}} & \rotatebox[origin=c]{90}{\textbf{SWAt}} & \rotatebox[origin=c]{90}{\textbf{Cl Aggr}} & \rotatebox[origin=c]{90}{road} & \rotatebox[origin=c]{90}{sidewalk} & \rotatebox[origin=c]{90}{building} & \rotatebox[origin=c]{90}{wall} & \rotatebox[origin=c]{90}{fence} & \rotatebox[origin=c]{90}{pole} & \rotatebox[origin=c]{90}{traffic light} & \rotatebox[origin=c]{90}{traffic sign} & \rotatebox[origin=c]{90}{vegetation} & \rotatebox[origin=c]{90}{terrain} & \rotatebox[origin=c]{90}{sky} & \rotatebox[origin=c]{90}{person} & \rotatebox[origin=c]{90}{rider} & \rotatebox[origin=c]{90}{car} & \rotatebox[origin=c]{90}{truck} & \rotatebox[origin=c]{90}{bus} & \rotatebox[origin=c]{90}{train} & \rotatebox[origin=c]{90}{motorcycle} & \rotatebox[origin=c]{90}{bicycle} & \rotatebox[origin=c]{90}{\textbf{mIoU}} \\
            \midrule
            \cmark & \cmark & & & & $67.4$ & $36.9$ & $74.7$ & $24.8$ & $25.4$ & $10.9$ & $21.0$ & $33.3$ & $72.8$ & $40.8$ & $91.2$ & $46.1$ & $23.1$ & $73.7$ & $31.1$ & $22.7$ & $3.1$ & $30.6$ & $11.9$ & $39.0 \pm 0.2$ \\
            \cmark & \cmark & \cmark & \cmark & & $75.4$ & $37.7$ & $73.4$ & $25.2$ & $25.2$ & $18.3$ & $26.6$ & $37.1$ & $73.5$ & $38.1$ & $91.4$ & $45.5$ & $13.8$ & $71.3$ & $30.9$ & $22.0$ & $3.0$ & $29.9$ & $19.1$ & $40.0 \pm 0.1$ \\
            \cmark & \cmark & \cmark & \cmark & \cmark & $75.5$ & $37.0$ & $69.1$ & $24.6$ & $25.6$ & $18.9$ & $26.7$ & $38.2$ & $72.5$ & $36.4$ & $89.4$ & $46.3$ & $17.2$ & $70.7$ & $32.6$ & $20.2$ & $4.1$ & $31.4$ & $21.0$ & $40.2 \pm 1.0$ \\
        \bottomrule
        \end{tabular}
    \end{adjustbox}
    \vspace{-5pt}
    \label{tab:perclass_map}
\end{table*}

\end{document}